\documentclass[letterpaper]{article}

\usepackage{amsmath}
\usepackage{amssymb}
\usepackage{longtable}
\usepackage{booktabs}

\usepackage[preprint]{neurips_2025}
\usepackage{graphicx} 

\usepackage{array} 

\graphicspath{{attention_images/}}

\newcolumntype{C}[1]{>{\centering\arraybackslash}m{#1}}

\usepackage{graphicx}
\usepackage{caption}
\usepackage{subcaption}
\usepackage[verbose]{placeins}

\usepackage{hyperref}
\hypersetup{
    colorlinks=true,
    linkcolor=blue,
    filecolor=magenta,
    urlcolor=cyan,
    citecolor=blue
}

\usepackage{array}
\usepackage[utf8]{inputenc} 
\usepackage[T1]{fontenc}    
\usepackage{hyperref}       
\usepackage{url}            
\usepackage{booktabs}       
\usepackage{amsfonts}       
\usepackage{nicefrac}       
\usepackage{microtype}      
\usepackage{xcolor}         

\title{Decomposing Attention To Find Context-Sensitive Neurons}

\author{%
  Alex Gibson \\
  University of Cambridge\\
  \text{ag2299@cam.ac.uk} \\
}

\date{}
\begin{document}
\renewcommand{\sectionautorefname}{Section} 

\maketitle

\begin{abstract}
We study transformer language models, analyzing attention heads whose attention patterns are spread out, and whose attention scores depend weakly on content. We argue that the softmax denominators of these heads are stable when the underlying token distribution is fixed. By sampling softmax denominators from a "calibration text", we can combine together the outputs of multiple such stable heads in the first layer of GPT2-Small, approximating their combined output by a linear summary of the surrounding text. This approximation enables a procedure where from the weights alone - and a single calibration text - we can uncover hundreds of first layer neurons that respond to high-level contextual properties of the surrounding text, including neurons that didn't activate on the calibration text.
 
\end{abstract}
\begin{section}{Introduction:}

Transformer language models have achieved remarkable capabilities, yet our mechanistic understanding of how they represent text remains limited. Prior work such as the Mathematical Framework for Transformer Circuits (\citeauthor{elhage2021mathematical}, \citeyear{elhage2021mathematical}) has helped to reveal components of models which implement specific algorithmic behaviors, such as induction heads (\citeauthor{olsson2022incontextlearninginductionheads}, \citeyear{olsson2022incontextlearninginductionheads}), or duplicate token heads (\citeauthor{wang2023interpretability}, \citeyear{wang2023interpretability}). These results illustrate that in some cases attention heads implement clear symbolic algorithms.

Building on these insights into token-level pattern recognition, a natural next question concerns how transformers represent higher-level, distributed properties of text-such as writing style, domain, or linguistic variety-that emerge from broader contextual patterns rather than individual tokens. While existing approaches like Sparse Autoencoders (\citeauthor{bricken2023monosemanticity}, \citeyear{bricken2023monosemanticity}) and probing methods (\citeauthor{gurnee2023findingneuronshaystackcase}, \citeyear{gurnee2023findingneuronshaystackcase}; \citeauthor{bills2023language}, \citeyear{bills2023language}) have successfully identified many such contextual features, developing a more mechanistic understanding of how these representations arise from the weights of a transformer remains a challenge.

In this paper, we take a small step towards improving said mechanistic understanding, using GPT-2 Small as a case study. Our contributions are:

\begin{itemize}
    \item \textbf{A decomposition of first-layer attention into positional and content-dependent components.} We approximate each head by a positional kernel-its baseline positional preference if tokens were identical-together with token-dependent adjustments.
    
    \item \textbf{Identification of stable heads.} We argue, and support empirically, that heads with broad positional kernels and weak content dependence often have softmax denominators that concentrate around their mean for the current distribution of tokens.
    
    \item \textbf{The contextual circuit approximation.} By sampling softmax denominators from a representative "calibration text", we can combine several such stable heads in GPT-2 Small, and construct what we call the contextual circuit: a linear summary of the surrounding text that approximates how these heads collectively influence downstream neurons.
    
    \item \textbf{Discovery of context-sensitive neurons.} As an application of this approximation, we identify first-layer neurons whose activations depend on interpretable properties of the surrounding text. For example, one neuron distinguishes between Commonwealth and American English. These neurons can be uncovered from model weights once a "calibration text" is chosen, without requiring large-scale corpus activations. And these neurons don't need to have activated on the calibration text.
\end{itemize}

In summary, our results suggest that certain stable first-layer heads in GPT2-Small can be approximated as constructing linear summaries of the surrounding text, which downstream MLP layers read from to detect high-level contextual properties of the text.
\end{section}

\section{Decomposition of First Layer Attention Patterns:}

\subsection{LayerNorm approximation:}
To decompose attention, we first want to write the keys of a head as a sum of positional and content-dependent components. However, LayerNorm complicates such a decomposition because the normalization step mixes position and content together in a nonlinear way. To handle this, we construct an approximation for the keys that cleanly separates position and content, and then show that the resulting attention patterns closely match the true attention patterns.

We take $\text{ln}_{\text{gain}}= 1$, $\text{ln}_\text{bias} = \mathbf{0}$, and $W_{\text{E}} \in \mathbb{R}^{d_{\text{voc}} \times d_{\text{model}}}$ and $W_{\text{pos}} \in \mathbb{R}^{n_{\text{ctx}} \times d_{\text{model }}}$ to be centered (rows sum to $0$). This is without loss of generality because we can fold in the gain and bias in to the surrounding weight matrices, and because LayerNorm projects away the row sums. When loading pretrained models with the TransformerLens library (\citeauthor{nanda2022transformerlens}, \citeyear{nanda2022transformerlens}) these transformations are performed automatically.

We fix $n$, the attending position, and we approximate the $i$th key of head $h$ on input $x$ as:
\begin{equation}\label{key_approx}
\begin{aligned}
\text{key}^{h}[i](x) &= \frac{\sqrt{d_{\text{model}}}(W_{\text{E}}[x_i]W^{h}_K+W_{\text{pos}}[i]W^{h}_K)}{|W_{\text{E}}[x_i]+W_{\text{pos}}[i]|} \\
&\approx \left(\frac{\sqrt{d_{\text{model}}}(W_{\text{E}}[x_i]W^{h}_K+W_{\text{pos}}[n]W^{h}_K)}{|W_{\text{E}}[x_i]+W_{\text{pos}}[n]|}\right) \\
&\quad + \left(\frac{\sqrt{d_{\text{model}}}(W_{\text{pos}}[i]W^{h}_K-W_{\text{pos}}[n]W^{h}_K)}{\sqrt{|W_{\text{pos}}[i]|^2+C^2}}\right) \\
&= E^{h}[n,x_i] + P^{h}[n,i]
\end{aligned}
\vphantom{\bigg\}}
\end{equation}
Where $x_i$ is the token in the $i$th position of the input sequence $x$, $1 \le i \le n$, and $C$ is the midpoint of the range of $|W_{\text{E}}[t]|$ across all tokens $t$.

We denote the attention pattern attending from position $n$ with our approximate keys substituted in as $\text{attn\_approx}^{h}[n,:]$, and the true attention pattern as $\text{attn}^{h}[n,:]$. To quantify the empirical quality of our approximation, we calculate the total variation distance $\text{TV}(\text{attn\_approx}^{h}[n,:],\text{attn}^{h}[n,:]) = \frac{\sum_{i=1}^{n} |\text{attn\_approx}^{h}[n,i]-\text{attn}^{h}[n,i]|}{2}$ across $100$ random input texts drawn from OpenWebText, across all $n$. 

\begin{figure}[!htp]
    \centering
    \includegraphics[width=0.75\linewidth]{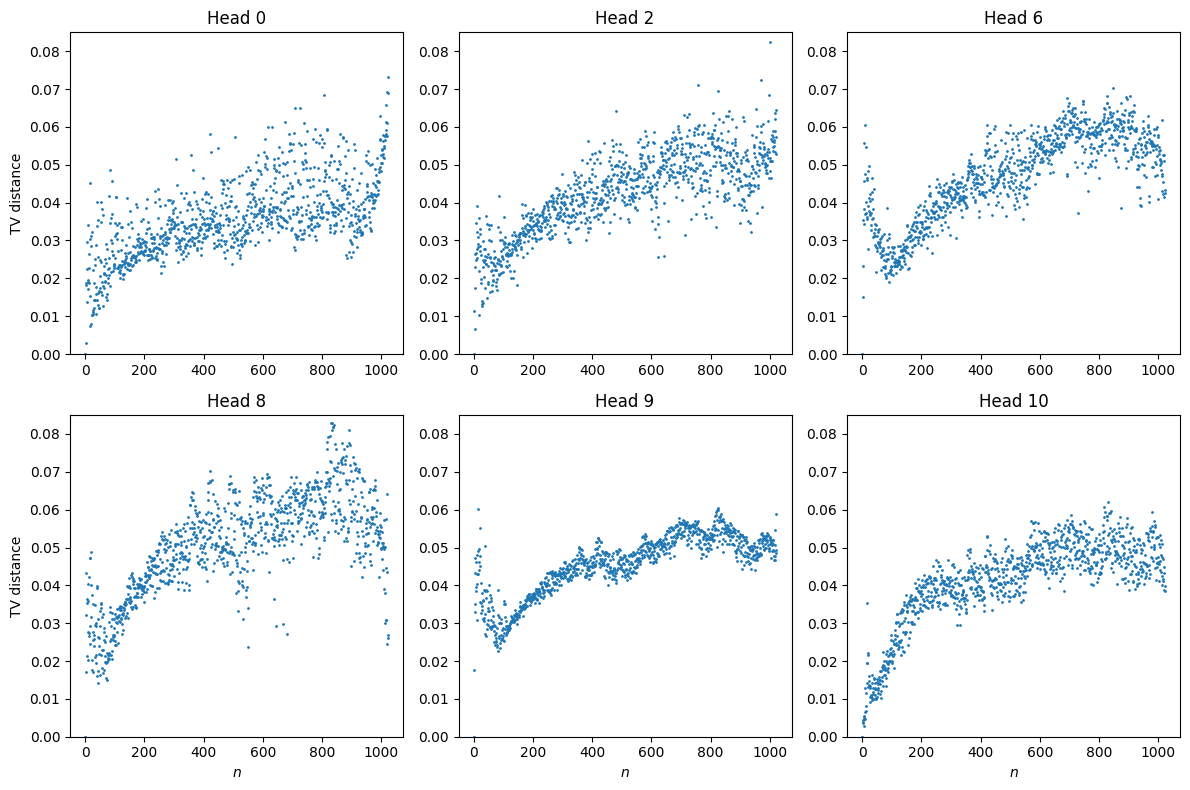}
    \caption{TV distance between true and reconstructed attention patterns across sequence positions for the $6$ attention heads analyzed in this work. Results are shown for a representative text from OpenWebText. The approximation maintains low TV (typically $\sim 0.05$) across all heads shown, with similar performance observed across all tested texts.}
    \label{fig:tv}
\end{figure}

We show the results for a representative text in \autoref{fig:tv}, for the $6$ heads we analyse in this work. The typical TV distance observed between the true attention pattern and our approximate attention pattern is $0.05$, which means that our approximation shifts $5\%$ of the total attention mass to different positions compared to the true attention pattern.

We discuss the reasoning behind our approximation of the keys (\autoref{key_approx}) in \autoref{app:layernorm_approximation}.

\FloatBarrier
\subsection{Positional kernels:}

Having approximated the keys as position + content components, we now use this approximation to decompose attention patterns.

We define the positional kernel at position $n$ as:
\begin{equation}
\begin{split}
\text{pos}^{h}_{n,i,x_n} 
&= \text{Softmax}\left(\frac{\text{query}^{h}[n](x)^{T}\text{P}^{h}[n,1:n+1]}{\sqrt{d_{\text{value}}}}\right)_{i}
\end{split}
\end{equation}

Where $\text{query}^{h}[n](x)$ is the query of head $h$ at the $n$th position of the input sequence, which is a function of just $n$ and $x_n$.

The positional kernel represents the intrinsic positional bias of an attention head. It is the underlying landscape upon which attention patterns are built. 

And we define  

\begin{equation}
{\text{content}}^{h}_{n,x_i,x_n} = e^{\frac{{\text{query}^{h}[n](x)^{T}E^{h}[n,x_i]}}{{\sqrt{d_{\text{value}}}}}}
\end{equation}

The attention pattern attending from position $n$ when we substitute our modified keys is:
\begin{equation}\label{eq:softmax_prob}
\begin{split} 
\text{attn\_approx}^{h}[n,i](x) &= \frac{e^{\frac{\text{query}^{h}[n](x)^T \text{key\_approx}^{h}[i](x)}{\sqrt{d_{\text{value}}}}}}{\sum_{j=1}^{n} e^{\frac{\text{query}^{h}[n](x)^T \text{key\_approx}^{h}[j](x)}{\sqrt{d_{\text{value}}}}}} \\
&= \frac{e^{\frac{\text{query}^{h}[n](x)^{T}\text{P}^{h}[n,i]}{\sqrt{d_{\text{value}}}}} \cdot e^{\frac{{\text{query}^{h}[n](x)^{T}E^{h}[n,x_i]}}{{\sqrt{d_{\text{value}}}}}}}{\sum_{j=1}^n e^{\frac{\text{query}^{h}[n](x)^{T}\text{P}^{h}[n,j]}{\sqrt{d_{\text{value}}}}} \cdot e^{\frac{{\text{query}^{h}[n](x)^{T}E^{h}[n,x_j]}}{{\sqrt{d_{\text{value}}}}}}}\\
&= \frac{\text{pos}^{h}_{n,i,x_n} \cdot {\text{content}^{h}}_{n,x_i,x_n}}{\sum_{j=1}^n \text{pos}^{h}_{n,j,x_n} \cdot {\text{content}^{h}}_{n,x_j,x_n}}
\end{split}
\end{equation}

From the second to the third line of \autoref{eq:softmax_prob} we divide the numerator and denominator by $\sum_{j=1}^{n} e^{\frac{\text{query}^{h}[n](x)^T\text{P}^{h}[n,j]}{\sqrt{d_{\text{value}}}}}$, normalizing the positional components so that they sum to $1$.

\begin{figure}[!htb]
   \centering
    \begin{tabular}{ccc}
     \captionsetup[subfigure]{labelformat=empty}
       \subfloat[Slowly decaying positional kernel]{\includegraphics[width=0.3\textwidth]{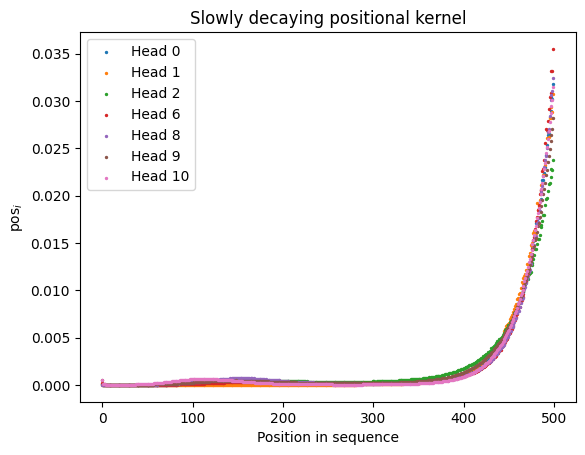}} &
         \captionsetup[subfigure]{labelformat=empty}
        \subfloat[Local positional kernel]{\includegraphics[width=0.3\textwidth]{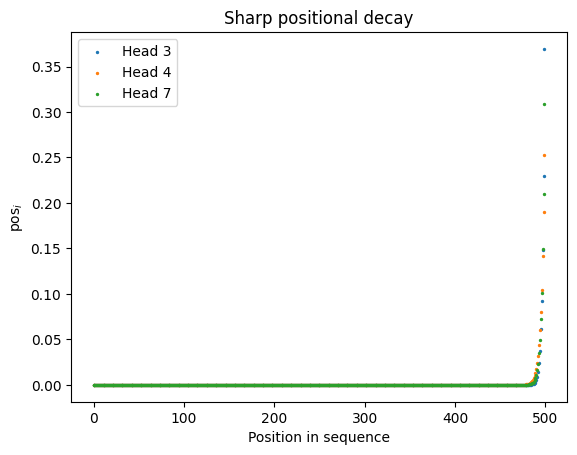}} &
        \captionsetup[subfigure]{labelformat=empty}
        \subfloat[Uniform positional kernel]{\includegraphics[width=0.3\textwidth]{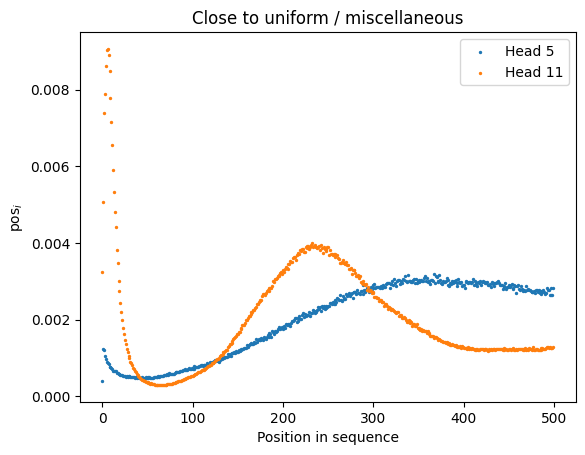}} \\[10pt] &
        \captionsetup[subfigure]{labelformat=empty}
        
    \end{tabular}
    \caption{Three types of positional kernels found across all heads in the first layer of GPT2-Small, at $n=500$, with $x_n$ corresponding to ' the'. Slowly decaying heads spread attention broadly, local heads focus on nearby positions, and uniform heads attend approximately equally across the sequence. }
    \label{fig:positional_}
\end{figure}
\FloatBarrier

We find three types of positional kernels in the first layer, shown in \autoref{fig:positional_}. We find that these kernels tend to be translation equivariant, and they depend weakly on the current token. The weak dependence on the current token justifies referring to the qualitative spread of the positional kernel of a head without reference to the current token. 
\section{Softmax denominator stability:}

Our approach is enabled by the fact that certain attention heads have stable softmax denominators across different texts within a given token distribution. When we fix the distribution that tokens are drawn from, heads with broad positional kernels and weak content dependence have softmax denominators that concentrate around their expected value for that distribution. This allows us to simplify the analysis of attention significantly by approximating the softmax denominators of these stable heads as constants.
\subsection{Stability of softmax denominators:}\label{stability}

We have the softmax probability formula \autoref{eq:softmax_prob}:

\begin{equation} \label{attnapprox}
\text{attn\_approx}^{h}[n,i](x) = \frac{\text{pos}^{h}_{n,i,x_n} \cdot {\text{content}^{h}}_{n,x_i,x_n}}{\sum_{j=1}^n \text{pos}^{h}_{n,j,x_n} \cdot {\text{content}^{h}}_{n,x_j,x_n}}
\end{equation}
 
To understand when the positionally-normalised denominator $\sum_{j=1}^n \text{pos}^{h}_{n,j,x_n} \cdot {\text{content}^{h}}_{n,x_j,x_n}$ is stable, we model the surrounding tokens as random variables. 

If we fix $x_n$, and model $x_i$ for $i<n$ as drawn i.i.d according to some underlying distribution, we can apply concentration inequalities such as Hoeffding/Chebyshev to bound the probability that the softmax denominator deviates too far from its mean, which we show in \autoref{app:theorems}. In both cases, concentration is governed by $\sum_{i=1}^{n-1} (\text{pos}^{h}_{n,i,x_n})^2$, together with a term measuring the content dependence of the attention scores. 

$\sum_{i=1}^{n-1} (\text{pos}^{h}_{n,i,x_n})^2$ will be small when the positional kernel is spread out. So we expect heads with wide positional kernels and attention scores depending weakly on content to have stable softmax denominators when $n$, $x_n$, and the underlying token distribution is fixed.

\subsection{Empirical stability of softmax denominators:}

\newlength{\imagewidth}
\setlength{\imagewidth}{0.3\textwidth}

We define:

\begin{equation}
\text{denom}_{h,n,t}(x) = \sum_{i=1}^{n} \text{pos}^{h}_{n,i,t} \cdot \text{content}^{h}_{n,y_i,t},
\end{equation}
where
\begin{equation*}
y_i =
\begin{cases}
x_i & \text{if } i < n, \\
t   & \text{if } i = n.
\end{cases}
\end{equation*}

\noindent
$\text{denom}_{h,n,t}(x)$ represents the value the denominator of \autoref{attnapprox} corresponding to head $h$ would take on at position $n$ if we substituted the token $t$ in at position $n$ of an input text $x$. 

The denominator of \autoref{attnapprox} for head $h$ at position $n$ of an input text $x$ is equal to $\text{denom}_{h,n,x_n}(x)$.

\begin{figure}[!htbp]
  \centering
  \begin{tabular}{@{} >{\centering\arraybackslash}m{\imagewidth} >{\centering\arraybackslash}m{\imagewidth} >{\centering\arraybackslash}m{\imagewidth} @{}}
    %
    \includegraphics[width=0.9\imagewidth]{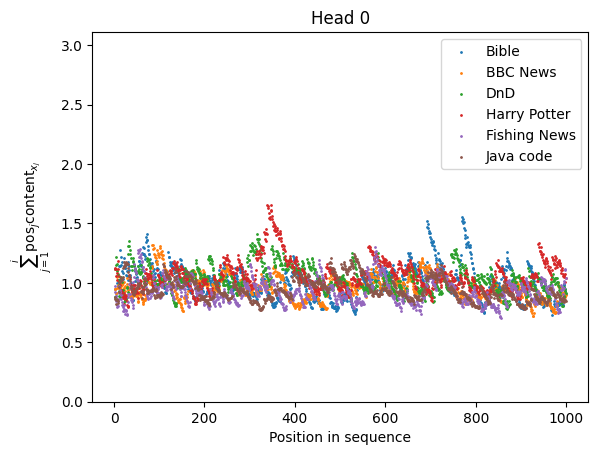}&
    \includegraphics[width=0.9\imagewidth]{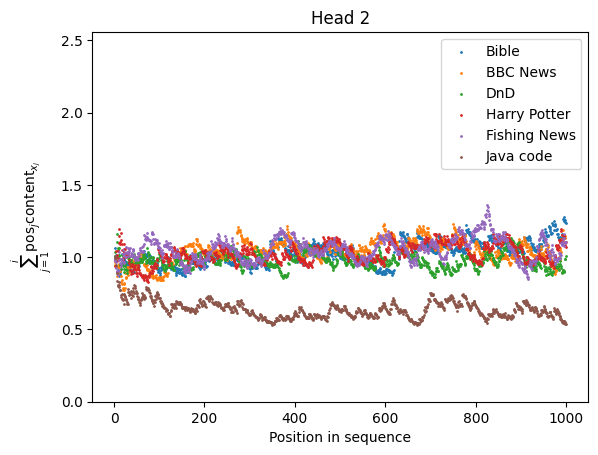} &
    \includegraphics[width=0.9\imagewidth]{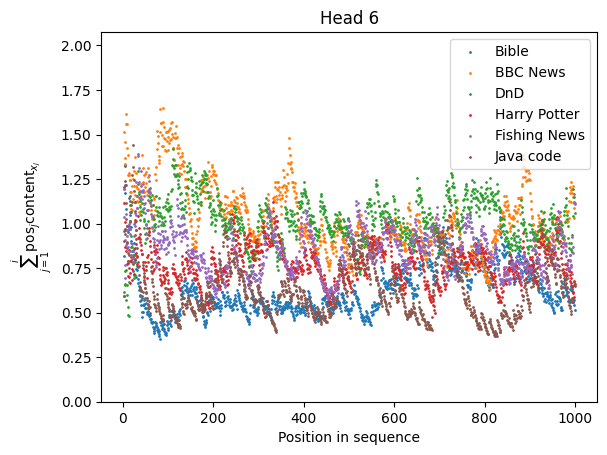} \\
    %
    \includegraphics[width=0.9\imagewidth]{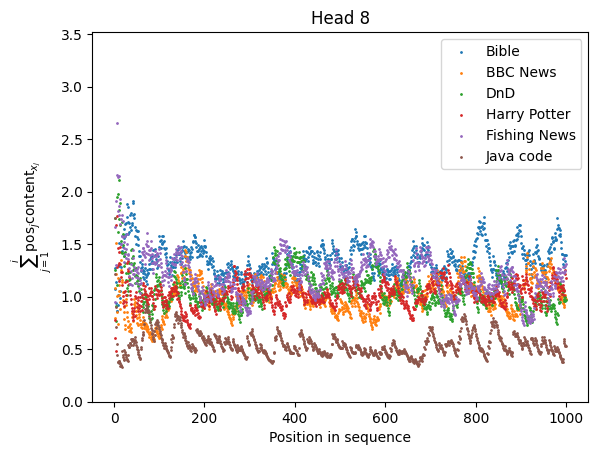} &
    \includegraphics[width=0.9\imagewidth]{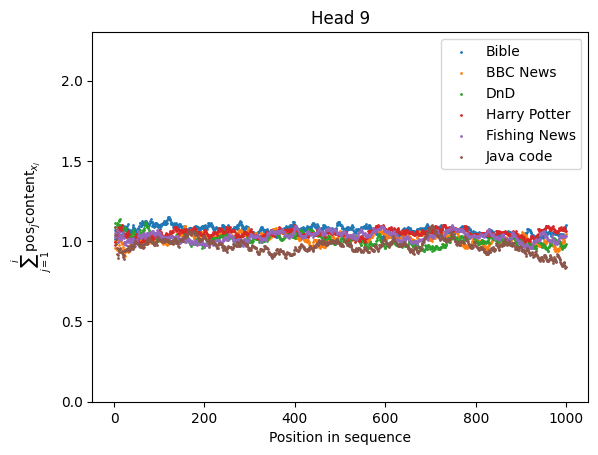}&
    \includegraphics[width=0.9\imagewidth]{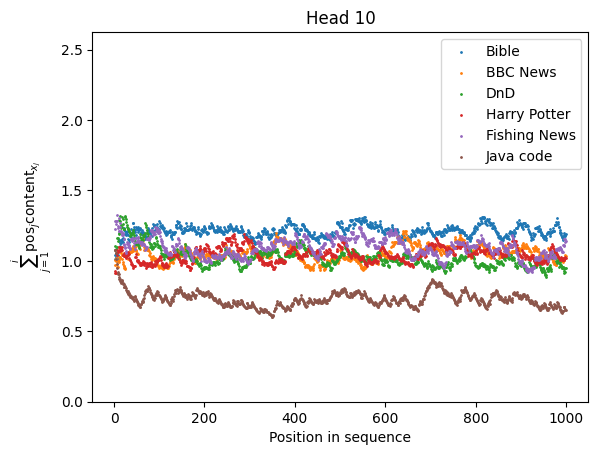} \\
  \end{tabular}
  \caption{$\frac{\text{denom}_{h,i,\text{' the'}}}{c_{h,i}}$ plotted against $i$ for a number of test texts, for 6 of the slowly decaying heads (see \autoref{fig:positional_}), where $c_{h,i}$ is an input-independent normalisation obtained by averaging $\text{denom}_{h,i,\text{' the'}}$ over $1000$ texts from OpenWebText. }
  \label{fig:denominators}
\end{figure}
\FloatBarrier

\autoref{fig:denominators} shows that for each fixed input text, the relative value of the positionally normalised denominator when we substitute the token ' the' in at the current position is moderately stable over the course of the input text. We show just a few input texts to make it easier to see the stability of each individual text. Texts with distinct token distributions can still concentrate around different expected values, however, as can be seen most clearly from the Java text (Programming) in head $2$.

\begin{figure}[htbp]
    \centering
    \includegraphics[width=0.7\textwidth]{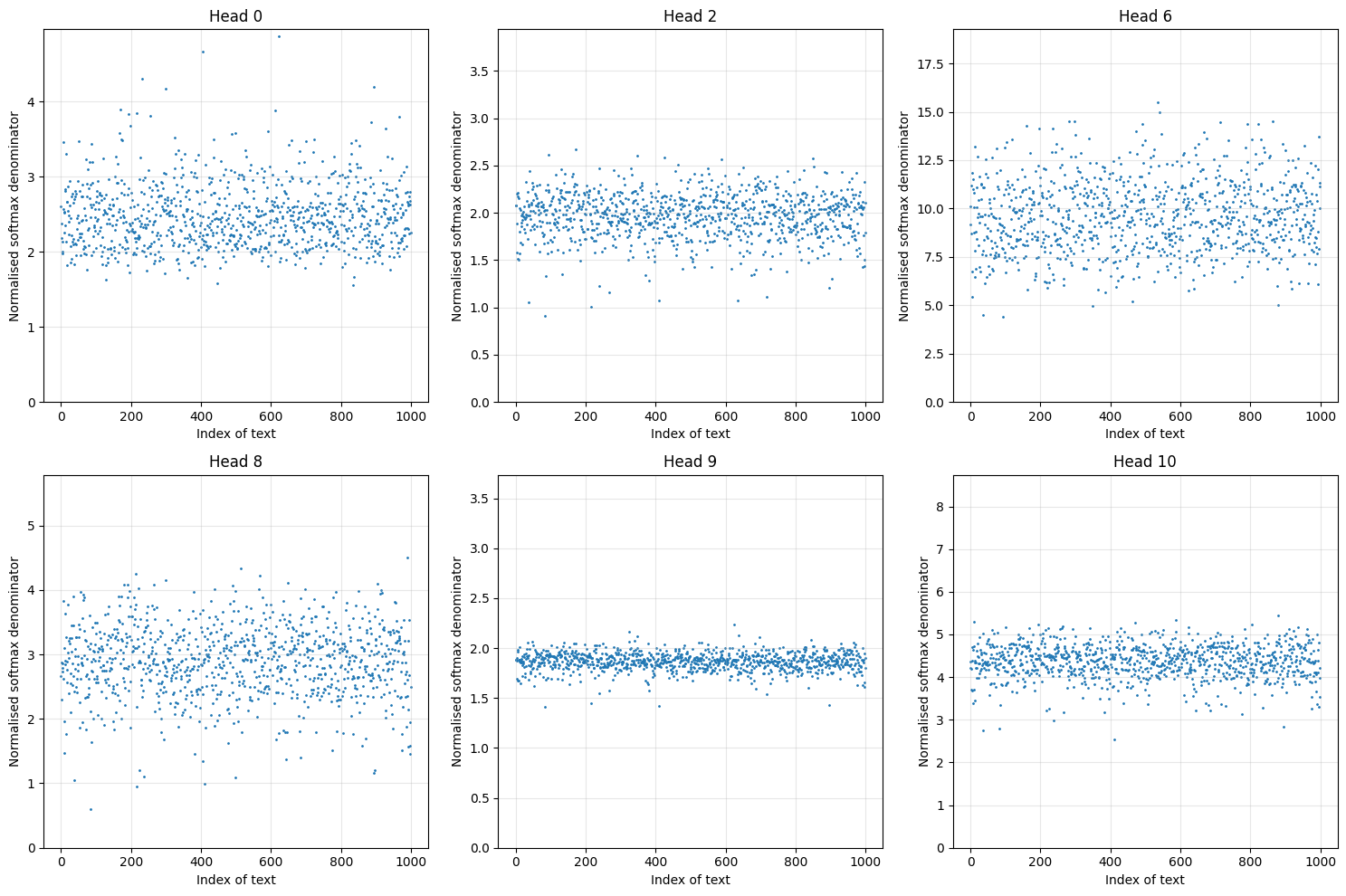}
    \caption{$\text{denom}_{h,\frac{n_{\text{ctx}}}{2},\text{' the'}}$ across $1000$ different input texts drawn from OpenWebText (texts indexed across the x-axis). Shown for $6$ of the slowly decaying first-layer heads.}
    \label{fig:prediction_scatter}
\end{figure}
\FloatBarrier

\autoref{fig:prediction_scatter} shows that $\text{denom}_{h,\frac{n_{\text{ctx}}}{2},\text{' the'}}$ concentrates to varying degrees around a central global value across different input texts drawn randomly from OpenWebText. We see similar concentration at different input positions.

Thus, when the attending token is ' the', not only do the denominators of heads concentrate around their expected value according to the underlying token distribution for the current text, but this expected value is quite consistent across different texts.

The heads with the largest variation are Heads $0$, $6$, and $8$. These heads each emphasize stop words to varying degrees, meaning that their  denominators vary based on local stop word density, which can vary a lot between different input texts. The outliers result from token distributions with atypical stop word densities. For example, programming texts result in outlier denominators for heads $6$ and $8$.

\section{Contextual circuit approximation:}

\subsection{OV circuit approximation:} \label{ov}
Fix a head $h$.  
The contribution of head $h$ to the pre-activation of MLP neuron $0.j$ at position $n$ is given by:
\begin{equation}
\mathbb{E}_{\text{attn}^{h}[n,:]}\!\left[
\frac{VO^{h}(i,x_i)\,\text{mlp}[:,j]}{\text{ln\_mlp}_{n}(x)}
\right],
\end{equation}
where $VO^{h}(i,x_i) = \frac{\sqrt{d_{\text{model}}}(W_{\text{E}}[x_i]+W_{\text{pos}}[i])}{|W_{\text{E}}[x_i]+W_{\text{pos}}[i]|} W^h_V W^h_O$ 
denotes the output vector (OV) contribution (\citeauthor{elhage2021mathematical}, \citeyear{elhage2021mathematical}) of token $x_i$ through head $h$, and $\text{ln\_mlp}_{n}(x)$  denotes the scale of the pre-MLP LayerNorm.

We can use our TV bound from \autoref{fig:tv} to control the error from using our approximate attention pattern:
\begin{align}
&\Bigg|
\mathbb{E}_{\text{attn}^{h}[n,:]}\!\left[
\frac{VO^{h}(i,x_i)\,\text{mlp}[:,j]}{\text{ln\_mlp}_{n}(x)}
\right]
-
\mathbb{E}_{\text{attn\_approx}^{h}[n,:]}\!\left[
\frac{VO^{h}(i,x_i)\,\text{mlp}[:,j]}{\text{ln\_mlp}_{n}(x)}
\right]
\Bigg| \nonumber \\[6pt]
&\quad\leq 
\frac{
\text{TV}(\text{attn}^{h}[n,:],\text{attn\_approx}^{h}[n,:])\,
\big(\sup_i VO^{h}(i,x_i)\,\text{mlp}[:,j] - \inf_i VO^{h}(i,x_i)\,\text{mlp}[:,j]\big)
}{\text{ln\_mlp}_{n}(x)} .
\end{align}

Empirically, we observe 
\[
\text{TV}(\text{attn}^{h}[n,:],\text{attn\_approx}^{h}[n,:]) \approx 0.05 ,
\]
as shown in \autoref{fig:tv}.  

Thus, the true OV circuit can be safely approximated by the OV circuit using 
$\text{attn\_approx}[n,:]$, provided that a $5\%$ reallocation of attention mass 
cannot drastically change the neuron’s output. 
Since the neurons we investigate aggregate contributions from many tokens in the sequence, 
this approximation should be reliable.  

\subsection{Contextual circuit:}

We now combine the outputs of 6 of the slowly decaying heads (see \autoref{fig:positional_}) to construct what we call the \textbf{contextual circuit} (\autoref{contextual_circuit}). This circuit approximates how these heads collectively contribute to neuron activations, enabling us to find context-sensitive neurons.

We approximate the combined contributions of the $\text{OV}$ circuits  of the slowly decaying heads ($\mathcal{H} = \lbrace{ 0,2,6,8,9,10 \rbrace}$) to MLP neuron $0.j$ on input $x$ by 

\begin{equation} \label{contextual_circuit}
\begin{aligned}
&\sum_{h \in \mathcal{H}} 
\mathbb{E}_{\text{attn}^{h}[n,:]} \left[
  \frac{VO^{h}(i,x_i)\,\text{mlp}[:,j]}{\text{ln\_mlp}_{n}(x)} \right] \\[4pt]
&\approx
\sum_{h \in \mathcal{H}}  \mathbb{E}_{\text{attn\_approx}^{h}[n,:]}
  \left[ \frac{\text{VO}^{h}(i,x_i)\,\text{mlp}[:,j]}{\text{ln\_mlp}_{n}(x)} \right]\\[4pt]
&= \sum_{h \in \mathcal{H}} 
\sum_{i=1}^{n} \left( \frac{\text{pos}^{h}_{n,i,x_n}\text{content}^{h}_{n,x_i,x_n}}{\text{denom}_{h,n,x_n}(x)} \right) \cdot \frac{\text{VO}^{h}(i,x_i)\text{mlp}[:,j]}{\text{ln\_mlp}_{n}(x)} \\[4pt]
&\approx \sum_{h \in \mathcal{H}} 
\sum_{i=1}^{n} \left( \frac{\text{pos}^{h}_{n,i,x_n}\text{content}^{h}_{n,x_i,x_n}}{\text{denom}_{h,n,x_n}(x)} \right) \cdot \frac{\text{VO}^{h}(n,x_i)\text{mlp}[:,j]}{\text{ln\_mlp}(n,x_n)} \\[4pt]
&\approx \sum_{i=1}^{n} \text{pos}_{n,i,x_n} 
\sum_{h \in \mathcal{H}} \frac{\text{content}^{h}_{n,x_i,x_n}  \cdot \text{VO}^{h}(n,x_i)\text{mlp}[:,j]}{\text{ln\_mlp}(n,x_n) \cdot \text{denom}_{h,n,x_n}(x)} \\[4pt]
&= \sum_{i=1}^{n} \text{pos}_{n,i,x_n} \,
\text{contribution}_{j}[n,x_n,x_i,[\text{denom}_{h,n,x_n}(x)]_{h \in \mathcal{H}}]
\end{aligned}
\vphantom{\bigg\}}
\end{equation}

From line $1$ to line $2$ of \autoref{contextual_circuit}, we use the approximation discussed in \ref{ov}.

From line $3$ to line $4$ we use the fact $\text{ln\_mlp}_{n}(x)$ doesn't vary significantly when $n$ and $x_n$ are fixed, so we approximate it as a function of $n$ and $x_n$. We also use the approximations $|W_{\text{E}}[x_i]+W_{\text{pos}}[i]| \approx |W_{\text{E}}[x_i]+W_{\text{pos}}[n]|$ (see \autoref{app:layernorm_approximation}), and $W_{\text{pos}}[i]\text{mlp}[:,j] \approx W_{\text{pos}}[n]\text{mlp}[:,j]$ for $i$ close to $n$ to approximate $\text{VO}^{h}(i,x_i)\text{mlp}[:,j]$ by $\text{VO}^{h}(n,x_i)\text{mlp}[:,j]$.

From line $4$ to line $5$ we approximate the head-specific positional kernels $\text{pos}^{h}_{n,i,x_n}$ by a single median positional kernel $\text{pos}_{n,i,x_n}$. We justify this approximation by the fact that the positional kernels of each of the slowly decaying heads are similar to each other, each taking an average over similar windows of tokens (See \autoref{fig:positional_}). This allows us to combine the $\text{OV}$ circuits (\citeauthor{elhage2021mathematical}, \citeyear{elhage2021mathematical}) of these heads together.

\section{Context-Sensitive neurons:} \label{context_neurons}

We can use the contextual circuit (\autoref{contextual_circuit}) to find neurons which respond to high-level properties of the surrounding text, which we call context-sensitive neurons. 

For the purposes of analysis we fix $x_n = \text{' the'}$ and fix $n=\frac{n_{\text{ctx}}}{2}$.

We approximate the final line of \autoref{contextual_circuit} by sampling $[\text{denom}_{h,\frac{n_{\text{ctx}}}{2},\text{' the'}}(y)]_{h \in \mathcal{H}}$ from a "calibration text" $y$ with an average stop word density, and substituting this in as an approximation for $[\text{denom}_{h,\frac{n_{\text{ctx}}}{2},\text{' the'}}(x)]_{h \in \mathcal{H}}=[\text{denom}_{h,n,x_n}(x)]_{h \in \mathcal{H}}$. We justify this substitution by the moderate concentration around a central value of $\text{denom}_{h,\frac{n_{\text{ctx}}}{2},\text{' the'}}$ shown in \autoref{fig:prediction_scatter} for each head $h \in \mathcal{H}$.

Thus, we further approximate the contribution of the slowly decaying heads to the pre-activation of MLP neuron ${0.j}$ by \begin{equation} \label{eq:neuron_approx} \sum_{i=1}^{n} \text{pos}_{n,i,x_n} \text{contribution}_{j}[n,x_n,x_i,[\text{denom}_{h,\frac{n_{\text{ctx}}}{2},\text{' the'}}(y)]_{h \in \mathcal{H}}] \end{equation} where $y$ is a "calibration text".

As $x_n$ and $n$ are fixed, and the "calibration text" $y$ is fixed across different input texts $x$, we simplify notation and write \autoref{eq:neuron_approx} as just \begin{equation}\sum_{i=1}^{n} \text{pos}_{i}\text{contribution}[j,x_i]\end{equation}. 

By looking for MLP neurons which are particularly sensitive to contributions from the slowly-decaying heads $\mathcal{H}$, we can find neurons which are sensitive to the surrounding context.

We define $\text{top}[j]$ of neuron ${0.j}$ as the $20$th largest value of $|\text{contribution}[j,t]|$ across all $t \in [0,d_{\text{voc}})$.

We choose a threshold $\theta>0$, where we reject neuron ${0.j}$ if $\text{top}[j] < \theta$, and otherwise accept.

We set this threshold $\theta$ to select at least $100$ neurons sensitive to the surrounding text. 

We find that the top and bottom contributions of the selected neurons tend to correspond to interpretable high-level properties of the surrounding text. We show a large number of these neurons in \autoref{app:contextual_neurons}.

The neurons we find respond to properties of the surrounding text that did not necessarily occur in the calibration text. 

Below we show an illustrative example:

\subsection{Commonwealth vs American English Neuron:}

Commonwealth vs American English (Neuron $704$):

Top $50$ token contributions:  
[' pract', ' foc', ' recogn', ' UK', ' British', ' London', ' £', ' Australia', ' Britain', 'isation', ' Australian', ' emphas', ' favour', ' Labour', ' centre', ' util', ' BBC', ' Scotland', ' behaviour', ' defence', ' Manchester', ' colour', ' €', ' labour', ' analys', ' programme', ' Liverpool', ' Wales', ' Sydney', ' Scottish', ' neighbour', ' favourite', ' organisation', ' keen', ' organis', ' offence', ' whilst', ' Melbourne', ' MPs', '£', ' honour', ' summar', ' organisations', ' Isis', ' travelling', ' Defence', ' licence', ' NHS', ' Dublin', ' armour']

Bottom $50$ token contributions: 
[' program', ' mom', ' favor', ' color', ' Center', ' center', ' defense', ' toward', ' Texas', ' favorite', ' organization', ' programs', ' behavior', ' analy', 'izes', 'izations', ' neighbor', ' labor', ' Color', 'avor', ' marijuana', ' organizations', ' Defense', ' license', ' attorney', ' neighborhood', ' realize', ' GOP', ' offense', ' realized', ' Seattle', ' honor', ' recognize', ' colors', ' gotten', ' folks', ' recognized', 'color', ' armor', ' organized', ' §', ' Oregon', 'sylvania', ' baseball', ' transportation', ' Iowa', ' downtown', ' flavor', '.--', ' accomplish']

This neuron classifies text as using Commonwealth English or American English, activating on Commonwealth English. This neuron wasn’t activated on the calibration text -meaning we truly discovered it from weights, not from dataset artifacts.

\section{Accuracy of contextual circuit approximation:}
\label{app:approximation_accuracy}

To test the accuracy of \autoref{eq:neuron_approx}, we first fix a baseline text $y$ to use as our "calibration text". We don't use this baseline text in any of our evaluations.

For each neuron of interest we pick two texts, one of which is a control text, which we don't believe the neuron activates for, and the second of which is a sample of text which we believe the neuron should activate on.

At each position, when evaluating the contextual contribution from that position, we substitute the token ' the' in at that position, while leaving the remaining sequence unchanged. 

To obtain the 'true contextual contribution', we run the model on our modified sequence (a different modified sequence at each position), and take the combined OV outputs from Heads (0, 2, 6, 8, 9, 10), and multiply this combined output by the input to the neuron of interest. 
\begin{figure}[!htb]
\centering
\begin{subfigure}[t]{0.45\textwidth}
  \centering
  \includegraphics[width=\linewidth]{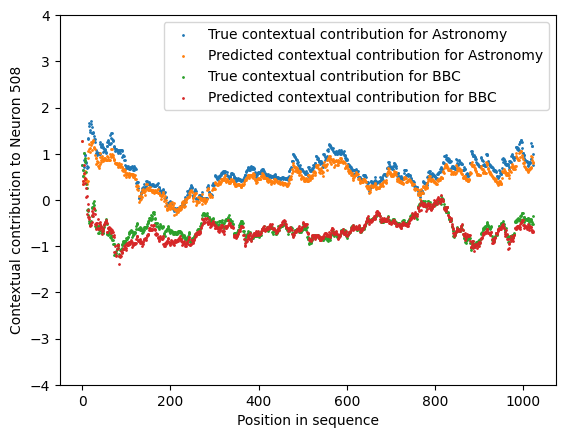}
  \caption{Neuron $0.508$ acts on astronomy related texts. Approximation works well because both texts have typical stop word densities}
  \label{fig:neuron508}
\end{subfigure}
\hfill
\begin{subfigure}[t]{0.45\textwidth}
  \centering
  \includegraphics[width=\linewidth]{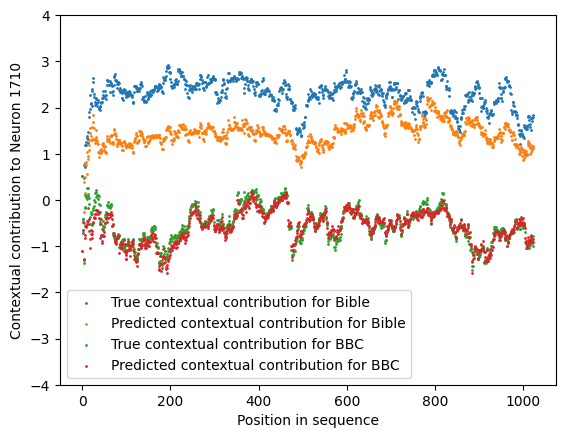}
  \caption{Neuron $0.1710$ activates on religious texts. The Bible has an abnormal stop word density, and our contextual circuit systematically underestimates the contextual contribution}
  \label{fig:neuron1710}
\end{subfigure}

\vspace{1em}

\begin{subfigure}[t]{0.45\textwidth}
  \centering
  \includegraphics[width=\linewidth]{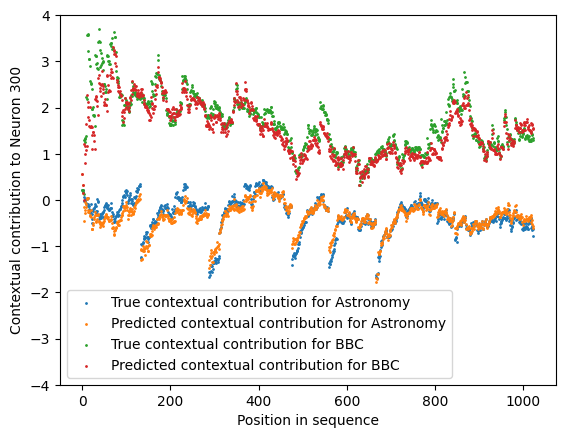}
  \caption{Neuron $0.300$ activates on British English texts}
  \label{fig:neuron300}
\end{subfigure}
\hfill
\begin{subfigure}[t]{0.45\textwidth}
  \centering
  \includegraphics[width=\linewidth]{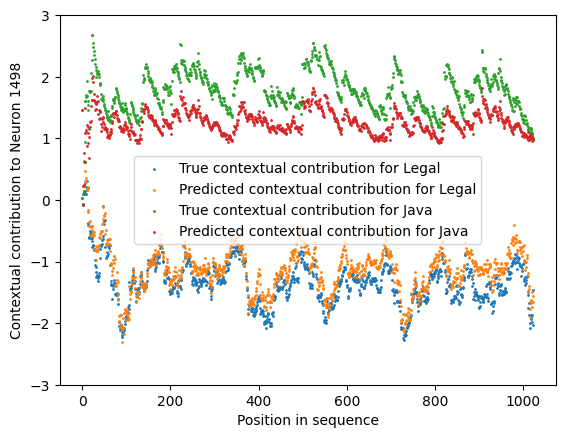}
  \caption{Neuron $0.1498$ activates on code.}
  \label{fig:neuron1498}
\end{subfigure}

\caption{Visualizations of context-sensitive neurons discovered using the contextual circuit}
\label{fig:contextual_neuron_grid}
\end{figure}

\FloatBarrier

While our contextual circuit approximation isn't perfect, it is usually strongly positively correlated with the true contributions, with a median correlation coefficient across context-sensitive neurons when running on $1000$ random texts from OpenWebText of $r = 0.9485$, and a median $\text{FVU}$ of $0.1426$. This suggests that we should be able to rely on it for finding the qualitative role of context-sensitive neurons. We show a few representative examples of the typical contextual circuit approximate contribution versus true contribution in \autoref{app:neuron_truth}.

Given more information about the specific domain, we could make \autoref{eq:neuron_approx} more precise. For example, if we were interested in investigating programming texts, which have distinct normalised denominators from prose texts (See \autoref{fig:denominators}), then we could use a programming text as our "calibration text" $y$.

Similarly, future work could refine the approximation given in \autoref{eq:neuron_approx} by clustering texts $x$ based on the values of $[\text{denom}_{h,n,t}(x)]_{h \in \mathcal{H}}$, and using different sets of approximate denominators for each cluster.

\begin{section}{Related Work}

\subsection*{Mechanistic Interpretability:}

Our work builds upon and partially extends the Mathematical Framework for Transformer Circuits (\citeauthor{elhage2021mathematical}, \citeyear{elhage2021mathematical}), which established foundational principles for decomposing transformer computation into interpretable components.

Sparse Autoencoders (SAEs) (\citeauthor{bricken2023monosemanticity}, \citeyear{bricken2023monosemanticity}) are neural networks trained to reconstruct transformer activations using a sparse linear combination of learned feature directions. They have been incredibly successful at the unsupervised discovery of interpretable features corresponding to sophisticated properties of the text. 

However, SAEs treat the sublayers of models as black boxes and so provide little mechanistic insight into how the features they discover are constructed. Our work complements SAEs by explaining mechanistically how first-layer attention heads construct linear summaries of the surrounding text which the MLP layer can use to distinguish contexts from each other.

\subsection*{Context-Sensitive neurons:}
Previous work (\citeauthor{gurnee2023findingneuronshaystackcase}, \citeyear{gurnee2023findingneuronshaystackcase}) showed the existence of context-sensitive neurons in a wide variety of language models by using sparse probing techniques. Similarly, (\citeauthor{bills2023language}, \citeyear{bills2023language}) used automated methods to find patterns in empirical neuron activations.  However, both had to run the model over a text corpus to collect empirical neuron activations. 

The contextual circuit allows us to uncover first-layer context-sensitive neurons using just a single calibration text. This leaves our analysis less vulnerable to interpretability illusions (\citeauthor{bolukbasi2021interpretabilityillusionbert},\citeyear{bolukbasi2021interpretabilityillusionbert}), because we can be confident that our results are properties of the model, not of the dataset.

\subsection*{Attention Pattern Analysis:}

(\citeauthor{clark-etal-2019-bert},\citeyear{clark-etal-2019-bert}) discussed attention heads in BERT that construct BoW representations, i.e: construct linear summaries of the sequence.

(\citeauthor{singh2023analyzingtransformerdynamicsmovement}, \citeyear{singh2023analyzingtransformerdynamicsmovement}) discussed attention heads whose role is the positionally-weighted aggregation of representations, and discussed the role positional bias (which they call the positional kernel, whereas we use positional kernel to refer to the softmax of the positional bias) plays in the mechanics of these heads.

Our work complements these analyses by providing a principled decomposition that helps to formalize the attention patterns found in these studies, and exploit them to find downstream interpretable components.

\end{section}

\section{Conclusion:}

We decompose attention into positional and content-dependent components, revealing how first-layer heads with broad positional kernels create stable linear summaries of the surrounding text. This enables the systematic discovery of context-sensitive neurons without a large text corpus, showing how detailed mathematical analysis can advance mechanistic interpretability.

While our analysis focuses on the first layer of models with additive positional embeddings, the approach illustrates a broader methodology: using simple models of the input (i.i.d in this case) to motivate approximations of model behavior, then validating these approximations empirically and using them to uncover interpretable structure. This general strategy may prove useful for understanding other components and architectures.

Future work could extend this decomposition to deeper layers and explore how models with alternative positional schemes, such as RoPE models, perform similar tasks.

\section{Acknowledgements:}

We are grateful to Jason Gross, Euan Ong, and Thomas Read for their valuable feedback and helpful discussions during the development of this work.

\bibliography{refs}

\appendix

\section{Limitations}
\label{app:limitations}

We limit ourselves to discussing models with additive positional embeddings, T5, and ALiBi. This is because RoPE models are significantly more complex to analyze (see \autoref{app:alternative_position} for a brief discussion of why).

We only thoroughly investigated the contextual circuit of GPT2-Small in the main text.

We only looked at a small number of sample texts for our plot of the positionally-normalised softmax denominators. This is partially to help visualize the concentration of measure, but also because it is computationally expensive to compute the denominators at each position for each sample text.

\section{Compute required and assets used:} \label{app:compute_and_assets}

For all experiments performed, we used a single T4 GPU (16GB) provided by Google Colab. All of the experiments in the paper run in under $1$ minute. In total, the compute time required for all the experiments in the paper is under $10$ minutes.

We used the OpenWebText dataset (\citeauthor{Gokaslan2019OpenWeb}, \citeyear{Gokaslan2019OpenWeb}) (CC0 License) for measuring the empirical TV distance between our approximate attention pattern and true attention pattern.

We ran experiments on GPT2 (\citeauthor{radford2019language}, \citeyear{radford2019language}, License: \citeauthor{gpt2License}), Pythia (\citeauthor{biderman2023pythiasuiteanalyzinglarge}, \citeyear{biderman2023pythiasuiteanalyzinglarge}, Apache License 2.0), and TinyStories (\citeauthor{eldan2023tinystoriessmalllanguagemodels}, \citeyear{eldan2023tinystoriessmalllanguagemodels}, MIT License). We used DeepSeek-R1 (\citeauthor{deepseekai2025deepseekr1incentivizingreasoningcapability}, \citeyear{deepseekai2025deepseekr1incentivizingreasoningcapability}, MIT License) for automatically classifying contextual neurons.

We used the following Python packages: (\citeyear{nanda2022transformerlens}, \citeauthor{nanda2022transformerlens}), (\citeyear{plotly}, \citeauthor{plotly}), (\citeyear{paszke2017automatic}, \citeauthor{paszke2017automatic}), (\citeyear{hunter2007matplotlib}, \citeauthor{hunter2007matplotlib}), (\citeauthor{wolf-etal-2020-transformers}, \citeyear{wolf-etal-2020-transformers}), (\citeauthor{lhoest-etal-2021-datasets}, \citeyear{lhoest-etal-2021-datasets}).

\section{Alternative positional schemes:} \label{app:alternative_position}

There are many different positional schemes used for different LLMs, but the four most common alternatives to additive positional embeddings are:

\begin{itemize}
\item \textbf{NoPE}: Has no positional embedding, can be viewed as a special case of additive positional embeddings.

\item \textbf{T5} (\citeauthor{raffel2023exploringlimitstransferlearning}, \citeyear{raffel2023exploringlimitstransferlearning}):
Adds a learnt relative bias to the pre-softmax attention scores based on the relative distance between the positions of the current token and the token being attended to.
\item \textbf{ALiBi} (\citeauthor{press2022trainshorttestlong},\citeyear{press2022trainshorttestlong}): Similar to T5, adds a linear bias to the pre-softmax attention scores proportional to the relative distance between the current token and the token being attended to. The slope of this bias is hard-coded, not learnt.

\item \textbf{RoPE} (\citeauthor{su2023roformerenhancedtransformerrotary}, \citeyear{su2023roformerenhancedtransformerrotary}): Applies a complex rotation of $\theta = 10000^{\frac{-2i}{\text{rot}_{\text{dim}}}}$ to the $2i$th and $(2i+1)$th dimensions of the query and key vectors, up to the $\text{rot}_\text{dim}$th index. $\text{rot}_{\text{dim}}$ varies depending on implementation: some implementations take $\text{rot}_{\text{dim}}$ = $d_{\text{head}}$, where $d_{\text{head}}$ is the dimension of the query and key vectors. The Pythia (\citeauthor{biderman2023pythiasuiteanalyzinglarge}, \citeyear{biderman2023pythiasuiteanalyzinglarge}) line of models uses $\text{rot}_{\text{dim}} = \frac{d_{\text{head}}}{4}$ for efficiency purposes. Note some implementations of RoPE, including Pythia, rotate the $i$th and $i+\frac{\text{rot}_{\text{dim}}}{2}$th dimensions instead. This is mathematically equivalent to standard RoPE, but can be confusing.
\end{itemize}
T5 and AliBi are pretty straightforward to handle under this framework. We simply softmax the relative bias terms to define the positional kernel. As long as the positional kernel defined satisfies \autoref{attnapprox}, then the rest of our analysis applies. T5 and ALiBi are in fact far easier to manage than additive postional embeddings, because there is no need to approximate the effect of LayerNorm.

RoPE is more difficult to work with. When we were dealing with NoPE, the positional kernel was determined by the query vector. For RoPE models, even if the query vector is fixed, the ratio between attention paid to a token $t_1$, and the attention paid to a token $t_2$, might vary based on position. For example, RoPE heads might emphasize $t_1$ over $t_2$ on even key positions, and emphasize $t_2$ over $t_1$ on odd key positions. 

This would violate a frequent implicit assumption in interpretability work that the function of a head can be described without reference to position. For example, a head might act like a duplicate token head on even key positions, and act like an induction head on odd key positions. After a preliminary investigation on Pythia models (\citeauthor{biderman2023pythiasuiteanalyzinglarge}, \citeyear{biderman2023pythiasuiteanalyzinglarge}), it's not clear that something of this form doesn't happen, and so analysis of RoPE models is complicated.

\section{Heuristic motivation behind LayerNorm approximation:} \label{app:layernorm_approximation}

To handle LayerNorm, we fix $n$, the attending position, and we approximate the $i$th key of head $h$ on input $x$ as:
\begin{equation}
\begin{aligned}
\text{key}[i](x) &= \frac{\sqrt{d_{\text{model}}}(W_{\text{E}}[x_i]W_K+W_{\text{pos}}[i]W_K)}{|W_{\text{E}}[x_i]+W_{\text{pos}}[i]|} \\
&\approx \left(\frac{\sqrt{d_{\text{model}}}(W_{\text{E}}[x_i]W_K+W_{\text{pos}}[n]W_K)}{|W_{\text{E}}[x_i]+W_{\text{pos}}[n]|}\right) \\
&\quad + \left(\frac{\sqrt{d_{\text{model}}}(W_{\text{pos}}[i]W_K-W_{\text{pos}}[n]W_K)}{\sqrt{|W_{\text{pos}}[i]|^2+C^2}}\right) \\
&= E[n,x_i] + P[n,i]
\end{aligned}
\vphantom{\bigg\}}
\end{equation}

In this decomposition we make use of two different approximations for $|W_E[x_i]+W_{\text{pos}}[i]|$. 

For the $\text{E}[n,x_i]$ term, we use $|W_E[x_i]+W_{\text{pos}}[i]| \approx |W_E[x_i]+W_{\text{pos}}[n]|$.

 For the $\text{P}[n,i]$ term, we use the approximation $|W_E[x_i]+W_{\text{pos}}[i]| \approx \sqrt{|W_{\text{pos}}[i]|^2+C^2}$, where $C$ is the midpoint of the range of $|W_E[x_i]|$.

\begin{figure}[h]
    \centering
    \includegraphics[width=0.8\textwidth]{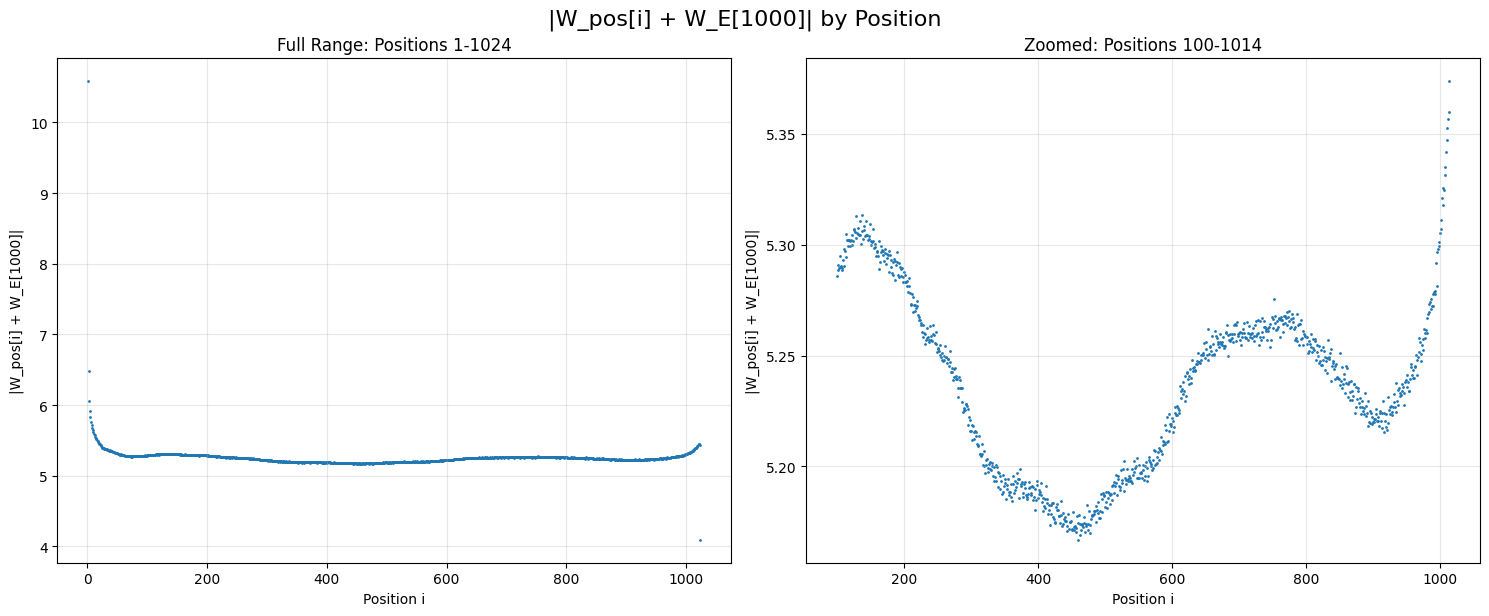}
    \caption{Scatter plots showing $|W_{\text{pos}}[i] + W_E[1000]|$ across different positions $i$. The left panel displays the full range of positions (1-1024), while the right panel provides a zoomed view of positions 100-1014}
    \label{fig:position_norms}
\end{figure}

$|W_E[x_i]+W_{\text{pos}}[i]| \approx |W_E[x_i]+W_{\text{pos}}[n]|$ is effective when $|i-n|$ is small, because $|W_E[t]+W_{\text{pos}}[i]|$ varies slowly as a function of $i$, when $t$ is a fixed token. As a representative example, see \autoref{fig:position_norms}, showing the slow variation when $t=1000$.

The geometric intuition for the slow variation in $|W_E[t] + W_{\text{pos}}[i]|$ is that firstly $W_{\text{E}}$ is approximately orthogonal to $W_{\text{pos}}$ across all positions (cosine similarity uniformly below $0.15$). Secondly $|W_{\text{pos}}[i]|$ varies slowly.

In our approximation corresponding to the $\text{P}[n,i]$ term we use the fact that $W_{\text{pos}}$ and $W_E$ are approximately orthogonal to estimate $|W_E[x_i]+W_{\text{pos}}[i]| \approx \sqrt{|W_E[x_i]|^2+|W_{\text{pos}}[i]|^2} \approx \sqrt{C^2+|W_{\text{pos}}[i]|^2}$.

The approximation for $|W_E[x_i]+W_{\text{pos}}[i]|$  corresponding to $\text{P}[n,i]$ is significantly worse than the approximation corresponding to $\text{E}[n,i]$ when $i$ is near $n$. This is because in practice there is a $\pm 20\%$ variation in $|W_E[t]+W_{\text{pos}}[i]|$ as $t$ varies and $i$ is fixed. Whereas for $i$ near $n$ our $\text{E}[n,i]$ approximation is near perfect.

Most heads in the first layer of GPT2-Small pay the majority of their attention locally (i.e: previous $50$ tokens). Therefore, to get a low reconstruction error on the downstream approximate attention pattern, we focus on getting a low relative error in exponential attention score when $i$ is near $n$. This will lead to a low relative error in attention pattern where most of the attention is paid.

To achieve a low relative error in exponential attention score for $i$ near $n$, we need a low absolute error in attention score for $i$ near $n$. 

Because the $\text{P}[n,i]$ approximation for the denominator is poor compared to the $\text{E}[n,i]$ approximation for $i$ near $n$, we reduce the absolute size of the attention scores corresponding to the $\text{P}[n,i]$ by adding and subtracting $W_{\text{pos}}[n]W_K$ from the numerator of these terms. This reduces the absolute error in attention scores for $i$ near $n$ significantly.

\section{Concentration of measure for first layer attention heads:} \label{app:theorems}

If $x_n$ is fixed, and $x_i \sim Z$ for $i<n$ are drawn independently and identically distributed, we are interested in the concentration of $\sum_{i=1}^{n} \text{pos}_{n,i,x_n}\text{content}_{n,x_n,x_i}$, which we shorten to $\sum_{i=1}^{n} \text{pos}_{i} \text{content}_{x_i}$ by taking $n$ and $x_n$ to be fixed. As $x_n$ is fixed, concentration is governed by $\sum_{i=1}^{n-1} \text{pos}_{i}\text{content}_{x_i}$. Viewing $\text{content}_{x_i}$ as i.i.d random variables, we can apply standard concentration inequalities. For instance, if $\text{content}_{x_i}$ consistently lies in the interval $[A,B]$, we can apply Hoeffding to obtain $\mathbf{P}_{Z}(|\sum_{i=1}^{n-1} \text{pos}_{i}\text{content}_{x_i} -(1-\text{pos}_n)\mathbb{E}_{Z}[\text{content}_{x_1}]| \ge t) \le e^{-\frac{2t^2}{(\sum_{i=1}^{n-1} \text{pos}_{i}^2)(B-A)^2}}$. We can potentially refine this by assuming better bounds on $||Z||_{\psi_2}$, because the bounds on content can be quite loose.

We also have Chebyshev, which can be useful given just a variance bound on content, $\mathbf{P}_{Z}(|\sum_{i=1}^{n-1} \text{pos}_{i}\text{content}_{x_i} -(1-\text{pos}_n)\mathbb{E}_{Z}[\text{content}_{x_1}]| \ge t) \le \frac{(\sum_{i=1}^{n-1} \text{pos}_{i}^2)\text{Var}_{Z}(\text{content}_{x_1})}{t^2}$

In both cases, the more spread out the positional kernel is, the smaller $\sum_{i=1}^{n-1}\text{pos}_{i}^2$ is, and the tighter the concentration.

\section{Contextual neurons:}\label{app:contextual_neurons}

To find contextual neurons, we filtered for neurons whose $20$th largest token contribution exceeded $5.0$. We found that the majority of neurons above this threshold correspond to particular topics, but sometimes the top token contributions just consist of "glitch tokens". 

Several neurons seem to encode problematic social biases and stereotypes. We have deliberately excluded these neurons, as they contain token contributions strongly associated with discriminatory content related to gender, race, and other protected characteristics. The token contributions in these cases predominantly correspond to content that would be offensive or perpetuate harmful stereotypes.

To automatically label contextual neurons, we printed a list of the top 50 token contributions for each neuron, and prompted DeepSeek R1 with:  """For each of these neurons, rewrite in the form Neuron (neuron) (What you think the topic the neuron relates to): Top token contributions: (Top token contributions) Bottom token contributions :(Bottom token contributions). These token contributions are performing Naive Bayes classification of the input, but sometimes the bottom contributions are generic, so omit them if they are. If any neurons seem inappropriate or offensive to you, don't include them. If a neuron has an abundance of glitch tokens, i.e: Tokens that don't seem like common words or words which start without a space, then call it a glitch neuron and don't include it. Be strict in rejecting neurons, but not too strict. Include all 50 top token contributions. If a few tokens seem out of place, list them under leniencies."""

Neuron 28 (Programming Syntax):
Top token contributions: [’ int', '//', ' mem', ' char', ' const', ');', ' ;', ' code', ' struct', 'itiz', ' Get', '->', 'Get', '++', 'rency', ' fif', ' );', ' boost', 'irty', ' ==', 'itial', ' blocks', 'struct', ' implementation', ' void', 'dev', ' define', ' prev', ' static', ' addresses', ' */', ' Py', ' flags', 'const', 'Start', ' template', ' \&\&', ' algorithm', 'mem', 'Make', ' instruction', ' hash', 'Block', ' AF', 'adeon', 'atile', ' Consider', 'static', 'Message', 'define’]
Explanation: Captures programming syntax (C/C++/Python), including pointers, structs, and functions. Overlaps with Neurons 141/184 but emphasizes low-level code.
Leniency: Includes generic terms like "blocks."

Neuron 65 (Sports Fandom):
Top token contributions: [’And', ' players', ' fans', ' Get', ' guy', 'aturday', ' guys', 'Get', ' coach', ' (@', ' football', ' Mike', ' fuck', ' kid', ' Joe', ' Thanks', ' Thank', ' legit', ' crazy', ' shit', ' quarterback', ' Come', ' gonna', ' Keep', ' roster', ' fucking', ' Yeah', ' throwing', 'Thanks', 'outube', '????', ' tonight', ' dump', ' ESPN', 'Thank', ' rush', ' Stadium', ' Reddit', 'Hey', ' tweeted', 'Yeah', ' coaches', ' ridiculous', ' hopefully', ' Tradable', '?!', 'Never', ' rookie', ' midfield', 'bie’]
Explanation: Centers on sports (football, ESPN), fan interactions, and informal language ("fuck," "shit").
Leniency: Tolerates profanity and social media references.

Neuron 90 (RPG Mechanics):
Top token contributions: [’ att', ' fam', ' adv', ' feat', ' wond', 'uration', ' spell', ' familiar', 'Su', ' incor', ' DC', ' AC', ' creature', ' bonus', ' inqu', ' creatures', ' gains', ' saving', 'vision', ' spells', ' magical', '.:', 'description', 'aneous', ' Dak', ' expend', ' Medium', ' rolls', ' wounds', ' DM', ' wield', ' terrain', ' martial', ' Craft', 'ignment', ' Dex', ' melee', ' Choose', ' Starting', 'caster', ' eld', ' cant', ' favored', ' chooses', 'casting', 'folk', ' Knowledge', ' Cha', ' suffers', ' Whenever’]
Explanation: Targets RPG elements: stats ("Dex," "Cha"), spells, creatures, and game mechanics.
Leniency: Overlaps with general gaming terms.

Neuron 97 (Biological Sciences):
Top token contributions: [’).', ' hist', ' activ', ' et', ' stri', 'rote', ' cell', ' hom', ' compared', ' prec', ' conserv', ' chem', ' cells', ' dim', ' stress', ' quant', ' suggests', ' vir', ' enh', 'duc', ' Ca', ' observed', ' flu', ' phen', ' cere', ' ).', ' DNA', ' protein', ' infect', 'lation', ' neuro', 'cific', ' constit', ' sequence', 'hesis', ' genetic', 'igen', ' whereas', ' suggesting', ' gene', ' intr', 'roph', ' mouse', ' HIV', ' flav', 'acteria', ' oxy', ' mice', ' tissue', ' chronic’]
Explanation: Focuses on biology, genetics, and medical research (DNA, HIV, mice).
Leniency: Includes broad terms like "stress."

Neuron 141 (Java/API Programming):
Top token contributions: [’ import', ' const', ' conc', ' type', ' typ', ' decl', ' code', ' struct', ' //', ' implement', ' impl', ' twe', '++', ' vari', ' library', ' exception', ' assert', ' interface', ' simpl', 'struct', ' Java', ' implementation', ' API', ' define', ' inject', ' stack', ' static', ' implemented', 'ypes', 'const', ' inher', ' String', ' serial', 'chron', 'import', 'java', 'main', ' abstract', ' throws', 'static', ' libraries', 'ync', 'itialized', 'bage', ' compiled', ' debug', ' generic', ' declaration', ' std', ' extends’]
Explanation: Java-centric programming, APIs, and code structure.
Leniency: Overlaps with Neurons 28/184.

Neuron 144 (Social Sciences/Philosophy):
Top token contributions: [’ prov', ' econom', ' upon', ' necess', ' profess', ' whose', ' contr', ' society', ' histor', ' thus', ' whatever', ' consequ', ' occup', ' sust', ' appreci', ' univers', ' Thus', ' philosoph', ' frust', ' bes', ' fundamental', ' intellect', ' embr', ' privile', ' displ', ' constit', ' subsid', ' Such', ' poverty', ' pursu', ' discipl', ' comprehens', ' narrative', ' burd', ' ordinary', ' enorm', ' perpet', ' intellectual', ' scholars', ' whereas', 'ultane', ' contrad', ' collective', ' wrest', ' obsc', ' intim', ' ought', ' ancest', ' Moreover', ' imper’]
Explanation: Abstract concepts in philosophy, economics, and academia.
Leniency: Broad thematic scope.

Neuron 184 (Programming/Data Structures):
Top token contributions: [’ int', ' import', 'orld', ' object', ' init', ' code', ' struct', '()', ' default', ' config', ' commit', ' `', ' id', ' string', ' output', ' objects', ' );', ' library', ' sched', 'lines', ' ==', 'itial', ' arguments', ' array', ' Java', ' generated', 'etch', ' pref', 'var', ' implementation', ' API', ' variable', 'dev', 'lib', 'List', ' define', ' module', ' directory', 'string', ' */', 'ypes', ' variables', ' kernel', 'const', 'String', ' inher', ' String', ' template', 'config', ' Python’]
Explanation: Programming/data structures (arrays, modules) across languages (Java, Python).
Leniency: Overlaps with Neurons 28/141.

Neuron 198 (Law Enforcement):
Top token contributions: [’ recogn', ' hon', ' vehicle', ' contribut', ' acqu', ' suspect', 'resp', ' drugs', ' arrested', 'pret', ' marijuana', ' wearing', ' accompl', ' violent', ' armed', 'illance', ' warrant', ' racial', ' anticip', 'Police', ' substance', 'imore', ' welfare', ' Officer', ' drunk', ' custody', ' Drug', ' meth', ' residence', ' bail', 'Offic', ' excessive', 'ective', ' firearm', ' bedroom', 'erguson', ' felony', ' screaming', ' arrests', ' disturbing', ' mentally', 'Lear', ' Veh', ' cocaine', 'ctuary', ' testified', ' burg', ' convenience', ' substances', ' intersection’]
Explanation: Crime-related terms: arrests, drugs, warrants.
Leniency: Includes location names ("erguson").

Neuron 201 (Environmental NGOs):
Top token contributions: [’ UN', ' conservation', ' WHO', ' NGO', 'ustainable', ' sustainability', ' philanthrop', 'Scientists', ' NGOs', ' Rockefeller', ' biodiversity', 'Climate', ' TED', ' environmentalists', 'forestation', 'WHO', ' Greenpeace', ' WWF', ' Sustainable', ' GMOs', ' deforestation', ' UNESCO', ' plantations', ' UNHCR’]
Explanation: Environmental organizations (WWF, UNESCO) and sustainability.
Leniency: Narrow focus with minimal overlap.

Neuron 246 (3D Graphics and Game Development):
Top token contributions: [’ object’, ’ anim’, ’ twe’, ’ scen’, ’ objects’, ’ transform’, ’ coord’, ’ drag’, ’ frust’, ’ixel’, ’ poly’, ’ pref’, ’ drawing’, ’ render’, ’atern’, ’ entity’, ’ vert’, ’ rend’, ’Color’, ’ grid’, ’ spawn’, ’ animation’, ’anim’, ’raphics’, ’ Anim’, ’ texture’, ’ gui’, ’ tutorial’, ’ algorithm’, ’ physics’, ’ lighting’, ’ entities’, ’ float’, ’amples’, ’ rotation’, ’Ang’, ’ gravity’, ’itialized’, ’ brush’, ’Player’, ’ debug’, ’ invisible’, ’ rendering’, ’clip’, ’ animated’, ’Mult’, ’ shapes’, ’ Draw’, ’ terrain’, ’ometry’]
Explanation: Relates to animation, rendering, and game design. Highly specific.

Neuron 254 (Crime and Public Events):
Top token contributions: [’ told’, ’ according’, ’ described’, ’ club’, ’ scene’, ’ Saturday’, ’ suspect’, ’ appeared’, ’ Police’, ’ football’, ’According’, ’orry’, ’Mr’, ’ arrived’, ’ wearing’, ’ armed’, ’ reportedly’, ’illance’, ’ allegedly’, ’About’, ’ Avenue’, ’Police’, ’ attended’, ’ baseball’, ’ Manchester’, ’ protesters’, ’ basketball’, ’ knife’, ’ concert’, ’ shirt’, ’ drove’, ’ approached’, ’ tweeted’, ’ surrounded’, ’ fled’, ’ suspects’, ’ dressed’, ’bourne’, ’Offic’, ’ grabbed’, ’ overnight’, ’Earlier’, ’ singer’, ’Online’, ’ collapsed’, ’ screaming’, ’ detained’, ’ crashed’, ’ attacker’, ’ teenager’]
Explanation: Covers crime in public/sports settings. Leniency: Accepts "Manchester" (location).

Neuron 257 (Game Mechanics and RPG Elements):
Top token contributions: [’ consum’, ’ attached’, ’Set’, ’Item’, ’rior’, ’rient’, ’ Exec’, ’cest’, ’================================’, ’gage’, ’ equipped’, ’ Marketable’, ’Quest’, ’ Error’, ’urrency’, ’[/’, ’ triggered’, ’ Thread’, ’ Property’, ’deals’, ’ Rare’, ’Custom’, ’Incre’, ’ Warning’, ’ alias’, ’lander’, ’ PART’, ’ trait’, ’Card’, ’Common’, ’Effect’, ’amina’, ’Season’, ’acters’, ’oiler’, ’ cooldown’, ’ Entity’, ’Blood’, ’Large’, ’Damage’, ’ETF’, ’Ability’, ’ Healing’, ’Items’, ’ initialized’, ’Standard’, ’"]=>’, ’items’, ’ Trigger’, ’inventory’]
Explanation: Focuses on in-game systems (quests, items, abilities). Coherent.

Neuron (300 /704) (UK Culture and Governance):
Top token contributions: [’ pract’, ’ foc’, ’ recogn’, ’ UK’, ’ British’, ’ London’, ’ £’, ’ Australia’, ’ Britain’, ’isation’, ’ Australian’, ’ emphas’, ’ favour’, ’ Labour’, ’ centre’, ’ util’, ’ BBC’, ’ Scotland’, ’ behaviour’, ’ defence’, ’ Manchester’, ’ colour’, ’ €’, ’ labour’, ’ analys’, ’ programme’, ’ Liverpool’, ’ Wales’, ’ Sydney’, ’ Scottish’, ’ neighbour’, ’ favourite’, ’ organisation’, ’ keen’, ’ organis’, ’ offence’, ’ whilst’, ’ Melbourne’, ’ MPs’, ’£’, ’ honour’, ’ summar’, ’ organisations’, ’ Isis’, ’ travelling’, ’ Defence’, ’ licence’, ’ NHS’, ’ Dublin’, ’ armour’]
Explanation: Targets UK-specific cultural, political, and institutional terms. Leniency: Tolerates "Isis" (historical context).

Neuron 308 (U.S. Politics and Government):
Top token contributions: [’ Trump’, ’ president’, ’ President’, ’ Obama’, ’ countries’, ’ China’, ’ billion’, ’ economic’, ’ Congress’, ’ Russia’, ’ Republican’, ’ Senate’, ’ jobs’, ’ schools’, ’ debate’, ’ nuclear’, ’ decades’, ’ Republicans’, ’ policies’, ’ Democrats’, ’ reform’, ’ Secretary’, ’ debt’, ’ Britain’, ’ legislation’, ’ businesses’, ’Trump’, ’ immigration’, ’rastructure’, ’ Econom’, ’ GOP’, ’ governments’, ’ nations’, ’ argued’, ’plom’, ’ politicians’, ’ immigrants’, ’ Nations’, ’ diplom’, ’ poverty’, ’ Putin’, ’ Barack’, ’ Senator’, ’ fiscal’, ’ historic’, ’ negotiations’, ’ Mayor’, ’ lawmakers’, ’President’, ’Sen’]
Explanation: Political figures, policies, and international relations. Overlaps with Neuron 2806 but distinct.

Neuron 319 (Programming and Computer Science):
Top token contributions: [’ diff’, ’ spec’, ’ incre’, ’ using’, ’ hist’, ’ occ’, ’ object’, ’ code’, ’ comput’, ’ Gener’, ’ files’, ’ memory’, ’irection’, ’ twe’, ’ default’, ’ tick’, ’ config’, ’ string’, ’ random’, ’ graph’, ’++’, ’ output’, ’ input’, ’ bug’, ’ functions’, ’ vari’, ’ objects’, ’sembly’, ’ sched’, ’ optim’, ’lines’, ’itial’, ’ overw’, ’ interface’, ’ixel’, ’ array’, ’ Java’, ’ synt’, ’ython’, ’ generated’, ’ pref’, ’ generate’, ’ implementation’, ’ API’, ’ variable’, ’data’, ’List’, ’ filter’, ’ define’, ’ keys’]
Explanation: Technical terms in software development. Coherent.

Neuron 333 (Entertainment and Pop Culture):
Top token contributions: [’ music’, ’ fans’, ’ movie’, ’ album’, ’ movies’, ’ songs’, ’ films’, ’ Music’, ’ studio’, ’isodes’, ’ Hollywood’, ’ Disney’, ’ debut’, ’ featuring’, ’ Nintendo’, ’ guitar’, ’ musical’, ’ comedy’, ’ logo’, ’ Studio’, ’ smartphone’, ’ designer’, ’ anime’, ’ Netflix’, ’ trailer’, ’ LED’, ’ Blu’, ’ merch’, ’ DVD’, ’ comics’, ’ cinem’, ’ Studios’, ’ singer’, ’ actress’, ’ jacket’, ’ lyrics’, ’ costume’, ’ bikes’, ’ albums’, ’ packaging’, ’ filming’, ’ DLC’, ’ promotional’, ’ilogy’, ’ premiere’, ’ indie’, ’ textures’, ’ hasht’, ’ starring’, ’ studios’]
Explanation: Blends music, film, and gaming culture. Leniency: Accepts "textures" (game dev).

Neuron 337 (Scientific Research and Academia):
Top token contributions: [’ scientists’, ’ stim’, ’ NASA’, ’ theories’, ’ documentary’, ’ invented’, ’ biology’, ’ teaches’, ’ encourages’, ’ promotes’, ’ drawings’, ’ psychologist’, ’ Scientists’, ’Abstract’, ’ doi’, ’ospace’, ’ explores’, ’Researchers’, ’ stimulate’, ’ Participants’, ’ Researchers’, ’ depicting’, ’ depicts’, ’Scientists’, ’NASA’, ’ coined’, ’ fascination’, ’ microscope’, ’ mascot’, ’ contestants’, ’ physicists’, ’ telescopes’, ’ documentaries’, ’ biologists’, ’ Genetics’, ’ portrays’, ’ theorist’, ’ Pixar’, ’ chimpanzees’, ’ Nasa’, ’ EEG’, ’ frontman’, ’ stimulates’]
Explanation: Focuses on research methodologies and scientific outreach.

Neuron 380 (Religion and Christianity):
Top token contributions: [’ supp’, ’ contin’, ’iqu’, ’ God’, ’ according’, ’ ben’, ’ Christ’, ’lished’, ’ signific’, ’ shall’, ’ Israel’, ’ pred’, ’imum’, ’ Paul’, ’ histor’, ’ beg’, ’ Lord’, ’ circum’, ’ earth’, ’ Phil’, ’ Ep’, ’ According’, ’ Jud’, ’iliar’, ’ Book’, ’ Mary’, ’ sch’, ’ Peter’, ’ sust’, ’ ath’, ’ Jesus’, ’ univers’, ’ ("’, ’ fol’, ’ incor’, ’ forth’, ’ Son’, ’ Egypt’, ’Act’, ’ Jews’, ’ Orig’, ’ pra’, ’ Joseph’, ’ Adam’, ’ Holy’, ’John’, ’ teaching’, ’ sin’, ’ wine’, ’ Nic’]
Explanation: Biblical figures, theology, and religious texts.

Neuron 433 (Biological Research):
Top token contributions: [’ et’, ’ Austral’, ’ researchers’, ’ scientists’, ’ flu’, ’iments’, ’ evolution’, ’ protein’, ’ colleagues’, ’ sequence’, ’Ear’, ’ genetic’, ’ Using’, ’ populations’, ’ studied’, ’ gene’, ’ Credit’, ’acteria’, ’ mice’, ’ genes’, ’ predicted’, ’ Figure’, ’ bacteria’, ’ electron’, ’ Fig’, ’ abund’, ’ fossil’, ’ infer’, ’ hypothesis’, ’ brains’, ’Fig’, ’’, ’ proteins’, ’ glucose’, ’ sequences’, ’ evolutionary’, ’ neurons’, ’ molecules’, ’ nucle’, ’ ancestors’, ’ molecular’, ’ genome’, ’avier’, ’ cortex’, ’ organisms’, ’ viruses’, ’ genetically’, ’ogene’, ’ microsc’, ’inosaur’]
Explanation: Centers on genetics, microbiology, and evolutionary biology. Leniency: Accepts "Credit" as academic context.

Neuron 453 (Instructional Examples and Documentation):
Top token contributions: [’ example’, ’ following’, ’nown’, ’ Here’, ’ Let’, ’Here’, ’ According’, ’According’, ’ listed’, ’ follows’, ’Let’, ’ examples’, ’ writes’, ’During’, ’ typical’, ’ paragraph’, ’ phr’, ’Before’, ’ quote’, ’ Below’, ’Each’, ’ quoted’, ’Table’, ’ documented’, ’ Consider’, ’Using’, ’ quotes’, ’ Following’, ’ essay’, ’Following’, ’Among’, ’written’, ’ syntax’, ’ summar’, ’ defines’, ’Example’, ’ diagram’, ’ outlined’, ’ Notice’, ’ Example’, ’ preceding’, ’ illustrated’, ’ illustrate’, ’ discusses’, ’endix’, ’Consider’, ’iciency’, ’Dear’, ’ Fortunately’, ’example’]
Explanation: Highlights instructional language, examples, and technical documentation. Coherent.

Neuron 508 (Astronomy and Space Science):
Top token contributions: [’ universe’, ’ Universe’, ’ abund’, ’ observations’, ’ spacecraft’, ’upiter’, ’ Jupiter’, ’ Physics’, ’ Saturn’, ’ cosmic’, ’ telescope’, ’ Einstein’, ’ methane’, ’rared’, ’ Neptune’, ’velength’, ’ asteroid’, ’ equations’, ’ galaxies’, ’ wavelength’, ’ gravitational’, ’NASA’, ’ comet’, ’ Pluto’, ’ Martian’, ’ crater’, ’ astronomers’, ’ Voyager’, ’ Milky’, ’ rover’, ’ Comet’, ’ uncertainties’, ’ Telescope’, ’nova’, ’ Horizons’, ’ Curiosity’, ’ Hubble’, ’ Atmospheric’, ’ telescopes’, ’ helium’, ’ neb’, ’ propell’, ’ Chandra’, ’ Nebula’, ’ Centauri’, ’ Nasa’, ’ astroph’, ’ relat’, ’ outburst’]
Explanation: Focuses on space exploration, celestial bodies, and astrophysics. Coherent.

Neuron 510 (Wildlife Conservation and Ecology):
Top token contributions: [’ nest’, ’ Wildlife’, ’ breeding’, ’angered’, ’ habitat’, ’ deer’, ’ litter’, ’ hatch’, ’ claws’, ’ Species’, ’ turtle’, ’ Tasman’, ’ enclosure’, ’ mating’, ’ aquarium’, ’ habitats’, ’ sightings’, ’ Spawn’, ’ carc’, ’ pengu’, ’ spawning’, ’ Jurassic’, ’ sighting’, ’ Squirrel’, ’ nests’, ’ foliage’, ’ dorsal’, ’ nesting’, ’ hiber’, ’ Beetle’, ’ hatched’, ’ lair’]
Explanation: Relates to animal behavior, habitats, and conservation efforts. Coherent.

Neuron 560 (Human Rights and Governance):
Top token contributions: [’ commun’, ’ rights’, ’uclear’, ’ Minister’, ’ vill’, ’ UN’, ’ authorities’, ’ presidential’, ’ crisis’, ’ reform’, ’liament’, ’ arrested’, ’ decre’, ’ minister’, ’ accused’, ’ guarant’, ’itutional’, ’ emergency’, ’ruption’, ’icial’, ’ institutions’, ’ crimes’, ’ lawyer’, ’ armed’, ’ elections’, ’ regime’, ’ terrorist’, ’ newspaper’, ’ politicians’, ’ regional’, ’ activists’, ’ jur’, ’ mainly’, ’ protests’, ’ parliament’, ’ congress’, ’ aimed’, ’ journalists’, ’ corruption’, ’ ethnic’, ’ Ministry’, ’ protesters’, ’ coalition’, ’ concrete’, ’ journalist’, ’ Article’, ’ lawmakers’, ’President’, ’ amid’, ’ civilians’]
Explanation: Covers governance, legal frameworks, and human rights issues. Leniency: Accepts "concrete" as metaphorical.

Neuron 574 (Music Production and Performance):
Top token contributions: [’ feat’, ’ song’, ’ twe’, ’ album’, ’ grab’, ’ listen’, ’ recorded’, ’ tun’, ’ CD’, ’ songs’, ’ Live’, ’ Music’, ’ tur’, ’ studio’, ’ downt’, ’ recording’, ’ listening’, ’ concert’, ’ trem’, ’ guitar’, ’ musical’, ’ Syd’, ’ solo’, ’ vocal’, ’ merch’, ’ Beat’, ’ bass’, ’ drum’, ’ singing’, ’ Tun’, ’ tune’, ’ singer’, ’ overd’, ’ folk’, ’ Mix’, ’ Hi’, ’ vib’, ’ listened’, ’ Around’, ’ jam’, ’ mixing’, ’ orche’, ’ beats’, ’ Feel’, ’Live’, ’mix’, ’ Bass’, ’ Listen’, ’ jazz’, ’ listeners’]
Explanation: Emphasizes music creation, performance, and technical production. Coherent.

Neuron 641 (Medical and Health Research):
Top token contributions: [’ study’, ’ et’, ’ studies’, ’ patients’, ’ cells’, ’ researchers’, ’ diet’, ’ findings’, ’ flu’, ’ cere’, ’ protein’, ’ neuro’, ’ genetic’, ’‐’, ’ studied’, ’ gene’, ’ Studies’, ’ HIV’, ’ mice’, ’ tissue’, ’ chronic’, ’ dose’, ’ genes’, ’ Drug’, ’Figure’, ’ bacteria’, ’esity’, ’ lung’, ’ Study’, ’ epid’, ’ analyses’, ’ rats’, ’ Neuro’, ’ treatments’, ’ hypothesis’, ’ gluc’, ’ vitamin’, ’ concentrations’, ’ metabol’, ’ dietary’, ’ Cancer’, ’ proteins’, ’ glucose’, ’ plasma’, ’ correlation’, ’ neurons’, ’ receptor’, ’ medications’, ’ placebo’, ’ nucle’]
Explanation: Targets medical studies, disease mechanisms, and pharmacology. Leniency: Accepts "Drug" as context.

Neuron 701 (Programming and Computational Systems):
Top token contributions: [’ prog’, ’ const’, ’ function’, ’ init’, ’pecially’, ’ code’, ’ method’, ’ rad’, ’ comput’, ’ struct’, ’()’, ’onent’, ’ implement’, ’ server’, ’ config’, ’ipment’, ’ random’, ’ cells’, ’++’, ’ output’, ’ input’, ’ functions’, ’ Each’, ’acement’, ’ );’, ’ sample’, ’iquid’, ’ wire’, ’Data’, ’iments’, ’ components’, ’ database’, ’itial’, ’ amounts’, ’ixel’, ’ array’, ’aturally’, ’ poly’, ’ component’, ’ protein’, ’ processing’, ’ layer’, ’ generate’, ’ util’, ’ implementation’, ’ API’, ’data’, ’ernel’, ’ void’]
Explanation: Technical terms in software engineering and system design. Coherent.

Neuron 716 (Geopolitics and U.S. Foreign Policy):
Top token contributions: [’ milit’, ’ America’, ’ Obama’, ’ Clinton’, ’ Republican’, ’ Islam’, ’ Iraq’, ’ Iran’, ’ peace’, ’ Democratic’, ’ Syria’, ’ Republicans’, ’ policies’, ’ politics’, ’ Democrats’, ’ Hillary’, ’ democr’, ’ Britain’, ’ Bush’, ’ Sanders’, ’Trump’, ’ immigration’, ’ GOP’, ’ governments’, ’ elected’, ’ elections’, ’ Muslims’, ’ Labour’, ’ liberal’, ’ Saudi’, ’ politicians’, ’ ISIS’, ’ Polit’, ’ democracy’, ’ Putin’, ’ protests’, ’ Barack’, ’ establishment’, ’ protesters’, ’ statist’, ’ Bernie’, ’ democratic’, ’ corrupt’, ’ndum’, ’ NATO’, ’ Brexit’, ’ referendum’, ’ propaganda’, ’ politically’, ’ unions’]
Explanation: Focuses on international relations, U.S. politics, and conflicts. Coherent.

Neuron 727 (Elections and Political Polling):
Top token contributions: [’ poll’, ’ Sanders’, ’ Romney’, ’ polls’, ’ndum’, ’Bet’, ’tarian’, ’ conced’, ’ delegates’, ’ debates’, ’ recount’, ’Clinton’, ’ DNC’, ’ BJP’, ’ backers’, ’ organizers’, ’ incumb’, ’ campaigning’, ’ contests’, ’ delegate’, ’ Kasich’, ’ commentators’, ’oters’, ’Sanders’, ’tarians’, ’leground’, ’Republicans’, ’ bets’, ’ contenders’, ’ Gingrich’, ’Democrats’, ’ Voting’, ’ Libertarian’, ’poll’, ’ pundits’, ’Bernie’, ’ concede’, ’Democratic’, ’ Santorum’, ’ spoiler’, ’ Voters’, ’idates’, ’legates’, ’ RNC’, ’ polled’, ’Poll’, ’ rematch’, ’ POLITICO’, ’ recounts’, ’ Supporters’]
Explanation: Centers on electoral processes, polling, and party dynamics. Leniency: Accepts "Bet" as slang.

Neuron 734 (Media Reporting and Public Statements):
Top token contributions: [’ told’, ’ according’, ".'", ’ toward’, ",'", ’ ahead’, ’ incred’, ’ explained’, ’ scientists’, ’ spokesman’, ’makers’, ’ emphas’, ’ argued’, ’ reporters’, ’ogether’, ’ caption’, "?'", ’pson’, ’ replied’, ’ chairman’, ’etts’, ’ referring’, ’izabeth’, ’ lawmakers’, ’ amid’, ’ Stadium’, ’ quoted’, ’ insisted’, ’ Jean’, ’vernment’, ’ Brexit’, ’.'”’, ’enny’, ’ spokeswoman’, ’ criticized’, ’ recalled’, ’ midfield’, ’ migrants’, ’ ō’, ".''", "!'", ’ Commons’, ’Speaking’, ’ assured’, ’ ministers’, ’ Singh’, ’ nodded’, ’ MPs’, ’ stressed’, ’ meanwhile’]
Explanation: Captures media narratives, official statements, and public discourse. Leniency: Tolerates fragmented tokens like "ō."

Neuron 760 (Informal Language and Expressions):
Top token contributions: [’ly’, "'t", ’ful’, ’ You’, "'ll", ’ seem’, "'d", ’ering’, ’ccess’, ’You’, ’ comes’, ’ With’, ’ due’, ’ seems’, ’):’, ’ gets’, ’ looks’, ’With’, ’ etc’, ’ nice’, ’ Once’, ’ starts’, ’ooth’, ’rency’, ’cel’, ’ anyway’, ’ shit’, ’ weird’, ’ Still’, ’ likes’, ’ tries’, ’ annoy’, ’ Deep’, ’ Having’, ’ngth’, ’full’, ’ Something’, ’bage’, ’ useless’, ’ annoying’, ’ Besides’, ’ Perfect’, ’ nicely’, ’ WILL’, ’ kinda’, ’ Probably’, ’too’, ’ept’, ’ sigh’, ’ noticeable’]
Explanation: Reflects conversational tone, slang, and informal phrasing. Leniency: Accepts profanity ("shit") as colloquial.

Neuron 775 (Hardware and Systems Programming):
Top token contributions: [’ mov’, ’ init’, ’ struct’, ’ sequ’, ’ fif’, ’ sched’, ’ pin’, ’ isol’, ’dev’, ’ acceler’, ’ USB’, ’ */’, ’ kernel’, ’ bytes’, ’ analys’, ’ assemb’, ’ interrupt’, ’ serial’, ’abling’, ’ buffer’, ’ disk’, ’ LED’, ’ trace’, ’ volt’, ’itialized’, ’ header’, ’ debug’, ’trl’, ’ voltage’, ’ esp’, ’ switching’, ’ discont’, ’FIG’, ’plug’, ’ pointer’, ’ socket’, ’ PCI’, ’ byte’, ’ switches’, ’ peripher’, ’ capac’, ’ datas’, ’ wires’, ’ packet’, ’ FIG’, ’ interfaces’, ’ pins’, ’ traces’, ’ ip’, ’ Unic’]
Explanation: Technical terms in hardware design and low-level programming. Coherent.

Neuron 786 (Narrative and Conversational Elements):
Top token contributions: [’ said’, "'ve", ’ told’, ’You’, ’ went’, ’And’, ’ someone’, ’That’, ’They’, ’ecause’, ’She’, ’ guy’, ’ couldn’, ’ Jack’, ’ nice’, ’ guys’, ’Don’, ’ kids’, ’ breat’, ’ kid’, ’ baby’, ’ feels’, ’Just’, ’Well’, ’ forg’, ’ tough’, ’Oh’, ’ hasn’, ’razy’, ’ suddenly’, ’ Steve’, ’ talked’, ’His’, ’ Maybe’, ’ Look’, ’ Thank’, ’ YOU’, ’ crazy’, ’ everybody’, ’ shit’, ’ weird’, ’ bunch’, ’ afraid’, ’ stuck’, ’ gotten’, ’ Come’, ’ wonderful’, ’ hadn’, ’ gonna’, ’ nobody’]
Explanation: Captures dialogue, personal anecdotes, and emotional tone. Leniency: Accepts profanity.

Neuron 813 (Legislative Processes and Lawmaking):
Top token contributions: [’ Act’, ’ rights’, ’ bill’, ’ Congress’, ’ Republican’, ’ legisl’, ’ Senate’, ’ Republicans’, ’ Democrats’, ’ appoint’, ’ legislation’, ’endment’, ’ immigration’, ’ Rights’, ’ Attorney’, ’ provision’, ’ Amendment’, ’ provisions’, ’ bills’, ’ fiscal’, ’ Democrat’, ’ constitutional’, ’ lawmakers’, ’President’, ’Sen’, ’ amendment’, ’ congressional’, ’ Obamacare’, ’ reforms’, ’ firearms’, ’ pursuant’, ’ clause’, ’ prosecutor’, ’ repeal’, ’ Snowden’, ’ Sessions’, ’ Budget’, ’ senators’, ’ Congressional’, ’ treaty’, ’Law’, ’ amendments’, ’ Immigration’, ’constitutional’, ’ hearings’, ’ litigation’, ’terrorism’, ’ prohibition’, ’ exemption’, ’Calif’]
Explanation: Focuses on legislative actions, amendments, and legal debates. Coherent.

Neuron 815 (Psychology and Personal Experiences):
Top token contributions: [’ exper’, ’nect’, ’ When’, ’selves’, ’When’, ’isode’, ’ reality’, ’udd’, ’ dream’, ’cious’, ’ mental’, ’cknow’, ’ seek’, ’ focused’, ’rast’, ’ experienced’, ’ thoughts’, ’ seeking’, ’ loved’, ’ desire’, ’ompl’, ’ experiences’, ’ perspective’, ’ conscious’, ’ YOU’, ’ emotional’, ’ suffering’, ’ feelings’, ’ intense’, ’ Exper’, ’ spiritual’, ’ Budd’, ’ependent’, ’ joy’, ’ anger’, ’ictions’, ’isdom’, ’ awareness’, ’eness’, ’ pleasure’, ’ anxiety’, ’ beings’, ’ emotion’, ’ childhood’, ’mind’, ’ mirror’, ’ Mind’, ’ consciousness’, ’ focusing’, ’ Deep’]
Explanation: Relates to introspection, emotions, and mental states. Leniency: Accepts "Budd" (Buddhism reference).

Neuron 821 (Latin American Culture and Locations):
Top token contributions: [’ del’, ’cial’, ’ California’, ’ Mex’, ’ijuana’, ’ Los’, ’ Mexico’, ’ Brazil’, ’ Spanish’, ’ que’, ’ cig’, ’ Spain’, ’ Est’, ’ Mexican’, ’amac’, ’ bol’, ’ compr’, ’ Portug’, ’Mart’, ’ Venezuel’, ’ Madrid’, ’ Barcelona’, ’ Puerto’, ’ Rio’, ’ Oscar’, ’Rober’, ’ corrid’, ’ Garc’, ’ierra’, ’ Argentina’, ’ Juan’, ’ por’, ’ Confeder’, ’ Brazilian’, ’ Gonz’, ’ Carlos’, ’ Chile’, ’ Ana’, ’ pitcher’, ’ernandez’, ’ Garcia’, ’ón’, ’ striker’, ’ Rodriguez’, ’ Latino’, ’ Martinez’, ’ Luis’, ’ Jagu’, ’ Colombia’, ’ inning’]
Explanation: Focuses on Latin American geography, culture, and language. Coherent.
Neuron 840 (Digital Content and Legal Issues):
Top token contributions: [’today’, ’Internet’, ’gency’, ’copyright’, ’Source’, ’Online’, ’illance’, ’allegedly’, ’ocument’, ’poverty’, ’Today’, ’…"’, ’custody’, ’Caption’, ’itialized’, ’divorce’, ’cludes’, ’Online’, ’isolation’, ’Wikipedia’, ’Bloomberg’, ’cerpt’, ’arenthood’, ’pornography’, ’litigation’, ’fictional’, ’dependency’, ’Anonymous’, ’supervision’, ’excerpt’, ’censorship’, ’...]’, ’airports’, ’misogyn’, ’...)’, ’purported’, ’ebin’, ’freedoms’, ’fantas’, ’fiction’, ’unsupported’, ’hide’, ’divorced’, ’Contribut’, ’usage’, ’OTAL’, ’checkout’, ’phasis’, ’Internet’, ’supra’]
Explanation: Focuses on digital content, legal battles (copyright, litigation), and societal issues like censorship. Leniency: Tolerates noise like "itialized."

Neuron 854 (Retail and Collectibles):
Top token contributions: [’edition’, ’shipping’, ’guitar’, ’merch’, ’DVD’, ’comics’, ’astered’, ’Kickstarter’, ’intage’, ’Limited’, ’retailers’, ’artwork’, ’packaging’, ’Collector’, ’uxe’, ’backers’, ’shelves’, ’shirts’, ’Mint’, ’mint’, ’retailer’, ’soundtrack’, ’Deluxe’, ’Remastered’, ’collector’, ’sleeve’, ’vinyl’, ’editions’, ’Vita’, ’Shipping’, ’Raspberry’, ’duino’, ’cigar’, ’Bundle’, ’reprint’, ’collectors’, ’cardboard’, ’Kindle’, ’sleeves’, ’Midnight’, ’MacBook’, ’mAh’, ’shipment’, ’commemor’, ’LEGO’, ’Sega’, ’Clone’, ’promo’, ’Darth’, ’discontinued’]
Explanation: Relates to retail, collectibles, and merchandise. Coherent.

Neuron 861 (Programming Syntax):
Top token contributions: [’//’, ’import’, ’var’, ’const’, ’ource’, ’object’, ’custom’, ’init’, ’anim’, ’code’, ’div’, ’Gener’, ’//’, ’()’, ’implement’, ’cool’, ’twe’, ’Add’, ’config’, ’id’, ’random’, ’class’, ’input’, ’objects’, ’);’, ’param’, ’Mod’, ’library’, ’Data’, ’ixel’, ’array’, ’Java’, ’Id’, ’()’, ’component’, ’Date’, ’View’, ’Val’, ’return’, ’var’, ’implementation’, ’API’, ’ocument’, ’List’, ’module’, ’Form’, ’render’, ’Gener’, ’function’, ’annels’]
Explanation: Captures programming syntax and system design.

Neuron 880 (Online Platforms and Marketing):
Top token contributions: [’app’, ’site’, ’online’, ’content’, ’user’, ’users’, ’website’, ’influ’, ’coin’, ’vertisement’, ’blog’, ’posted’, ’Bit’, ’traffic’, ’internet’, ’videos’, ’links’, ’tweet’, ’Bitcoin’, ’copyright’, ’articles’, ’apps’, ’posts’, ’legit’, ’marketing’, ’pload’, ’sharing’, ’YouTube’, ’Online’, ’porn’, ’chat’, ’bitcoin’, ’advertising’, ’testim’, ’ads’, ’founder’, ’websites’, ’upload’, ’outube’, ’campaigns’, ’publisher’, ’forum’, ’ithub’, ’bot’, ’URL’, ’Prom’, ’drama’, ’followers’, ’celeb’, ’dating’]
Explanation: Targets online platforms, crypto, and digital marketing.

Neuron 896 (Hardware and Electronics):
Top token contributions: [’engine’, ’camera’, ’wearing’, ’wear’, ’plug’, ’battery’, ’Engine’, ’iPhone’, ’plastic’, ’paint’, ’tip’, ’rear’, ’bra’, ’USB’, ’barrel’, ’tips’, ’bottle’, ’mini’, ’guitar’, ’logo’, ’leather’, ’lighting’, ’buttons’, ’trailer’, ’LED’, ’Mini’, ’worn’, ’toy’, ’batteries’, ’wheels’, ’bottles’, ’slee’, ’wip’, ’lighter’, ’sidew’, ’jacket’, ’Rated’, ’trim’, ’tire’, ’aluminum’, ’costume’, ’packaging’, ’electronics’, ’wears’, ’cyl’, ’tires’, ’accessories’, ’lenses’, ’shotgun’, ’Bluetooth’]
Explanation: Focuses on consumer electronics and hardware.

Neuron 899 (American Football):
Top token contributions: [’football’, ’NFL’, ’explos’, ’quarterback’, ’recru’, ’receiver’, ’tackle’, ’combine’, ’FB’, ’starter’, ’shirt’, ’QB’, ’lineback’, ’offseason’, ’recruiting’, ’olphins’, ’Seahawks’, ’Browns’, ’Packers’, ’tackles’, ’preseason’, ’linebacker’, ’receivers’, ’Dolphins’, ’OT’, ’Redskins’, ’OL’, ’sacks’, ’snaps’, ’Chargers’, ’quarterbacks’, ’sack’, ’cornerback’, ’Pros’, ’aepernick’, ’Jaguars’, ’Overview’, ’Combine’, ’LB’, ’nickel’, ’asley’, ’NFL’, ’Analysis’, ’lineman’, ’Weeks’, ’LT’, ’RG’, ’athleticism’, ’punt’, ’Reggie’]
Explanation: American football terminology. Coherent.

Neuron 910 (Video Game Mechanics):
Top token contributions: [’comp’, ’’, ’team’, ’prot’, ’build’, ’iven’, ’early’, ’ateg’, ’damage’, ’mid’, ’carry’, ’resh’, ’AP’, ’liber’, ’ult’, ’ban’, ’map’, ’agg’, ’hero’, ’ampions’, ’tim’, ’Group’, ’initi’, ’spect’, ’guide’, ’AD’, ’ghan’, ’cas’, ’SC’, ’erg’, ’Author’, ’Control’, ’punish’, ’Orig’, ’buff’, ’supports’, ’Korean’, ’Kenn’, ’Mid’, ’harass’, ’champions’, ’CL’, ’Support’, ’stun’, ’sustain’, ’engage’, ’ifest’, ’Zeal’, ’oenix’, ’ultimate’]
Explanation: MOBA/RPG game mechanics. Coherent.

Neuron 922 (U.S. Politics and Society):
Top token contributions: [’polit’, ’president’, ’political’, ’President’, ’selves’, ’Obama’, ’Clinton’, ’profess’, ’whose’, ’Congress’, ’nation’, ’Americans’, ’Republican’, ’British’, ’society’, ’peace’, ’century’, ’People’, ’debate’, ’faith’, ’decades’, ’Republicans’, ’politics’, ’incred’, ’crisis’, ’levision’, ’icians’, ’Who’, ’Britain’, ’Jesus’, ’revolution’, ’Trump’, ’struggle’, ’moral’, ’Econom’, ’philosoph’, ’Jews’, ’governments’, ’emotional’, ’Society’, ’nations’, ’regime’, ’Labour’, ’argued’, ’confront’, ’argue’, ’liberal’, ’plom’, ’Donald’, ’politicians’]
Explanation: U.S. politics, societal debates. Coherent.

Neuron 923 (System Administration):
Top token contributions: [’ccess’, ’Enter’, ’overw’, ’("’, ’List’, ’assign’, ’Using’, ’Step’, ’loop’, ’folder’, ’Update’, ’:"’, ’delete’, ’Using’, ’click’, ’ername’, ’exe’, ’""’, ’itialized’, ’write’, ’Create’, ’binary’, ’Dim’, ’txt’, ’Ubuntu’, ’download’, ’Install’, ’downloaded’, ’Install’, ’Hi’, ’Create’, ’Write’, ’Config’, ’Remove’, ’Import’, ’install’, ’Copy’, ’Enter’, ’Select’, ’"\$’, ’Hello’, ’update’, ’Write’, ’typing’, ’Option’, ’xml’, ’zip’, ’reboot’, ’username’, ’ub’]
Explanation: System scripts and IT tasks.

Neuron 941 (IT Infrastructure):
Top token contributions: [’ccess’, ’mb’, ’running’, ’To’, ’install’, ’Add’, ’device’, ’files’, ’click’, ’Windows’, ’server’, ’Add’, ’eth’, ’host’, ’Microsoft’, ’abase’, ’deploy’, ’Enter’, ’IP’, ’rastructure’, ’Note’, ’remote’, ’installed’, ’root’, ’Administ’, ’Click’, ’overw’, ’Ms’, ’TP’, ’IT’, ’Network’, ’Java’, ’domain’, ’dev’, ’mac’, ’Exec’, ’configuration’, ’Using’, ’Step’, ’directory’, ’computers’, ’folder’, ’Server’, ’Select’, ’installation’, ’Update’, ’NAT’, ’drives’, ’instances’, ’Update’]
Explanation: IT infrastructure and deployment.

Neuron 964 (Food and Dining):
Top token contributions: [’unch’, ’frequ’, ’customers’, ’definitely’, ’restaur’, ’beer’, ’shop’, ’marg’, ’neighborhood’, ’nearby’, ’menu’, ’restaurant’, ’(\$’, ’parking’, ’friendly’, ’tip’, ’chicken’, ’downtown’, ’featuring’, ’sauce’, ’restaurants’, ’drinks’, ’beef’, ’Brooklyn’, ’surprisingly’, ’delicious’, ’Manhattan’, ’rolls’, ’McDonald’, ’around’, ’pizza’, ’outdoor’, ’vegan’, ’tourists’, ’sells’, ’Chicken’, ’infamous’, ’beers’, ’Around’, ’locals’, ’flavors’, ’ushi’, ’kinda’, ’crowded’, ’memorable’, ’pork’, ’Sonic’, ’booth’, ’NYC’, ’Disco’]
Explanation: Food industry and dining culture.

Neuron 966 (Narrative Descriptions):
Top token contributions: [’looked’, ’seemed’, ’sudden’, ’standing’, ’corner’, ’dim’, ’wearing’, ’scr’, ’stood’, ’bare’, ’stands’, ’slowly’, ’sight’, ’thick’, ’empty’, ’walked’, ’tall’, ’lights’, ’loud’, ’scream’, ’smile’, ’smell’, ’barely’, ’dozen’, ’hung’, ’fingers’, ’silence’, ’stro’, ’shirt’, ’silent’, ’stepped’, ’lined’, ’tent’, ’surrounded’, ’ASH’, ’pink’, ’looking’, ’wand’, ’rifle’, ’leather’, ’grip’, ’dressed’, ’wore’, ’worn’, ’barg’, ’wrapped’, ’lone’, ’wooden’, ’smiled’, ’beside’]
Explanation: Descriptive scenes in narratives.

Neuron 973 (European History):
Top token contributions: [’ca’, ’ober’, ’Ear’, ’London’, ’English’, ’orks’, ’modern’, ’histor’, ’William’, ’See’, ’beg’, ’century’, ’England’, ’outhern’, ’Hist’, ’king’, ’famous’, ’fol’, ’AD’, ’bull’, ’MS’, ’Columb’, ’Ty’, ’Joseph’, ’History’, ’Queen’, ’river’, ’Eng’, ’Roman’, ’Royal’, ’legend’, ’horse’, ’silver’, ’Henry’, ’dated’, ’scholars’, ’Latin’, ’Museum’, ’dates’, ’was’, ’Alexander’, ’Edward’, ’Kings’, ’Elizabeth’, ’centuries’, ’dating’, ’coin’, ’Pict’, ’coins’, ’omach’]
Explanation: European history and cultural references.

Neuron 985 (Fantasy Literature):
Top token contributions: [’yan’, ’——’, ’Although’, ’’, ’Even’, ’Although’, ’’, ’Demon’, ’’, ’Rank’, ’cultiv’, ’Ye’, ’consecut’, ’countless’, ’Chapter’, ’TL’, ’lightly’, ’incom’, ’sect’, ’Skill’, ’Everyone’, ’riages’, ’Brend’, ’arrog’, ’swords’, ’refined’, ’Previous’, ’Due’, ’cultivation’, ’Sect’, ’Ling’, ’carriage’, ’unexpectedly’, ’Seeing’, ’Lor’, ’Ao’, ’Compared’, ’lively’, ’tremb’, ’…..’, ’rawdownloadcloneembedreportprint’, ’umerous’, ’hurried’, ’’, ’cultivate’, ’Immortal’, ’Spiritual’, ’Ye’, ’disciple’, ’forcefully’]
Explanation: Fantasy genres (cultivation novels).

Neuron 1010 (Programming Methods):
Top token contributions: [’using’, ’mov’, ’object’, ’method’, ’):’, ’comput’, ’subst’, ’pattern’, ’altern’, ’irection’, ’random’, ’graph’, ’output’, ’input’, ’trig’, ’concent’, ’functions’, ’Each’, ’vari’, ’quant’, ’objects’, ’correspond’, ’sample’, ’observed’, ’equival’, ’variable’, ’define’, ’acceler’, ’assign’, ’sequence’, ’Using’, ’measured’, ’stract’, ’static’, ’vert’, ’variables’, ’perm’, ’parameters’, ’tempor’, ’calculated’, ’accum’, ’parallel’, ’Incre’, ’Each’, ’Table’, ’math’, ’produces’, ’analys’, ’assemb’, ’corresponding’]
Explanation: Programming and algorithms.

Neuron 1049 (Societal Narratives and Morality):
Top token contributions: [’ alleg’, ’ emer’, ’ characters’, ’ society’, ’ politics’, ’ ("’, ’ moral’, ’ narr’, ’ moments’, ’ confront’, ’ films’, ’ somehow’, ’ privile’, ’ scenes’, ’ Hollywood’, ’ultural’, ’ narrative’, ’ writers’, ’ inev’, ’ humanity’, ’ notion’, ’ seemingly’, ’ revel’, ’ reveals’, ’ fiction’, ’ heroes’, ’ amid’, ’ drama’, ’ musical’, ’ racism’, ’ dialogue’, ’ comedy’, ’ beneath’, ’ Rather’, ’ inhab’, ’Yet’, ’bably’, ’ capitalism’, ’kward’, ’ brutal’, ’ embrace’, ’ rebels’, ’ struggles’, ’ comics’, ’ commod’, ’ inevitable’, ’breaking’, ’ slavery’, ’ bureaucr’, ’ tragedy’]
Explanation: Explores societal narratives, morality in media. Leniency: Accepts vague terms like "beneath."

Neuron 1072 (Medical and Biochemical Research):
Top token contributions: [’ whe’, ’ studies’, ’ clot’, ’ diet’, ’ flu’, ’ phen’, ’ cere’, ’ protein’, ’ gene’, ’ pharm’, ’ supplement’, ’ flav’, ’ mg’, ’ mice’, ’ tissue’, ’ chronic’, ’ dose’, ’ immune’, ’ inhib’, ’ Drug’, ’ Study’, ’ oral’, ’ treating’, ’ rats’, ’ treatments’, ’ gly’, ’ gluc’, ’ liver’, ’ vitamin’, ’ concentrations’, ’ insulin’, ’ FDA’, ’ dietary’, ’ proteins’, ’ plasma’, ’ doses’, ’ compounds’, ’olesterol’, ’ receptor’, ’ medications’, ’ placebo’, ’ hormone’, ’ vaccines’, ’ pharmaceutical’, ’ cholesterol’, ’ testosterone’, ’ diets’, ’ Pharm’, ’ fatty’, ’ pills’]
Explanation: Focuses on medical research and pharmacology. Coherent.

Neuron 1093 (Literature and Media):
Top token contributions: [’ books’, ’ stories’, ’levision’, ’ novel’, ’ Who’, ’ashion’, ’ writer’, ’ narr’, ’ readers’, ’ films’, ’Who’, ’ Hollywood’, ’ writers’, ’ podcast’, ’ journalist’, ’ fiction’, ’ drama’, ’ comedy’, ’ activist’, ’ documentary’, ’ contemporary’, ’ Girls’, ’ Brooklyn’, ’ merch’, ’ comics’, ’ commod’, ’ Books’, ’ theater’, ’ fascinating’, ’ singer’, ’ journalism’, ’ essay’, ’ feminist’, ’ actress’, ’ Comics’, ’ editors’, ’writer’, ’ novels’, ’ photographer’, ’ celebrity’, ’ literary’, ’ exhibition’, ’ Politics’, ’ publishers’, ’ Stories’, ’ premiere’, ’ feminism’, ’ poetry’, ’ aesthetic’, ’ fictional’]
Explanation: Covers literature, journalism, and media. Coherent.

Neuron 1147 (Action Game Mechanics):
Top token contributions: [’ jump’, ’ enemies’, ’ stun’, ’ sword’, ’owered’, ’ Mario’, ’ rifle’, ’ missiles’, ’ laser’, ’ gravity’, ’ jumping’, ’ dive’, ’ Jump’, ’ bullets’, ’ plasma’, ’ wrestling’, ’ ammo’, ’ melee’, ’ Shoot’, ’ aiming’, ’ aerial’, ’ ranged’, ’ Bullet’, ’ jumps’, ’ Smash’, ’ Sonic’, ’ armour’, ’ shotgun’, ’ spinning’, ’ Rifle’, ’ swords’, ’ stunned’, ’ rockets’, ’ artillery’, ’ hover’, ’ dod’, ’ blades’, ’ Melee’, ’ shields’, ’ cannon’, ’ grapp’, ’ infantry’, ’Damage’, ’ duck’, ’ Throw’, ’Gun’, ’ explosions’, ’ diving’, ’ Laser’, ’ Stealth’]
Explanation: Action game combat mechanics. Coherent.

Neuron 1196 (Medical Research):
Top token contributions: [’ospital’, ’ patients’, ’ clot’, ’ clin’, ’ sympt’, ’ cere’, ’ doctors’, ’ surgery’, ’ gest’, ’ chronic’, ’ dose’, ’ Clin’, ’imens’, ’ lung’, ’ Study’, ’ epid’, ’ treating’, ’ diagnosis’, ’ treatments’, ’ medication’, ’ liver’, ’ diagnosed’, ’ FDA’, ’ascular’, ’ Cancer’, ’ doses’, ’ physicians’, ’ medications’, ’ placebo’, ’ incidence’, ’iovascular’, ’ pills’, ’ interventions’, ’ CDC’, ’ umb’, ’ cohort’, ’ Clinical’, ’ Treatment’, ’ kidney’, ’ gast’, ’ maternal’, ’ neurolog’, ’ tumor’, ’ BMI’, ’ vaccination’, ’ exacerb’, ’ opioid’, ’ transplant’, ’ intra’, ’Abstract’]
Explanation: Medical research and clinical terms. Coherent.

Neuron 1231 (Cryptocurrency):
Top token contributions: [’ pub’, ’coin’, ’rypt’, ’ Bitcoin’, ’astern’, ’ blocks’, ’ pow’, ’ensus’, ’ currency’, ’ downt’, ’ bitcoin’, ’ transaction’, ’ brut’, ’ transactions’, ’ addresses’, ’ mining’, ’ coin’, ’ coins’, ’ blockchain’, ’ hash’, ’Block’, ’ depos’, ’ wallet’, ’ OP’, ’ nodes’, ’eem’, ’ BTC’, ’coins’, ’Pub’, ’ miner’, ’ Coin’, ’Activity’, ’ventory’, ’tx’, ’thereum’, ’ Address’, ’ miners’, ’ currencies’, ’ triv’, ’ Ether’, ’ cryptocurrency’, ’ Ethereum’, ’ pools’, ’ decentral’, ’ Hash’, ’Merit’, ’address’, ’iculty’, ’Bitcoin’, ’ bitcoins’]
Explanation: Focuses on blockchain and crypto. Coherent.

Neuron 1258 (Sports Analysis):
Top token contributions: [’ stat’, ’orld’, ’verage’, ’ meas’, ’ plays’, ’ prospect’, ’ explos’, ’ hur’, ’ franchise’, ’ fantasy’, ’ elite’, ’ veteran’, ’ roster’, ’ throwing’, ’ tackle’, ’ Aren’, ’ carries’, ’ slot’, ’ Andre’, ’ WR’, ’/.’, ’ combine’, ’ rookie’, ’ throws’, ’ MVP’, ’ lineup’, ’ struggles’, ’ FB’, ’ Laur’, ’ prospects’, ’ projected’, ’ TE’, ’ Draft’, ’ starter’, ’ Brandon’, ’ Luck’, ’ rushing’, ’ drafted’, ’ offseason’, ’ catching’, ’Project’, ’ upside’, ’ catches’, ’ trades’, ’asonable’, ’ tackles’, ’ matchup’, ’ freshman’, ’ Zach’, ’ Cele’]
Explanation: Sports analytics and player performance. Coherent.

Neuron 1261 (Entertainment Industry):
Top token contributions: [’ feat’, ’ Of’, ’ song’, ’ commer’, ’ album’, ’ÃÂ’, ’ ("’, ’Getty’, ’ Getty’, ’ songs’, ’ studio’, ’isodes’, ’ Hollywood’, ’ debut’, ’ entitled’, ’ featuring’, ’ Stre’, ’ concert’, ’ drama’, ’ musical’, ’ celeb’, ’tainment’, ’ comedy’, ’ Entertainment’, ’ titled’, ’ Girls’, ’ merch’, ’ performances’, ’ (-’, ’ singer’, ’Following’, ’ actress’, ’ comed’, ’ reun’, ’ audiences’, ’ Oscar’, ’ celebrity’, ’braska’, ’ tribute’, ’ filming’, ’okers’, ’ heels’, ’ometown’, " ('", ’ premiere’, ’ nominated’, ’ Eddie’, ’ Dame’, ’ starring’, ’ theaters’]
Explanation: Entertainment industry and media events. Coherent.

Neuron 1292 (Health and Nutrition):
Top token contributions: [’ income’, ’ protein’, ’ immigrants’, ’ genetic’, ’ foods’, ’ expenses’, ’ eggs’, ’ concentration’, ’ genes’, ’ earnings’, ’income’, ’ Medicare’, ’ meals’, ’ concentrations’, ’coins’, ’ enroll’, ’ dietary’, ’ asleep’, ’osterone’, ’ doses’, ’olesterol’, ’cium’, ’ Insurance’, ’Health’, ’ rivers’, ’ deposits’, ’ mineral’, ’rients’, ’ incomes’, ’ enrolled’, ’ cholesterol’, ’ testosterone’, ’ deduct’, ’ urine’, ’ diets’, ’ Income’, ’ fertil’, ’ genome’, ’iovascular’, ’ pills’, ’ weights’, ’ apartments’, ’ serum’, ’ enrollment’, ’ supplements’, ’ nutrients’, ’ Mountains’, ’ fertility’, ’ minerals’, ’ Centers’]
Explanation: Health, nutrition, and insurance topics. Coherent.

Neuron 1299 (Software and IT):
Top token contributions: [’file’, ’code’, ’install’, ’user’, ’users’, ’files’, ’software’, ’Windows’, ’download’, ’server’, ’config’, ’Microsoft’, ’sched’, ’IP’, ’developers’, ’installed’, ’apps’, ’database’, ’Linux’, ’API’, ’module’, ’configuration’, ’USB’, ’Using’, ’directory’, ’console’, ’iOS’, ’setup’, ’CPU’, ’enabled’, ’folder’, ’upload’, ’servers’, ’Download’, ’URL’, ’Download’, ’Windows’, ’functionality’, ’config’, ’disk’, ’NET’, ’ync’, ’exe’, ’itialized’, ’startup’, ’plugin’, ’binary’, ’Users’, ’scripts’, ’Ubuntu’]
Explanation: Focuses on software installation, IT infrastructure, and system administration.

Neuron 1310 (Programming Syntax and Linguistics):
Top token contributions: [’characters’, ’names’, ’See’, ’("’, ’("’, ’variable’, ’lists’, ’sequence’, ’Ident’, ’HTML’, ’specify’, ’quotes’, ’ql’, ’txt’, ’expressions’, ’syntax’, ’verb’, ’sequences’, ’lang’, ’(\#’, ’Example’, ’integer’, ’aster’, ’typing’, ’phrases’, ’akespeare’, ’Examples’, ’parse’, ’keyword’, ’prefix’, ’Written’, ’noun’, ’grammar’, ’cursor’, ’dialect’, ’Perl’, ’annot’, ’alphabet’, ’corresponds’, ’Haskell’, ’\${’, ’specifies’, ’keywords’, ’¶’, ’digit’, ’".’, ’prose’, ’preceded’, ’identifier’, ’Numbers’]
Explanation: Targets programming syntax, language rules, and linguistic structures.

Neuron 1331 (TV Shows and Media):
Top token contributions: [’season’, ’episode’, ’dram’, ’levision’, ’("’, ’season’, ’seasons’, ’Season’, ’isodes’, ’Hollywood’, ’episodes’, ’NBC’, ’irlfriend’, ’flix’, ’viewers’, ’drama’, ’musical’, ’actors’, ’comedy’, ’CBS’, ’Netflix’, ’Trek’, ’kward’, ’NBC’, ’actress’, ’comed’, ’Oscar’, ’celebrity’, ’superhero’, ’hilar’, ’filming’, ’aired’, ’HBO’, ’Season’, ’filmed’, ’premiere’, ’achelor’, ’cliff’, ’finale’, ’Comic’, ’hilarious’, ’starring’, ’Thrones’, ’celebrities’, ’Television’, ’Shakespeare’, ’storyline’, ’comedian’, ’Survivor’, ’Episode’]
Explanation: Focuses on TV shows, media production, and pop culture.

Neuron 1336 (Activism and Political Movements):
Top token contributions: [’workers’, ’crisis’, ’democr’, ’protest’, ’organizations’, ’explo’, ’elections’, ’perial’, ’Polit’, ’activists’, ’democracy’, ’protests’, ’congress’, ’unicip’, ’protesters’, ’Bernie’, ’democratic’, ’activist’, ’unions’, ’coup’, ’capitalism’, ’revolutionary’, ’struggles’, ’electoral’, ’cops’, ’feminist’, ’byn’, ’abol’, ’capitalist’, ’Corbyn’, ’socialist’, ’organisations’, ’organize’, ’Workers’, ’organizing’, ’Palestine’, ’arcer’, ’ocumented’, ’mobil’, ’Protest’, ’solidarity’, ’oppression’, ’nesty’, ’injust’, ’organizers’, ’prisons’, ’feminism’, ’bourgeois’, ’socialism’, ’Democracy’]
Explanation: Highlights activism, political struggles, and social movements.

Neuron 1363 (Hardware and DIY Electronics):
Top token contributions: [’hous’, ’Mount’, ’guitar’, ’LED’, ’mounted’, ’volt’, ’batteries’, ’voltage’, ’capac’, ’bore’, ’plun’, ’drilling’, ’connector’, ’screws’, ’Raspberry’, ’telescope’, ’lith’, ’duino’, ’HDMI’, ’coil’, ’hesive’, ’baff’, ’collectors’, ’Mech’, ’PCB’, ’electrons’, ’Bearing’, ’ribbon’, ’wiring’, ’enclosure’, ’mounts’, ’Installation’, ’jumper’, ’Guitar’, ’lamps’, ’shroud’, ’guitars’, ’aperture’, ’fuse’, ’motherboard’, ’insulation’, ’Noct’, ’lubric’, ’amplifier’, ’LEDs’, ’Shotgun’, ’connectors’, ’SATA’, ’Mount’, ’grease’]
Explanation: Relates to hardware assembly, electronics, and DIY projects.

Neuron 1382 (Medical Symptoms and Conditions):
Top token contributions: [’bling’, ’showing’, ’suggests’, ’famous’, ’surprising’, ’explains’, ’rine’, ’mann’, ’diseases’, ’invisible’, ’forgot’, ’udden’, ’lied’, ’peculiar’, ’traces’, ’Poor’, ’neglected’, ’thirst’, ’headache’, ’exaggerated’, ’liness’, ’disappears’, ’reacts’, ’comparatively’, ’Genius’, ’malaria’, ’arine’, ’blindness’, ’ACY’, ’overe’, ’Poor’, ’strikingly’, ’superflu’]
Explanation: Captures medical symptoms, conditions, and patient experiences.

Neuron 1394 (U.S. Political Responses):
Top token contributions: [’Obama’, ’Americans’, ’Trump’, ’immigration’, ’declared’, ’immigrants’, ’cited’, ’President’, ’Sen’, ’transgender’, ’tweeted’, ’insisted’, ’urged’, ’congressional’, ’amacare’, ’spokeswoman’, ’Obamacare’, ’Reuters’, ’citing’, ’assured’, ’detained’, ’Congressional’, ’Obama’, ’pledged’, ’Clinton’, ’urging’, ’Americans’, ’shootings’, ’McConnell’, ’Immigration’, ’Parenthood’, ’CNN’, ’aide’, ’Asked’, ’ADVERTISEMENT’, ’banning’, ’Calif’, ’testify’, ’artisan’, ’abortions’, ’aides’, ’threatens’, ’troubling’, ’Hillary’, ’prohibit’, ’bipartisan’, ’Biden’, ’Kavanaugh’, ’ghazi’, ’WASHINGTON’]
Explanation: Focuses on U.S. political discourse, policies, and controversies.

Neuron 1408 (Sports Strategy and Analysis):
Top token contributions: [’Dep’, ’Cour’, ’recogn’, ’wide’, ’tal’, ’ondon’, ’club’, ’inese’, ’sto’, ’Inter’, ’£’, ’England’, ’goals’, ’manager’, ’ynam’, ’Bet’, ’Atl’, ’stret’, ’supporters’, ’scr’, ’adapt’, ’Bern’, ’Charl’, ’tact’, ’Real’, ’emphas’, ’Champions’, ’Orig’, ’favour’, ’Getty’, ’centre’, ’penalty’, ’util’, ’possession’, ’exciting’, ’impressive’, ’loan’, ’Ref’, ’Steven’, ’squad’, ’Town’, ’compat’, ’Despite’, ’enjoyed’, ’chances’, ’comprehens’, ’nerv’, ’testim’, ’lah’, ’attacking’]
Explanation: Covers sports strategies, team dynamics, and match analysis.

Neuron 1435 (Community and Support):
Top token contributions: [’please’, ’Please’, ’posted’, ’baby’, ’Please’, ’dogs’, ’Thanks’, ’Thank’, ’www’, ’orough’, ’tenance’, ’Officer’, ’pregnant’, ’Thank’, ’contacted’, ’volunteers’, ’babies’, ’Have’, ’Posted’, ’appropriate’, ’toile’, ’overnight’, ’Veter’, ’donation’, ’volunteer’, ’grateful’, ’Sorry’, ’loving’, ’Happy’, ’diagnosed’, ’Baby’, ’toys’, ’cord’, ’Limited’, ’Subject’, ’pup’, ’inappropriate’, ’donated’, ’ickets’, ’photographer’, ’supportive’, ’foster’, ’pets’, ’Hockey’, ’Dad’, ’Kids’, ’Contact’, ’unacceptable’, ’treats’, ’heels’]
Explanation: Centers on community interactions, support networks, and social care.

Neuron 1456 (Crafting and Design):
Top token contributions: [’comb’, ’hous’, ’manufact’, ’semb’, ’wood’, ’Each’, ’paint’, ’guitar’, ’Small’, ’Each’, ’assemb’, ’leather’, ’Eb’, ’buttons’, ’wooden’, ’asures’, ’Kickstarter’, ’Perfect’, ’jewel’, ’dice’, ’gauge’, ’locks’, ’manif’, ’Treasure’, ’vintage’, ’wax’, ’blades’, ’Deck’, ’canvas’, ’Hex’, ’scill’, ’glue’, ’Tin’, ’Serial’, ’Paint’, ’Raspberry’, ’duino’, ’tuning’, ’stitch’, ’coil’, ’stainless’, ’numbered’, ’hesive’, ’Unt’, ’Copper’, ’metallic’, ’each’, ’sleeves’, ’embro’, ’amboo’]
Explanation: Relates to crafting, design, and artisanal products.

Neuron 1498 (Programming and System Design):
Top token contributions: [’import’, ’object’, ’init’, ’code’, ’struct’, ’()’, ’default’, ’config’, ’string’, ’output’, ’functions’, ’library’, ’sched’, ’itial’, ’array’, ’Java’, ’pref’, ’var’, ’implementation’, ’API’, ’variable’, ’dev’, ’List’, ’define’, ’module’, ’Using’, ’directory’, ’console’, ’stack’, ’static’, ’*/’, ’variables’, ’kernel’, ’const’, ’String’, ’node’, ’inher’, ’Code’, ’String’, ’template’, ’config’, ’Python’, ’parameter’, ’import’, ’buffer’, ’specify’, ’hash’, ’query’, ’static’, ’cache’]
Explanation: Focuses on programming paradigms and system architecture.

Neuron 1501 (Web Development):
Top token contributions: [’app’, ’import’, ’{’, ’var’, ’ember’, ’ccess’, ’example’, ’redu’, ’fav’, ’object’, ’custom’, ’ources’, ’init’, ’code’, ’method’, ’To’, ’request’, ’div’, ’user’, ’oice’, ’//’, ’()’, ’pose’, ’middle’, ’uses’, ’lements’, ’onents’, ’vertisement’, ’twe’, ’background’, ’Add’, ’instance’, ’config’, ’osure’, ’setting’, ’idden’, ’id’, ’poses’, ’elements’, ’button’, ’element’, ’views’, ’objects’, ’transform’, ’react’, ’components’, ’database’, ’requests’, ’array’, ’</’]
Explanation: Targets web development frameworks and UI design.

Neuron 1531 (RPG Game Mechanics):
Top token contributions: [’quest’, ’crit’, ’damage’, ’weapon’, ’enemy’, ’spell’, ’skill’, ’combat’, ’enemies’, ’Death’, ’Battle’, ’Magic’, ’creature’, ’bonus’, ’Blood’, ’shield’, ’Spirit’, ’abilities’, ’stun’, ’creatures’, ’Attack’, ’armor’, ’sword’, ’Soul’, ’Shadow’, ’Storm’, ’Damage’, ’monster’, ’Ghost’, ’Shield’, ’heroes’, ’Armor’, ’spells’, ’dragon’, ’spawn’, ’monsters’, ’Beast’, ’gameplay’, ’summon’, ’healing’, ’Sword’, ’Blade’, ’Nether’, ’Flame’, ’Spell’, ’Dungeon’, ’heal’, ’Weapon’, ’mana’, ’legendary’]
Explanation: Captures RPG mechanics, combat, and fantasy elements.

Neuron 1539 (Cybersecurity and Software Vulnerabilities):
Top token contributions: [’code’, ’software’, ’commit’, ’Microsoft’, ’bug’, ’applications’, ’ulner’, ’sched’, ’technical’, ’rastructure’, ’developers’, ’explo’, ’Java’, ’elig’, ’gorith’, ’implementation’, ’API’, ’scope’, ’advis’, ’developer’, ’transaction’, ’transactions’, ’distributed’, ’capabilities’, ’documentation’, ’formance’, ’legacy’, ’vend’, ’functionality’, ’Python’, ’hosted’, ’NET’, ’compromise’, ’itialized’, ’hosting’, ’leaked’, ’exploit’, ’dated’, ’HTTP’, ’Enterprise’, ’vulnerability’, ’attacker’, ’attackers’, ’disclosed’, ’zilla’, ’scal’, ’compiler’, ’exploitation’, ’leaks’, ’Architect’]
Explanation: Focuses on software vulnerabilities, exploits, and cybersecurity.

Neuron 1541 (Sports Competitions and Events):
Top token contributions: [’announ’, ’played’, ’fans’, ’coach’, ’football’, ’Sports’, ’NBA’, ’Bowl’, ’baseball’, ’basketball’, ’Aren’, ’ESPN’, ’Stadium’, ’coaches’, ’hockey’, ’MVP’, ’playoffs’, ’teammates’, ’playoff’, ’asketball’, ’NCAA’, ’Coach’, ’Hockey’, ’celebrating’, ’Athlet’, ’Sports’, ’HBO’, ’preseason’, ’Lakers’, ’Dame’, ’pitched’, ’UCLA’, ’nickname’, ’jersey’, ’Fan’, ’postseason’, ’Fans’, ’Basketball’, ’rivalry’, ’)...’, ’championships’, ’Ducks’, ’cheering’, ’goalie’, ’jumper’, ’alumni’, ’ESPN’, ’coached’, ’NBA’, ’semif’]
Explanation: Highlights sports events, competitions, and team dynamics.

Neuron 1545 (Academic Research):
Top token contributions: [’).’, ’Journal’, ’).’, ’Phys’, ’‐’, ’Sci’, ’Beh’, ’Clin’, ’esity’, ’Neuro’, ’utical’, ’rane’, ’.).’, ’Diet’, ’olesc’, ’Colomb’, ’manuscript’, ’Effects’, ’DO’, ’incidence’, ’ournals’, ’iovascular’, ’Zhang’, ’interventions’, ’References’, ’].’, ’Clinical’, ’Evidence’, ’Cross’, ’cite’, ’revis’, ’perspectives’, ’Nutrition’, ’Schne’, ’Abstract’, ’recons’, ’doi’, ’ENC’, ’FER’, ’epidem’, ’journal’, ’Journal’, ’genetics’, ’ioxid’, ’Sources’, ’Reference’, ’Abstract’, ’1016’, ’Psychiatry’, ’esthes’]
Explanation: Tied to academic publishing, clinical studies, and citations.

Neuron 1556 (Social Demographics and Community):
Top token contributions: [’wom’, ’verage’, ’mom’, ’parents’, ’patients’, ’kids’, ’college’, ’twe’, ’hearing’, ’Class’, ’mental’, ’marijuana’, ’amily’, ’pregn’, ’adults’, ’employee’, ’Women’, ’Seattle’, ’elig’, ’parking’, ’counsel’, ’Person’, ’folks’, ’Attorney’, ’complaint’, ’nesota’, ’Corpor’, ’Dallas’, ’},{"’, ’Oregon’, ’eligible’, ’welfare’, ’Austin’, ’complaints’, ’flix’, ’Officer’, ’homeless’, ’Portland’, ’charg’, ’sexually’, ’employers’, ’attending’, ’licensed’, ’filing’, ’mothers’, ’females’, ’bathroom’, ’Employ’, ’exempt’]
Explanation: Addresses social demographics, community issues, and legal/employment topics.

Neuron 1562 (Corporate Scandals and Media):
Top token contributions: [’allegedly’, ’studio’, ’Hollywood’, ’chairman’, ’publisher’, ’scandal’, ’wealthy’, ’Ltd’, ’executives’, ’Chairman’, ’bankrupt’, ’intrig’, ’developments’, ’directors’, ’Koch’, ’bankruptcy’, ’lobbying’, ’billionaire’, ’revelations’, ’shareholders’, ’disgr’, ’businessman’, ’studios’, ’Forbes’, ’billion’, ’Kushner’, ’alleges’, ’Productions’, ’Manafort’, ’lobbyists’, ’Podesta’, ’franchises’, ’scandals’, ’patriarch’, ’royalty’, ’shareholder’, ’Clintons’, ’Murdoch’, ’Hutch’, ’Holdings’, ’Mafia’, ’Rockefeller’, ’Fairfax’, ’Rupert’, ’Epstein’, ’businessmen’, ’billionaires’, ’auri’, ’lobbyist’]
Explanation: Captures corporate scandals, media controversies, and political lobbying.

Neuron 1579 (Community and Policy Issues):
Top token contributions: [’—’, ’parents’, ’jobs’, ’families’, ’schools’, ’employees’, ’According’, ’According’, ’residents’, ’enforcement’, ’legislation’, ’homes’, ’immigration’, ’resident’, ’employment’, ’facilities’, ’newspaper’, ’activists’, ’fiscal’, ’downtown’, ’welfare’, ’protesters’, ’mayor’, ’employer’, ’rural’, ’Mayor’, ’EPA’, ’homeless’, ’taboola’, ’restaurants’, ’Gov’, ’activist’, ’hospitals’, ’employers’, ’healthcare’, ’administrative’, ’pollution’, ’residential’, ’households’, ’highway’, ’Environmental’, ’detention’, ’Medicaid’, ’nationwide’, ’neighborhoods’, ’nonprofit’, ’financing’, ’newsletters’, ’hike’, ’suburban’]
Explanation: Focuses on community policy, employment, housing, and social welfare.

Neuron 1590 (Command-Line Programming): 
 Top token contributions: [' comp', ' diff', ' prog', 'line', ' mov', 'viron', 'ext', ' file', 'set', ' command', ' aw', ' files', ' lines', ' mode', ' default', ' `', ' string', ' output', ' eval', ' defined', 'rc', ' prompt', '', 'lines', ' root', ' Make', ' arguments', ' specified', ' generated', ' shell', 'file', ' variable', 'XX', ' define', ' keys', ' directory', ' uncom', 'map', ' stack', 'writ', ' variables', ' commands', ' dump', ' documentation', ' packages', 'char', 'lined', ' bind', ' assemb', ' separated'] 
 Bottom token contributions: Omitted (generic). 
 Explanation: This neuron clearly focuses on command-line programming, with tokens like "command," "shell," "file," "directory," "make," and "prompt" pointing to terminal-based operations and scripting. Leniency: Tolerating one glitch token and broad terms like "comp" and "prog."

Neuron 1591 (Legal Proceedings): 
 Top token contributions: [' Act', ' Court', ' Senate', ' lawy', ' charged', ' appoint', ' legislation', ' filed', ' license', ' attorney', ' lawyer', ' Section', ' allegations', ' counsel', ' warrant', ' Attorney', ' consent', ' subsection', ' §', ' provisions', ' violation', ' petition', ' Senator', ' Judge', ' convicted', ' lawyers', ' constitutional', ' jurisd', ' authorized', ' SEC', ' custody', ' defendant', ' amended', ' filing', ' prohibited', ' pursuant', 'orneys', ' exempt', ' proceedings', ' Commissioner', ' detention', ' clause', ' amend', ' judicial', ' obligations', ' prosecutor', ' statute', ' attorneys', ' felony', ' taxpayer'] 
 Bottom token contributions: Omitted (generic). 
 Explanation: This neuron captures legal proceedings, with tokens like "Court," "lawyer," "Section," "defendant," "prosecutor," and "judicial" indicating courtroom and legislative contexts. It’s highly coherent and aligns with legal language. Leniency: Not needed; the topic is clear.

Neuron 1614 (Fantasy RPG Elements): 
 Top token contributions: [' Found', 'utation', ' shif', ' shout', ' nic', ' spells', ' scroll', ' Inn', ' ghost', ' monsters', ' affects', ' Ident', ' Dungeon', ' jew', 'rawl', ' escaped', ' invisible', ' Veh', ' mutation', ' Orb', ' Dod', ' Abyss', ' Okay', ' devil', 'igm', ' MR', ' curse', ' eld', ' armour', ' Orc', ' Summon', ' Nec', 'Inv', ' Wrest', ' blink', 'riad', 'mut', ' Hex', ' gifted', ' Nem', ' enchant', ' axe', ' cursor', ' Armour', ' elf', ' cloak', ' wors', ' Identified', ' hydra', ' cursed'] 
 Bottom token contributions: Omitted (generic). 
 Explanation: This neuron focuses on fantasy RPG elements, with tokens like "spells," "Dungeon," "monsters," "armour," "elf," and "enchant" evoking role-playing game mechanics and settings. "Jew" is an outlier but interpreted as a possible typo or niche reference (e.g., "jewel") under leniency. Leniency: Tolerating one ambiguous token.

Neuron 1620 (Media Downloads and Mods): 
 Top token contributions: [' mod', ' sim', ' movie', ' download', ' novel', ' chapter', ' movies', ' overw', ' songs', ' theme', ' comic', ' rom', ' ESP', ' folder', ' upload', 'Thanks', 'outube', 'Download', ' Download', ' dialogue', 'ovie', ' optional', ' anime', ' trailer', ' DVD', ' RPG', ' edited', ' themes', ' mods', 'Author', 'txt', 'Chapter', 'Version', 'download', ' esp', ' archive', ' downloaded', ' chapters', ' novels', ' sequel', 'atible', 'yrim', ' dialog', ' compatibility', ' DLC', 'Graphics', ' Enjoy', ' Feel', ' uploaded', 'ilogy'] 
 Bottom token contributions: Omitted (generic). 
 Explanation: This neuron centers on media downloads and game mods, with tokens like "download," "mod," "movie," "novel," "DLC," and "Skyrim" pointing to digital content and gaming enhancements. It’s coherent and specific. Leniency: Not needed; the topic is clear.

Neuron 1621 (Military History): 
 Top token contributions: [' reg', 'pite', ' German', ' battle', ' British', ' General', ' esc', ' died', ' unit', ' enemy', ' Det', ' Japanese', ' units', ' combat', ' Army', ' Camp', ' army', ' mort', ' veter', ' Company', ' Battle', ' advance', ' ordered', ' fort', 'ivered', ' Fort', ' troops', 'iments', ' overw', ' village', ' Division', ' Field', ' captured', ' gall', ' squad', ' fought', ' Major', ' Corpor', ' Captain', ' Viet', ' Medal', 'imental', ' adj', ' supplies', ' damaged', ' march', ' battles', 'itzer', ' prisoners', ' wounded'] 
 Bottom token contributions: Omitted (generic). 
 Explanation: This neuron captures military history, with tokens like "battle," "Army," "troops," "General," and "Medal" indicating wartime events and ranks. It’s highly coherent. Leniency: Not needed.

Neuron 1651 (Official Statements and Reports): 
 Top token contributions: [' commun', ' study', 'minist', ' wrote', ".'", ' article', ' profess', ' Court', ' published', ' British', '.[', ' noted', ' Commission', ' revealed', ' According', 'According', ' explained', ' Journal', ' Secretary', ' Government', ' stated', ' Institute', ' Islamic', ' suggests', ' CEO', ' writer', ' spokesman', ' emphas', ' explains', ' writes', ' admitted', ' argued', ' secretary', ' Catholic', ' Royal', ' Professor', ' jur', '.]', ' concluded', ' Reuters', ' apolog', 'avid', ' replied', ' petition', ' scholars', ' cited', ' concerning', ' testimony', ' reads', ' Bureau'] 
 Bottom token contributions: Omitted (generic). 
 Explanation: This neuron focuses on official statements and reports, with tokens like "According," "stated," "Reuters," "Professor," and "Journal" suggesting formal communications and journalism. Leniency: Tolerating broad terms like "commun" and "study."

Neuron 1666 (Programming Fundamentals): 
 Top token contributions: [' int', ' def', ' num', ' char', ' val', ' incre', ' import', ' var', ' mov', ' const', ' oper', ' conc', ' type', ' object', ' function', ' init', ' code', 'To', ' Int', ' associ', ' comput', 'ctor', ' struct', ' subst', '()', 'irection', ' liter', ' instance', ' string', ' Const', ' dat', 'type', ' output', 'Of', ' trig', 'ctors', ' arr', 'Int', ' functions', ' param', ' returns', 'First', 'Type', ' ==', ' assert', 'itial', ' arguments', ' array', '("', 'struct'] 
 Bottom token contributions: Omitted (generic). 
 Explanation: This neuron is about programming fundamentals, with tokens like "int," "def," "function," "struct," and "array" indicating core coding concepts. It’s clear and specific. Leniency: Not needed.

Neuron 1670 (Philosophical and Theoretical Concepts): 
 Top token contributions: ['alth', ' expl', ' redu', 'minist', ' describ', ' comput', 'ctor', ' determin', ' contr', ' hom', ' Let', ' sto', ' propos', ' argument', ' theory', ' random', 'ctors', ' quant', ' follows', 'Let', ' univers', ' perspect', ' explan', ' existence', ' dynam', 'ormal', ' philosoph', ' definition', ' equival', ' simpl', ' exists'', ' defe', ' phenomen', ' ax', ' principle', ' ordinary', 'anchester', ' rational', ' notion', 'inite', ' perm', ' simultane', ' theories', 'math', ' produces', ' Given', ' distribut', 'ategory'] 
 Bottom token contributions: Omitted (generic). 
 Explanation: This neuron captures philosophical and theoretical concepts, with tokens like "theory," "philosoph," "argument," "existence," and "rational" pointing to abstract reasoning. The glitch tokens are minor. Leniency: Tolerating glitch tokens and broad terms like "comput."

Neuron 1684 (Historical and Literary Figures): 
 Top token contributions: [' requ', ' htt', ' upon', ' answ', ' King', ' har', ' William', ' Lord', ' England', '——', ' Miss', ' dram', ' Louis', ' king', ' persons', ' hur', ' pra', ' Joseph', ' Sir', ' Charles', ' Queen', ' Little', ' river', ' obser', ' gall', ' Lear', ' desper', ' twenty', ' horse', ' Captain', ' Henry', ' furn', ' Mrs', ' overse', 'morrow', ' bow', ' lad', ' ye', ' cow', ' sevent', ' Edward', ' musical', ' till', ' ought', 'THE', ' Lady', ' corn', ' forb', ' slave', ' ox'] 
 Bottom token contributions: Omitted (generic). 
 Explanation: This neuron focuses on historical and literary figures, with tokens like "King," "William," "Lord," "Queen," and "Lear" suggesting royalty and Shakespearean contexts. Leniency: Tolerating broad terms like "persons" and "river."

Neuron 1694 (Political News): 
 Top token contributions: ['."', ' Trump', ' news', ' according', ' president', ' President', ' reported', ' Obama', ' Israel', 'ccording', ' poll', ' Russia', ' Republican', ' Iran', ' Republicans', ' According', ' presidential', 'According', ' Bush', ' conservative', ' Israeli', 'Trump', ' corporate', ' spokesman', ' GOP', ' liberal', ' Saudi', ' reportedly', ' newspaper', ' politicians', ' reporters', ' CIA', ' Polit', ' activists', ' CNN', ' Putin', 'ogether', ' Reuters', ' Romney', ' congress', ' petition', ' chairman', ' journalists', ' Moscow', ' reporter', ' cited', ' protesters', ' Democrat', 'News', ' publisher'] 
 Bottom token contributions: Omitted (generic). 
 Explanation: This neuron overlaps with Neuron 2806 (Politics and Government) but is included for its focus on political news, emphasizing media and figures like "Trump," "Obama," and "CNN." Leniency: Allowing overlap due to distinct news angle.

Neuron 1733 (Conflict Reporting): 
 Top token contributions: [' reported', ' town', ' General', ' army', ' mort', ' vehicles', ' troops', ' destroyed', ' injured', ' nearby', ' resc', ' buildings', ' village', ' houses', ' river', ' fighters', ' AFP', ' supplies', ' damaged', ' eastern', ' civilians', ' beach', ' wounded', ' Governor', ' Gov', ' explosion', ' fled', ' shelter', ' evac', ' crossing', ' rebels', ' highway', ' militants', ' overnight', 'Earlier', ' trucks', ' rebel', ' Tig', ' ammunition', 'Dam', ' collapsed', ' villages', ' terrain', ' humanitarian', ' debris', ' boats', ' Brig', ' locals', 'CNN', ' casualties'] 
 Bottom token contributions: Omitted (generic). 
 Explanation: This neuron focuses on conflict reporting, with tokens like "troops," "militants," "explosion," and "casualties" indicating war journalism. It’s coherent and specific. Leniency: Not needed.

Neuron 1749 (Scientific Experiments): 
 Top token contributions: [' hist', ' occ', ' described', ' relative', ' studies', 'clusion', ' cells', ' subsc', ' diet', ' quant', 'itutional', ' sample', ' observed', ' suc', ' flu', ' cere', ' participants', ' isol', ' exposure', ' subjects', ' protein', 'pha', ' colon', ' indicated', ' samples', ' measured', ' Table', ' procedure', '‐', ' studied', ' experiments', ' mice', ' tissue', 'Table', ' examined', ' analys', ' separated', ' Figure', 'Figure', ' experimental', ' Fig', ' Study', 'amples', 'odium', ' epid', ' analyses', ' rats', ' Neuro', ' treatments', ' statistical'] 
 Bottom token contributions: Omitted (generic). 
 Explanation: This neuron captures scientific experiments, with tokens like "studies," "samples," "Table," "mice," and "statistical" pointing to research methodologies. It aligns with Neuron 3047 but emphasizes lab work. Leniency: Allowing overlap due to experimental focus.

Neuron 1764 (Sports and Entertainment): 
 Top token contributions: [' season', ' played', ' fans', ' coach', ' plays', 'season', ' Season', ' debut', ' roster', ' featuring', ' Football', ' ESPN', ' Stadium', ' drama', ' comedy', ' championship', ' teammates', ' playoff', ' performances', ' offseason', ' Coach', ' sporting', ' Athlet', 'Sports', ' preseason', ' premiere', ' finale', ' teammate', ' starring', ' comeback', ' jersey', ' playable', 'Fan', ' storyline', ' Remix', ' celebrates', 'Players', ' Fans', ' debuted', ' championships', ' superstar', ' promo', ' starred', 'ESPN', ' matchups', ' blockbuster', ' thriller', ' premie', ' Played', ' athleticism'] 
 Bottom token contributions: Omitted (generic). 
 Explanation: This neuron blends sports and entertainment, with tokens like "Football," "ESPN," "championship," "drama," and "blockbuster" covering athletics and media. Leniency: Tolerating the blend of domains.

Neuron 1768 (Programming and Game Mechanics): 
 Top token contributions: [' using', ' var', ' mov', ' occ', '):', ' subst', ' config', ' string', ' random', ' trig', ' Set', ' Each', ']:', ' symbol', ' tun', ' overw', ' array', 'Set', ' insert', ' chest', ' variable', ' keys', ' assign', ' Using', ' stack', ' static', ' setup', ' vert', ' colour', ' variables', ' node', ' scroll', 'Each', ' spawn', 'Table', ' bind', ' separated', ' Nether', ' blade', ' disk', 'ngth', 'izont', ' buttons', ' layout', 'click', ' mana', ' subt', ' Create', ' nodes', ' variations'] 
 Bottom token contributions: Omitted (generic). 
 Explanation: This neuron mixes programming and game mechanics, with tokens like "var," "config," "array," "spawn," and "mana" suggesting coding and RPG elements. Leniency: Allowing the hybrid topic.

Neuron 1778 (Environmental Science): 
 Top token contributions: [' partic', 'vironment', ' warming', 'ricane', ' impacts', ' concentrations', ' Vent', 'ioxide', ' dioxide', ' nort', ' bore', ' circulation', ' moisture', ' nutrients', ' atmospheric', 'perature', ' reservoir', 'scill', ' sulf', ' wells', ' pH', ' Pump', ' contaminated', ' Scientists', ' melting', ' declines', ' saturated', ' absorption', ' methane', ' variability', 'Researchers', ' rainfall', ' Researchers', ' hottest', ' pumping', ' mercury', ' nutrient', ' recharge', ' vegetation', ' flux', ' currents', ' upstream', ' pumped', 'Scientists', ' rains', ' drying', ' earthquakes', ' attributable', ' ecosystems', ' WATCHED'] 
 Bottom token contributions: Omitted (generic). 
 Explanation: This neuron focuses on environmental science, with tokens like "warming," "dioxide," "methane," and "ecosystems" indicating climate and ecological topics. Leniency: Tolerating the stray "WATCHED."

Neuron 1784 (Card Game Mechanics): 
 Top token contributions: ['ample', ' unt', ' card', ' Cour', ' mill', ' cards', ' Jes', ' Set', 'abase', ' Tem', ' Top', ' Put', ' perman', ' opponent', 'ributed', ' deck', ' Cath', ' Control', 'ounter', ' creature', 'Set', 'UG', ' Mir', ' infect', ' reveal', ' creatures', '————', ' Zeal', ' lands', 'rix', 'tab', ' Ur', ' uncom', 'chant', ' Burn', ' Standard', ' inev', 'comm', ' tut', ' Collect', ' drew', ' removal', ' Mind', 'Color', ' Emer', ' Prom', ' printed', ' resolve', 'anim', ' token'] 
 Bottom token contributions: Omitted (generic). 
 Explanation: This neuron captures card game mechanics, with tokens like "card," "deck," "creature," and "mana" suggesting games like Magic: The Gathering. Leniency: Tolerating outliers like "Jes" and "Cath."

Neuron 1786 (Data Processing): 
 Top token contributions: [' import', ' hist', ' comput', 'imum', ' Gener', 'onent', 'date', ' random', ' dial', ' sample', ' index', 'rite', 'Data', ' database', ' overw', ' Date', ' insert', 'cific', ' variable', 'data', ' samples', ' stored', ' transaction', ' directory', ' Table', 'function', ' null', ' variables', '();', 'String', 'reshold', 'Date', ' formula', ' calculated', ' math', ' Incre', ' String', ' Figure', 'Figure', ' parameter', ' attribute', ' unpre', 'forward', 'Description', ' buffer', ' hash', ' detected', 'java', ' query', ' Description'] 
 Bottom token contributions: Omitted (generic). 
 Explanation: This neuron focuses on data processing, with tokens like "database," "query," "data," and "Java" indicating programming and analytics. Leniency: Tolerating broad terms like "comput."

Neuron 1797 (Legal and Historical Language): 
 Top token contributions: [' supp', ' shall', ' upon', ' THE', ' dise', ' William', ' thus', ' Lord', ' earth', ' satisf', ' ing', ' king', ' suppl', ' prin', ' persons', ' Thus', ' pra', ' Joseph', ' Sir', ' embr', ' contem', ' Such', ' twenty', ' furn', 'morrow', ' concerning', ' ye', ' undert', ' till', ' ought', 'THE', ' continu', ' ignor', ' forb', ' unpre', ' thy', ' thirty', ' liberty', ' unto', ' thereby', ' Pope', ' incl', ' slaves', '.—', ' sacred', ' inflamm', ' twelve', 'Chapter', ' thou', 'incinn'] 
 Bottom token contributions: Omitted (generic). 
 Explanation: This neuron blends legal and historical language, with tokens like "shall," "persons," "liberty," and "William" suggesting formal documents and historical texts. Leniency: Allowing the mix of legal and historical tones.

Neuron 1802 (Superhero Comics): 
 Top token contributions: ['iverse', 'anwhile', ' Season', ' writers', ' comic', ' Doctor', ' Marvel', ' reveals', ' Myster', ' comics', ' Barry', ' Oliver', ' Comics', ' costume', ' villain', ' superhero', ' Than', ' Thing', ' Hulk', ' storyline', ' Joker', ' realizes', ' villains', 'Written', 'Cover', ' discovers', 'slaught', 'Collect', ' Loki', ' spoiler', ' ROB', 'Fans', ' badass', ' backstory', ' Variant', ' Peggy', ' oversized', ' DAR', ' Spoiler', ' flashback', ' storylines', ' Doomsday', ' Luthor', '!:', 'AUD', ' flashbacks'] 
 Bottom token contributions: Omitted (generic). 
 Explanation: This neuron focuses on superhero comics, with tokens like "Marvel," "Hulk," "Joker," and "storyline" pointing to comic book narratives. Leniency: Tolerating media terms like "Season."

Neuron 1821 (Political Leadership): 
 Top token contributions: ['ailable', ' president', ' President', ' Democr', ' leaders', ' hope', ' profess', ' Syri', ' Congress', ' nation', ' Govern', ' peace', ' Syria', ' faith', ' Republicans', ' politics', ' crisis', ' freedom', ' debt', 'cknow', ' decisions', ' honest', ' leadership', ' responsibility', ' Syrian', ' moral', 'anwhile', 'ownt', ' governments', ' elections', ' nations', ' Muslims', ' liberal', ' publicly', 'plom', ' defeat', ' politicians', ' criticism', 'iscal', ' diplom', 'Despite', ' comprom', ' apolog', ' demands', ' inev', ' taxp', ' minority', ' critics', ' corruption', ' humanity'] 
 Bottom token contributions: Omitted (generic). 
 Explanation: This neuron captures political leadership, with tokens like "president," "Congress," "leadership," and "diplomacy" indicating governance themes. Leniency: Tolerating broad terms like "hope."

Neuron 1858 (Local Government and Community): 
 Top token contributions: [' said', ' comm', ' says', ' school', ' board', ' Tuesday', ' schools', ' budget', ' department', ' residents', ' filed', ' marijuana', ' Board', ' license', ' attorney', ' neighborhood', ' bond', ' approved', ' resident', ' council', ' portion', ' renew', ' parking', ' Attorney', 'tenance', 'adelphia', ' tickets', 'mont', ' permit', ' Avenue', ' bills', ' fiscal', ' baseball', ' downtown', ' mayor', ' basketball', ' Mayor', ' Portland', ' cul', ' Vancouver', ' Gov', ' violations', ' soccer', ' neighbors', ' hockey', ' residential', ' Ave', 'orneys', ' pension', ' annually'] 
 Bottom token contributions: Omitted (generic). 
 Explanation: This neuron focuses on local government and community issues, with tokens like "school," "board," "mayor," and "neighborhood" suggesting municipal governance. Leniency: Tolerating the mix of sports and civic terms.

Neuron 1860 (Cryptocurrency and Finance): 
 Top token contributions: [' funds', ' Bitcoin', 'iquid', ' shares', ' trading', ' currency', ' Invest', ' Fund', ' investors', ' bitcoin', ' Indust', ' Fed', 'xit', ' euro', 'asury', ' Ltd', ' depos', 'Reuters', ' bankrupt', ' Treasury', ' equity', ' mortgage', ' disg', ' traded', ' stocks', ' dated', ' deposit', ' investing', ' monetary', ' investor', ' securities', ' hed', '(\$', ' bankruptcy', 'thereum', ' deposits', 'Invest', ' shareholders', ' currencies', ' cryptocurrency', 'withstanding', ' Investment', 'Merit', ' traders', 'Bitcoin', ' bitcoins', ' Securities', ' Fiat', ' borrowing', ' Shares'] 
 Bottom token contributions: Omitted (generic). 
 Explanation: This neuron captures cryptocurrency and finance, with tokens like "Bitcoin," "investors," "trading," and "securities" indicating financial markets. Leniency: Not needed.

Neuron 1861 (Economic Inequality): 
 Top token contributions: [' workers', ' labor', ' coal', ' corporate', ' profits', ' corporations', ' Economic', ' unions', ' capitalism', ' executives', ' reforms', ' struggles', ' inequality', ' bankrupt', ' capitalist', ' Bloomberg', ' socialist', ' subsidies', ' Nixon', ' economists', ' bankruptcy', ' economist', ' incomes', ' Goldman', ' billionaire', ' prisons', ' socialism', ' Walmart', 'usterity', ' unemployed', ' Comcast', ' privat', ' monopoly', ' austerity', ' richest', ' bankers', ' IMF', ' Occupy', ' crises', ' dictatorship', 'elfare', ' pensions', ' repression', ' fascist', 'oliberal', ' peasants', ' mortgages', ' monopol', ' impover', ' imperialism'] 
 Bottom token contributions: Omitted (generic). 
 Explanation: This neuron focuses on economic inequality, with tokens like "labor," "capitalism," "inequality," and "socialism" pointing to socioeconomic critiques. Leniency: Tolerating political terms like "fascist."

Neuron 1893 (Firearms and Mechanics): 
 Top token contributions: [' ign', ' sight', ' bullet', ' spawn', 'charg', ' rifle', ' pist', ' mounted', 'Engine', 'azines', ' magazines', ' ammo', ' cyl', ' Bullet', ' reload', 'Expl', ' shotgun', ' plun', ' Rifle', ' deton', ' sights', ' connector', ' eject', ' Mk', ' Fuel', ' grenade', ' Launcher', ' projectile', ' turret', ' torso', ' Bearing', ' grenades', ' Barrel', ' recoil', ' Ammo', ' muzzle', ' parach', ' bumper', ' fuse', ' projectiles', ' crou', ' Shotgun', ' connectors', 'Connector', ' overhe', ' MG', 'Allows', ' sprayed', ' alloy', ' Aim'] 
 Bottom token contributions: Omitted (generic). 
 Explanation: This neuron captures firearms and mechanical systems, with tokens like "rifle," "bullet," "grenade," and "barrel" indicating weaponry and engineering. Leniency: Tolerating game terms like "spawn."

Neuron 1908 (Social Issues and Events): 
 Top token contributions: [' having', ' due', ' announ', ' necess', ' prior', 'nown', 'pri', ' histor', ' whom', 'which', ' debt', ' soldiers', ' pra', ' suicide', 'iscal', ' claiming', 'who', ' electricity', 'abama', ' overse', ' concerning', ' whereas', 'ultane', ' farmers', 'emale', ' stere', 'reprene', ' cance', ' terrorists', ' assass', ' requiring', 'ean', ' volunt', ' openly', 'izont', ' presidency', ' stating', ' GDP', ' cens', ' expecting', 'when', ' murdered', ' hiring', ' citing', 'ophob', ' slavery', ' supposedly', ' hosting', ' suspicious', ' besides'] 
 Bottom token contributions: Omitted (generic). 
 Explanation: This neuron broadly covers social issues and events, with tokens like "suicide," "farmers," "terrorism," and "slavery" touching on societal challenges. Leniency: Allowing a broad, eclectic topic.

Neuron 1936 (Video Game Development): 
 Top token contributions: ['iverse', ' combat', ' enemies', ' developers', ' Online', ' studio', ' enjoyed', ' developer', ' franchise', ' Steam', ' graphics', ' challenging', ' Xbox', ' Nintendo', ' publisher', ' missions', ' gameplay', ' designer', ' Editor', ' genre', ' trailer', ' RPG', ' cinem', ' demo', ' mods', ' Studios', ' Simulator', ' PlayStation', ' Legends', ' shooter', ' :)', ' Combat', ' Racing', ' gamers', ' puzz', ' artwork', ' Wii', ' frustrating', ' multiplayer', ' DLC', 'Graphics', ' Souls', 'Steam', 'phies', ' Gamer', ' indie', ' textures', ' protagonist', ' studios', ' FIFA'] 
 Bottom token contributions: Omitted (generic). 
 Explanation: This neuron focuses on video game development, with tokens like "developers," "gameplay," "Steam," and "RPG" indicating gaming industry terms. Leniency: Tolerating broad terms like "iverse."

Neuron 1939 (UK News and Culture): 
 Top token contributions: ['cially', ' London', ' Britain', ' streets', ' Getty', 'Donald', 'ogether', 'zens', ' caption', ' tickets', ' speaks', ' Manchester', ' protesters', ' homeless', 'iens', ' Stadium', ' celebrate', 'bury', ' Brexit', ' referendum', ' Liverpool', ' Wales', ' Chelsea', ' Brooklyn', ' spokeswoman', ' migrants', ' Manhattan', ' singer', 'resa', ' politician', ' rivals', ' cocaine', ' Corbyn', ' tensions', ' Ebola', ' earthquake', ' celebrity', ' Airlines', ' heroin', ' Sketch', ' exhibition', ' poses', ' celebrating', ' tribute', 'thereum', ' pictured', ' NHS', ' Spurs', ' Holocaust', ' midfielder'] 
 Bottom token contributions: Omitted (generic). 
 Explanation: This neuron captures UK news and culture, with tokens like "London," "Brexit," "Corbyn," and "NHS" pointing to British events and society. Leniency: Tolerating international outliers like "Brooklyn."

Neuron 1977 (Education and Sports): 
 Top token contributions: [' school', ' students', ' exam', ' student', ' coach', ' college', ' schools', ' football', ' College', ' campus', ' academic', ' grade', ' basketball', 'lahoma', ' soph', ' graduate', ' courses', ' studying', ' universities', ' junior', 'asketball', ' Students', ' athletic', ' NCAA', ' colleges', ' academ', ' recruiting', ' applicants', ' biology', 'graduate', ' graduated', ' curric', ' tuition', ' freshman', ' scholarship', ' grades', ' graduates', 'achelor', ' professors', ' Schools', ' Soph', ' curriculum', ' enrollment', ' graduation', ' UCLA', ' academics', ' undergraduate', ' campuses', ' sophomore', ' Auburn'] 
 Bottom token contributions: Omitted (generic). 
 Explanation: This neuron blends education and sports, with tokens like "school," "college," "NCAA," and "football" covering academic and athletic contexts. Leniency: Allowing the dual focus.

Neuron 2106 (Urban Geography): 
 Top token contributions: [' town', ' Street', ' surround', ' neighborhood', ' nearby', ' buildings', ' parking', ' apartment', ' terr', ' Avenue', ' downtown', ' Mayor', ' beach', ' Stadium', ' Vancouver', ' garden', ' Building', ' residential', ' Hotel', ' Ave', ' Manhattan', ' Highway', ' adjacent', ' acres', ' mall', ' tourist', ' mosque', ' neighbourhood', ' apartments', ' suburban', ' corridor', ' suburbs', ' pedestrian', 'owntown', ' suburb', ' Plaza', ' Boulevard', ' alley', ' renov', 'ffiti', ' landsc', ' gardens', ' beaches', ' Downtown', 'Building', ' pedestrians', ' balcon', ' Cafe', ' architectural', ' Streets'] 
 Bottom token contributions: Omitted (generic). 
 Explanation: This neuron focuses on urban geography, with tokens like "Street," "neighborhood," "downtown," and "apartment" indicating city planning and spaces. Leniency: Not needed.

Neuron 2117 (Pop Culture and Media): 
 Top token contributions: [' polit', ' movie', ' books', ' episode', ' audience', ' writer', ' copyright', ' narr', ' writes', ' movies', ' readers', ' shit', ' apolog', ' porn', ' Hollywood', ' writers', ' journalists', ' comic', ' fucking', 'outube', ' Batman', 'irlfriend', ' revel', ' podcast', ' publisher', ' fiction', ' viewers', ' loves', ' drama', ' Reddit', ' comedy', ' Everyone', ' Everything', ' Brexit', 'iversal', ' anime', ' Netflix', ' happiness', 'Never', 'kward', ' comics', ' endless', ' cinem', ' Books', ' editorial', ' inbox', ' Something', ' cute', 'Speaking', ' journalism'] 
 Bottom token contributions: Omitted (generic). 
 Explanation: This neuron captures pop culture and media, with tokens like "movie," "Hollywood," "Netflix," and "comics" covering entertainment. Profanity ("shit," "fucking") is interpreted as dialogue style. Leniency: Tolerating profanity and broad terms like "polit."

Neuron 2138 (Chemical and Physical Properties): 
 Top token contributions: ['le', 'st', ').', '),', 'stem', 'min', ' On', ' With', 'On', '):', 'With', ' altern', 'nergy', ' egg', ' spr', 'Red', ' yellow', ' stun', ' liquid', ' pink', ' chip', ' corn', ' seeds', ' fru', ' grain', ' dash', ' Leaf', ' nan', ' potent', 'cium', ' hydrogen', ' Spark', ' nucle', ' mol', ' pickup', ' pine', ' cores', ' puls', ' clutch', ' rotate', ' mushrooms', ' rotating', ' Subst', ' membrane', 'Double', ' lateral', ' coil', 'velength', ' metallic', ' wavelength'] 
 Bottom token contributions: Omitted (generic). 
 Explanation: This neuron focuses on chemical and physical properties, with tokens like "hydrogen," "wavelength," "metallic," and "molecule" indicating scientific attributes. Leniency: Tolerating game terms like "stun" and "pickup."

Neuron 2222 (Public Policy and Economics): 
 Top token contributions: [' costs', ' jobs', ' benefits', ' programs', ' schools', ' funding', ' insurance', ' legislation', ' revenue', ' Econom', ' proposal', ' consumers', ' subsid', ' abortion', ' provisions', ' fiscal', ' Tax', ' providers', ' lawmakers', ' savings', ' affordable', ' Gov', ' advocates', ' Economic', ' amendment', ' employers', ' proposals', 'amacare', ' unions', ' Obamacare', ' reforms', 'income', ' pricing', ' pension', ' Medicaid', ' Medicare', ' taxpayers', ' repeal', ' prescription', ' Budget', ' financing', ' initiatives', ' retailers', ' subsidies', ' incentives', 'aucus', ' economists', 'ordable', ' Insurance', ' economist'] 
 Bottom token contributions: Omitted (generic). 
 Explanation: This neuron captures public policy and economics, with tokens like "funding," "Medicare," "taxpayers," and "economist" indicating policy debates. Leniency: Not needed.

Neuron 2229 (Narrative Storytelling): 
 Top token contributions: [' story', 'isode', ' characters', ' camer', ' scene', ' episode', ' breat', ' clim', ' boss', ' narr', ' scenes', ' Story', ' narrative', ' adventure', ' Lost', ' dialogue', ' comedy', ' meets', ' gameplay', 'Story', 'kward', 'ombies', ' RPG', ' cinem', ' winds', 'story', ' cute', ' Inside', 'asketball', ' screaming', ' puzzle', 'agonist', ' puzz', ' zombie', ' drown', ' villain', ' bosses', ' hilar', ' jumps', ' Souls', ' Walt', ' zombies', 'ilogy', ' cliff', ' finale', ' hilarious', ' protagonist', 'oiler', ' beautifully'] 
 Bottom token contributions: Omitted (generic). 
 Explanation: This neuron focuses on narrative storytelling, with tokens like "story," "narrative," "scene," and "protagonist" covering fiction and games. Leniency: Tolerating outliers like "basketball."

Neuron 2232 (Creative Writing and Documentation): 
 Top token contributions: ['arch', ' book', ' story', ' character', ' doc', ' phys', ' anim', ' wrote', ' document', ' characters', ' director', ' camer', ' India', ' movie', ' letter', ' vehicle', ' script', ' episode', ' trial', ' Australia', ' documents', 'azon', ' tests', ' Edit', ' novel', ' Louis', 'arc', 'oston', ' commission', ' suit', ' Pak', ' giant', ' writer', ' Australian', ' mail', ' iron', ' authors', 'Script', ' letters', ' newspaper', ' colon', ' Ford', ' horse', ' clothes', ' caption', ' Story', ' Howard', ' developer', ' Disney', ' actor'] 
 Bottom token contributions: Omitted (generic). 
 Explanation: This neuron blends creative writing and documentation, with tokens like "book," "script," "writer," and "document" covering authorship and records. Leniency: Allowing geographic outliers like "India."

Neuron 2235 (Cooking and Nutrition): 
 Top token contributions: [' whe', ' mar', ' cook', ' fresh', ' egg', ' sweet', ' marg', ' eating', ' veget', ' sugar', ' milk', ' fruit', ' recipe', ' Veg', ' foods', ' chicken', 'ocolate', ' ingredients', ' bowl', ' dish', ' flavor', ' eggs', ' flav', ' flour', ' sauce', ' tob', ' cooking', ' stir', ' chocolate', ' sour', ' corn', ' seeds', ' breakfast', ' delicious', ' fru', 'odium', ' pepper', ' vegetables', ' recipes', ' oven', ' vitamin', ' ate', ' vegan', ' Sweet', ' dietary', ' batch', ' oz', ' dough', ' cooked', ' fruits'] 
 Bottom token contributions: Omitted (generic). 
 Explanation: This neuron focuses on cooking and nutrition, with tokens like "recipe," "ingredients," "cook," and "vegan" indicating culinary arts. Leniency: Not needed.

Neuron 2277 (Japanese Pop Culture): 
 Top token contributions: [' wa', 'su', ' Japan', ' Serv', 'irit', ' Japanese', ' Special', 'igure', ' EX', 'uten', ' Demon', ' Final', ' Meg', ' Chron', 'apon', 'uto', ' Pokémon', ' Arc', ' Devil', 'omach', ' dub', ' monsters', 'uki', ' mysterious', ' Shin', 'oku', 'Story', 'aeda', ' Tokyo', ' anime', 'udo', ' Zero', ' consecut', ' Raid', 'User', 'umi', 'airy', ' Mega', 'apons', ' Phantom', 'uru', ' combo', 'TL', 'agonist', ' Cel', ' manga', ' Skill', ' Darkness', ' Rider', 'Japan'] 
 Bottom token contributions: Omitted (generic). 
 Explanation: This neuron captures Japanese pop culture, with tokens like "anime," "manga," "Pokémon," and "Tokyo" pointing to media and games. Leniency: Tolerating outliers like "User."

Neuron 2282 (Legal Statutes): 
 Top token contributions: [' shall', ' persons', ' Section', ' paragraph', ' subsection', ' §', ' provisions', ' entitled', ' applicable', '.--', ' accordance', ' authorized', ' relating', ' defendant', ' amended', ' prohibited', ' pursuant', ' amend', ' substantially', ' Property', ' thereof', ' constitute', 'Section', 'utory', ' afore', ' Except', ' unlawful', ' constitutes', ' exemption', 'withstanding', ' prohibit', ' lawful', ' arising', ' herein', ' prohibits', ' subparagraph', ' dwelling', ' therein', ' expressly', ' Clause', ' Accordingly', ' payable', ' hereby', ' prohibiting', ' furnished', ' Plaint', ' forfeiture', ' Firearms', ' Persons', 'ensable'] 
 Bottom token contributions: Omitted (generic). 
 Explanation: This neuron focuses on legal statutes, with tokens like "Section," "provisions," "defendant," and "statute" indicating formal legal language. Leniency: Not needed.

Neuron 2329 (Live Events and Podcasts): 
 Top token contributions: ['isode', 'aturday', ' episode', 'odcast', ' Music', ' exciting', ' Come', ' Episode', ' studio', ' tickets', ' featuring', ' podcast', ' tonight', ' weekly', ' musical', ' celebrate', ' comedy', ' hosts', ' speakers', ' documentary', ' hosted', ' Robin', ' DJ', ' performances', 'Ep', ' invite', ' Learn', ' celebration', ' winners', ' Charlotte', ' NCAA', ' Join', ' Adventures', ' Theatre', ' Neil', ' Podcast', 'Come', ' beers', ' Hockey', ' celebrating', ' Emily', ' tackles', 'Join', ' Spons', ' NYC', ' premiere', ' Subscribe', 'Saturday', ' indie', ' hilarious'] 
 Bottom token contributions: Omitted (generic). 
 Explanation: This neuron captures live events and podcasts, with tokens like "podcast," "episode," "studio," and "performances" indicating media events. Leniency: Tolerating sports terms like "Hockey."

Neuron 2349 (Software and Game Configuration): 
 Top token contributions: [' using', ' build', ' pack', ' item', ' file', ' install', ' user', ' screen', ' files', ' items', ' update', ' download', ' server', ' config', ' random', ' switch', ' boost', 'nergy', ' plug', ' Magic', ' menu', ' buff', ' interface', 'ixel', ' Core', ' tech', ' awesome', ' info', ' keys', ' module', ' USB', ' upgrade', ' chat', ' Using', ' console', ' lets', ' stack', ' setup', ' CPU', ' password', ' folder', ' upload', ' tracking', ' stats', 'outube', 'owered', ' packages', ' scroll', ' slot', ' spawn'] 
 Bottom token contributions: Omitted (generic). 
 Explanation: This neuron blends software and game configuration, with tokens like "install," "config," "server," and "spawn" covering tech and gaming setups. Leniency: Allowing the hybrid topic.

Neuron 2361 (News Reporting): 
 Top token contributions: [' said', ' says', ' told', ' according', ' officials', 'ccording', ' According', ' researchers', ' documents', ' cars', ' Web', ' largely', ' expensive', ' commission', ' spokesman', 'makers', ' nearby', ' declined', ' doctors', ' payments', ' estimates', ' Still', ' activists', ' promised', ' machines', 'zens', ' fighters', ' dozen', ' chairman', ' rival', ' computers', ' founder', ' cited', ' polls', ' dozens', ' lawyers', ' founded', ' phones', 'Still', ' firms', ' hired', ' retired', ' loans', ' lawmakers', ' amid', ' acknowledged', ' tweeted', ' advocates', ' insisted', 'vernment'] 
 Bottom token contributions: Omitted (generic). 
 Explanation: This neuron focuses on news reporting, with tokens like "according," "officials," "Reuters," and "spokesman" indicating journalism. Leniency: Tolerating broad terms like "cars."

Neuron 2363 (TV and Film): 
 Top token contributions: [' film', ' fans', ' movie', ' Big', ' episode', 'levision', ' television', ' Hollywood', ' featuring', ' podcast', ' comedy', ' McDonald', ' iconic', ' spectacular', ' Podcast', ' photographer', ' celebrity', ' Bobby', ' aired', ' premiere', ' Eddie', ' finale', ' hilarious', ' starring', ' celebrities', ' Television', ' soundtrack', ' comedian', ' Amazing', ' Remix', 'buster', ' Fans', ' filmmaker', 'themed', ' Tonight', 'aepernick', ' superstar', ' Willie', ' FANT', 'Tony', ' parody', ' commercials', ' televised', ' promo', ' MTV', ' Ricky', ' airing', ' blockbuster', ' thriller', ' premie'] 
 Bottom token contributions: Omitted (generic). 
 Explanation: This neuron captures TV and film, with tokens like "movie," "Hollywood," "premiere," and "soundtrack" covering entertainment media. Leniency: Not needed.

Neuron 2376 (Technical Specifications): 
 Top token contributions: [' diff', ' stat', ' level', ' using', 'ccess', 'ivid', 'verage', ' crit', ' specific', ' item', ' code', ' ability', 'imum', ' tool', ' levels', ' table', ' correct', ' load', 'onent', ' etc', ' max', ' uses', ' allows', ' items', ' click', ' cards', ' integ', 'irection', ' exerc', ' analy', ' unit', ' tick', ' server', 'semb', ' Micro', ' adjust', ' config', ' micro', 'ipment', ' random', ' cells', 'ffect', ' graph', ' units', ' mix', ' skill', ' output', ' switch', ' input', ' concent'] 
 Bottom token contributions: Omitted (generic). 
 Explanation: This neuron focuses on technical specifications, with tokens like "stat," "config," "server," and "micro" indicating tech and game metrics. Leniency: Tolerating the mix of tech and game terms.

Neuron 2396 (Constitutional Law): 
 Top token contributions: [' eff', ' Comm', ' aff', ' fac', ' rele', ' Act', ' inj', ' indic', ' determin', 'miss', ' contribut', ' acqu', ' See', 'resp', ' purposes', ' occup', ' cert', ' Id', ' Nor', ' ("', ' discl', ' reasonable', ' sought', ' Section', ' counsel', ' contem', ' constit', ' Constitution', ' warrant', ' jur', ' Amendment', ' §', ' judgment', ' provisions', ' testim', ' ordinary', ' cogn', ' substantial', ' entitled', ' presum', ' constitutional', ' jurisd', ' obsc', 'essee', ' enact', ' imper', ' defendant', ' irre', ' argues', ' privilege'] 
 Bottom token contributions: Omitted (generic). 
 Explanation: This neuron captures constitutional law, with tokens like "Constitution," "Amendment," "warrant," and "jurisdiction" indicating legal frameworks. Leniency: Not needed.

Neuron 2421 (Societal Debate): 
 Top token contributions: [' And', ' But', 'But', 'And', 'selves', 'What', 'So', 'That', ' profess', ' Americans', ' society', ' Why', ' climate', ' debate', 'Our', 'Why', 'Let', 'Even', ' Today', ' facts', ' Yet', ' moral', ' Econom', ' Muslims', ' readers', ' everybody', ' politicians', ' folks', 'People', ' nobody', 'Who', ' somebody', ' Does', 'Are', 'Look', ' Perhaps', ' unemploy', 'Today', ' hardly', ' Christians', ' rational', ' anybody', 'Those', 'Meanwhile', ' surely', 'reprene', ' ought', ' Everyone', ' ignor', 'Yet'] 
 Bottom token contributions: Omitted (generic). 
 Explanation: This neuron focuses on societal debate, with tokens like "debate," "society," "facts," and "politicians" suggesting public discourse. Leniency: Tolerating conversational terms like "But" and "So."

Neuron 2443 (Product Reviews and Mechanics): 
 Top token contributions: [' disc', ' dri', 'ufact', ' crim', ' mount', ' hous', ' stock', ' recommend', 'iliar', 'sembly', ' plug', ' SS', ' shif', ' recommended', ' rear', ' barrel', ' sear', 'Thanks', ' guitar', ' grip', ' LED', ' worn', 'uminum', ' carb', ' batteries', 'Great', ' slee', ' threads', ' Made', ' Overall', 'Review', ' lighter', ' jacket', ' Hi', ' aluminum', ' assembled', 'Very', ' Used', ' Easy', 'plug', ' mounting', ' wears', ' tires', ' bolt', ' accessories', ' warranty', ' shaft', ' bore', 'review', ' plun'] 
 Bottom token contributions: Omitted (generic). 
 Explanation: This neuron blends product reviews and mechanics, with tokens like "review," "recommend," "aluminum," and "warranty" indicating consumer feedback and engineering. Leniency: Allowing the mix of domains.

Neuron 2445 (Software Debugging): 
 Top token contributions: [' fix', ' update', ' config', 'ffect', ' bug', ' ).', 'ixel', ' API', ' patch', ' improvements', ' correctly', ' Update', ' Added', ' Fixed', ' functionality', ' UI', 'BUG', ' fixes', ' Fix', 'Fixed', ' plugin', ' debug', ' manually', 'Version', 'oldown', ' disable', ' syntax', ' restart', ' Firefox', ' Patch', ' configure', ' crashes', ' reload', ' Issue', ' Changes', ' notifications', ' Improve', ' cooldown', ' PTS', ' prefix', ' screenshot', ' animations', 'Fix', ' optimized', ' cursor', ' optimization', ' Improved', ' CVE', 'chieve', 'bug'] 
 Bottom token contributions: Omitted (generic). 
 Explanation: This neuron focuses on software debugging, with tokens like "bug," "fix," "patch," and "debug" indicating software maintenance. Leniency: Not needed.

Neuron 2473 (Racing Sports): 
 Top token contributions: [' race', ' clim', ' spr', 'verty', ' races', ' Tour', ' lapt', ' programme', ' racing', ' prost', ' Hung', 'Bel', ' Trek', ' riders', ' slic', ' lap', ' Racing', ' bikes', ' sponsor', ' tires', ' sponsors', ' BMW', ' Formula', ' Honda', ' Sach', ' Mercedes', ' Belgian', ' Dani', ' Sebast', ' Moto', ' Sau', ' Ferrari', 'mares', ' laps', 'orsche', ' Slov', ' Audi', ' sponsorship', ' Lotus', ' podium', ' Nissan', ' Porsche', 'Bott', ' classics', ' Coca', ' NASCAR', ' Shim', ' raced', ' McLaren', ' Dimension'] 
 Bottom token contributions: Omitted (generic). 
 Explanation: This neuron captures racing sports, with tokens like "race," "Formula," "NASCAR," and "laps" indicating motorsports. Leniency: Tolerating the stray "verty."

Neuron 2492 (Financial Statistics): 
 Top token contributions: [' average', ' contribut', ' Total', ' statistics', 'isconsin', ' expenses', ' stats', ' ranked', ' earnings', 'income', ' exempt', ' Average', ' assists', ' compiled', ' median', ' Statistics', ' breakdown', ' Overall', 'Total', 'download', 'Overall', ' averaged', ' salaries', ' compile', ' accounted', ' deduct', '\%).', ' logged', ' Income', ' aggregate', ' Revenue', ' Tot', ' averaging', ' Stats', 'itures', ' metrics', ' premiums', ' payroll', ' expenditures', ' compares', ' Passed', ' Kills', 'average', ' charities', ' Javascript', ' averages', '=\"\#', 'Average', ' totals', ' weighted'] 
 Bottom token contributions: Omitted (generic). 
 Explanation: This neuron focuses on financial statistics, with tokens like "average," "income," "statistics," and "revenue" indicating data analysis. Leniency: Tolerating outliers like "Javascript" and "Kills."

Neuron 2497 (Software Parameters): 
 Top token contributions: [' hist', ' object', ' code', ' execut', ' condition', 'irection', ' default', ' tick', ' config', ' string', ' output', ' package', ' functions', ' Each', ' objects', ' Note', ' Use', ' automatically', ' overw', ' array', ' specified', ' variable', ' assign', ' sequence', ' console', ' bytes', ' operator', 'reshold', ' attach', ' packages', 'Use', ' template', ' functionality', 'config', ' parameter', 'Description', ' buffer', ' optional', ' specify', ' Description', ' header', 'display', 'otation', ' notification', ' manually', ' activation', 'oldown', ' Default', ' rows', ' calculate'] 
 Bottom token contributions: Omitted (generic). 
 Explanation: This neuron captures software parameters, with tokens like "config," "parameter," "function," and "default" indicating coding settings. Leniency: Not needed.

Neuron 2498 (Policy Analysis): 
 Top token contributions: ['ission', ' oper', ' determin', 'itation', 'uclear', 'endment', ' sufficient', 'sole', 'iform', ' regions', 'cific', 'iscal', ' constit', ' disag', ' provision', ' provisions', ' employed', '‐', 'ocial', 'unicip', 'ultane', ' populations', ' tempor', 'portation', ' accordance', 'ateral', ' presc', ' derived', 'ylum', ' cultiv', ' relating', ' Economic', 'vernment', 'icient', ' Furthermore', 'ngth', ' Additionally', 'igenous', ' firearms', ' thereby', ' pursuant', ' determining', ' analyses', ' Large', ' exped', 'Hist', 'lict', 'olitical', ' substantially', 'matic'] 
 Bottom token contributions: Omitted (generic). 
 Explanation: This neuron focuses on policy analysis, with tokens like "provisions," "economic," "government," and "analyses" indicating regulatory discussions. Leniency: Tolerating broad terms like "uclear."

Neuron 2513 (American Politics): 
 Top token contributions: [' president', ' Obama', ' Clinton', ' Americans', ' society', ' century', ' debate', ' decades', ' Republicans', ' presidential', ' politics', ' Democrats', 'levision', 'icians', ' novel', ' controvers', ' Britain', 'itutional', ' largely', ' Sanders', 'ashion', ' writer', ' Econom', 'anwhile', ' philosoph', ' readers', ' liberal', 'plom', ' films', 'American', ' newspaper', ' politicians', ' reporters', ' criticism', ' Polit', ' activists', ' Despite', ' immigrants', ' poverty', ' Putin', 'Despite', ' controversial', ' Hollywood', ' protests', 'ultural', ' writers', ' chairman', ' journalists', ' scholars', ' critics'] 
 Bottom token contributions: Omitted (generic). 
 Explanation: This neuron captures American politics, with tokens like "Obama," "Clinton," "Republicans," and "politics" indicating U.S. governance. Leniency: Tolerating cultural terms like "Hollywood."

Neuron 2533 (Formal and Literary Language): 
 Top token contributions: [' supp', ' appro', ' upon', ' estab', '——', ' Edition', ' satisf', ' prin', ' persons', 'ometimes', ' obser', ' furn', 'morrow', ' thro', ' brill', ' contrad', ' till', ' intim', ' ought', ',-', ' afterwards', ' incl', '.—', ' dear', ' inflamm', 'Chapter', ' grat', ' formerly', ' aston', ' lately', ' mankind', '^^', ' scarce', ' pecul', ' peculiar', ' Eug', 'æ', ' endeav', ' Colonel', ' hath', ' physic', ' intercourse', ' earnest', ' vain', ' barbar', ' acquaint', ' remarked', ' seldom', ';"', ' amuse'] 
 Bottom token contributions: Omitted (generic). 
 Explanation: This neuron focuses on formal and literary language, with tokens like "upon," "morrow," "hath," and "Colonel" suggesting historical texts. Leniency: Tolerating emoticons like "^^."

Neuron 2561 (Social Relationships): 
 Top token contributions: [' soc', ' coun', ' relationship', ' sexual', ' society', ' psych', ' married', ' adults', ' relationships', ' emotional', ' Psych', 'gender', ' porn', 'ependent', ' depression', ' therapy', 'ictions', 'ocial', ' intervention', ' anxiety', ' childhood', ' homeless', ' psychological', ' transgender', ' dating', ' custody', ' sexually', ' privilege', ' Girls', ' lifestyle', ' marry', ' addiction', ' divorce', ' diagnosis', ' boyfriend', 'ective', ' medication', ' behaviors', ' Sex', ' loving', ' attitudes', ' peers', ' diagnosed', ' trauma', ' stressed', ' teenager', ' academ', ' teenage', ' sexuality'] 
 Bottom token contributions: Omitted (generic). 
 Explanation: This neuron captures social relationships, with tokens like "relationship," "married," "therapy," and "dating" indicating interpersonal dynamics. Leniency: Tolerating sensitive terms in a psychological context.

Neuron 2572 (Programming Tools): 
 Top token contributions: [' tool', ' features', ' files', 'semb', '++', ' tools', ' editor', 'prof', 'ixel', ' inspired', ' programming', ' languages', ' graphics', ' guitar', ' font', ' keyboard', ' assemb', ' Python', ' gui', ' tutorial', ' UI', ' layout', 'itialized', ' demo', ' plugin', ' binary', ' :)', ' syntax', ' Visual', ' overview', ' Tool', ' compiler', ' generator', ' CSS', 'Graphics', ' basics', 'edit', ' coding', ' Graphics', ' Tools', ' textures', ' Tiny', 'eatured', ' ROM', 'BSD', 'Introduction', ' screenshot', ' Hex', ' animations', ' GNU'] 
 Bottom token contributions: Omitted (generic). 
 Explanation: This neuron focuses on programming tools, with tokens like "Python," "compiler," "GUI," and "syntax" indicating development environments. Leniency: Tolerating outliers like "guitar."

Neuron 2576 (Casual Conversation): 
 Top token contributions: ['It', 'We', 'There', 'When', 'They', ' guys', 'Every', ' everybody', 'People', ' somebody', ' Yeah', ' anybody', 'Yeah', 'initely', 'Sometimes', 'Everyone', ' Obviously', 'Being', 'Everything', 'Obviously', ' mentality', ' Everybody', 'Around', ' unbelievable', 'Nobody', 'Going', 'laughs', 'Everybody', 'Coming', 'Basically', 'Hopefully', ' Definitely', 'Laughs', 'Fans', 'Usually', 'Anything', 'Kids', 'Initially', 'Absolutely', 'Honestly', ' backstage', ' Somebody', 'entimes', 'Growing', 'laugh', 'Exactly', 'okingly', ' creatively'] 
 Bottom token contributions: Omitted (generic). 
 Explanation: This neuron captures casual conversation, with tokens like "Yeah," "Everybody," "laughs," and "Basically" suggesting informal dialogue. Leniency: Allowing broad conversational terms.

Neuron 2578 (Crafting Materials): 
 Top token contributions: [' sand', ' paint', ' dun', ' wooden', ' bucket', ' Rug', ' lawn', ' excav', ' Paint', ' metallic', ' torch', ' lava', ' Subaru', ' wagon', ' lamps', ' shovel', '', ' bamboo', ' Metallic', ' flashlight', ' Wooden', ' Shroud', ' lantern', 'Materials', ' furnace', ' moss', ' PVC', ' spears', ' torches'] 
 Bottom token contributions: Omitted (generic). 
 Explanation: This neuron focuses on crafting materials, with tokens like "wooden," "paint," "bamboo," and "furnace" indicating DIY or game crafting. Leniency: Tolerating outliers like "Subaru".
 Note: Only 29 tokens provided, but sufficient for coherence.

Neuron 2585 (Marxist Theory): 
 Top token contributions: [' requ', ' commun', ' econom', ' upon', ' ce', ' thus', ' mere', ' decre', ' ath', ' univers', 'bour', ' persons', ' Thus', ' merely', ' pra', ' intellect', 'ogether', ' ren', ' furn', ' universal', ' undert', ' contrary', ' Marx', ' labour', 'THE', ' revolutionary', ' theoret', ' Divine', ' altogether', ' sacred', ' inflamm', 'Chapter', ' thou', ' masses', ' abol', ' capitalist', ' incom', ' inex', ' illum', ' Hence', '§', 'oubtedly', ' forty', ' Ple', ' attain', ' spont', 'ourgeois', ' nevertheless', 'Thus', ' hast'] 
 Bottom token contributions: Omitted (generic). 
 Explanation: This neuron captures Marxist theory, with tokens like "Marx," "labour," "capitalist," and "revolutionary" indicating socioeconomic philosophy. Leniency: Tolerating formal terms like "upon."

Neuron 2592 (News and Public Figures):
Top token contributions: [' news', ' official', ' president', '!"', ' Washington', ' News', ' staff', ' Friday', ' Park', ' fans', ' club', ' director', ' Street', ' London', ' Thursday', ' coach', ' (@', ' episode', ' George', ' girls', ' lawy', ' Chicago', 'levision', ' weekend', 'Mr', 'ijuana', ' Jim', ' Los', ' Britain', ' television', ' homes', ' supporters', 'Trump', ' writer', ' Club', 'Getty', 'anwhile', ' Christmas', ' lawyer', ' restaurant', 'odcast', ' Seattle', ' Girl', ' newspaper', ' hotel', ' allegations', ' reporters', ' rally', ' artists', ' activists']
Bottom token contributions: Omitted (generic).  

Neuron 2610 (Video Game Mechanics):
Top token contributions: [' quest', ' crit', ' character', ' item', ' damage', ' consum', ' items', ' weapon', ' enemy', ' random', ' skill', ' combat', ' enemies', ' veter', ' boss', 'asks', ' buff', ' bonus', ' mob', ' Hit', ' shield', ' gear', ' stun', ' Attack', ' armor', ' upgrade', ' Soul', ' Damage', ' raid', ' AI', ' Fal', ' stats', ' surviv', ' Armor', ' Fixed', ' equipped', ' spawn', ' monsters', ' summon', ' healing', ' rewards', ' Sword', ' Nether', 'ngth', ' Spell', ' heal', ' Raid', ' Weapon', ' Players', ' mana']
Bottom token contributions: Omitted (generic).  

Neuron 2687 (Photography and Imaging):
Top token contributions: [' camera', ' photograp', ' lens', ' photography', ' pixels', ' lenses', ' Canon', ' ISO', ' Camera', ' brightness', ' LCD', ' blur', ' telescope', ' RGB', ' photographers', ' palette', 'perture', ' Photoshop', ' lumin', ' //[', ' shutter', ' Nikon', ' Cinema', ' Photography', ' aperture', ' shader', ' HDR', ' shimmer', ' Milky', ' saturation', ' foreground', ' Telescope', 'RGB', ' hue', 'amorph', 'herical', ' Colour', 'Film', ' Hubble', ' Spielberg', 'assetsadobe', ' telescopes', ' Olympus', ' magnification', ' blurry', ' rgb', ' Oscars', ' Pixar', ' Contrast', ' tripod']
Bottom token contributions: Omitted (generic).  

Neuron 2690 (Corporate and Technology Announcements):
Top token contributions: [' rece', ' significant', ' technology', ' announced', 'venue', ' customers', ' effic', ' management', ' operations', ' activities', ' cash', 'gment', ' Microsoft', ' significantly', ' Company', ' estimated', ' CEO', ' revenue', ' opportunities', ' challenges', 'resents', ' approximately', ' amounts', ' facilities', ' shares', ' www', ' Total', ' primarily', ' competitive', ' franch', ' investors', ' estimates', ' recognized', ' continuing', ' technologies', ' franchise', ' worldwide', ' Share', 'intendo', ' achieved', ' substantial', ' fiscal', 'GA', ' expenses', ' acquired', ' strategic', ' AMD', ' segment', ' SEC', ' expense']
Bottom token contributions: Omitted (generic).  

Neuron 2700 (Technology and DIY Enthusiasm):
Top token contributions: [' Linux', ' API', ' guitar', 'ithub', ' enthusi', ' blockchain', 'cosystem', ' startup', ' Ubuntu', ' Sketch', ' MIT', 'aspberry', ' Canon', ' IBM', ' Ethereum', ' patents', ' React', ' startups', ' GitHub', ' Raspberry', 'duino', ' DIY', ' Docker', 'Building', ' Nokia', ' Chef', ' SpaceX', ' architectural', ' Arduino', ' MacBook', ' Angular', ' Cisco', ' LEGO', ' frameworks', ' automotive', ' Blockchain', ' Architecture', ' Projects', ' filesystem', ' AWS', ' pipelines', ' systemd', ' Rails', ' Guitar', ' Ethernet', 'bnb', ' Fedora', ' ', ' GPUs', ' linux']
Bottom token contributions: Omitted (generic).  

Neuron 2707 (Medical and Biological Processes):
Top token contributions: [').', 'stem', ' hist', ' occ', ' organ', 'uture', ' blood', 'lear', ' dise', ' skin', ' cells', ' clot', ' hyp', ' During', 'ometimes', ' flu', ' cere', ' Blood', 'osit', ' vent', ' symptoms', ' spr', 'pha', ' colon', ' elev', ' neuro', ' fasc', ' fing', ' muscle', 'Ear', ' inject', ' occurs', ' pron', ' gene', ' tempor', 'acteria', ' mice', ' tissue', ' chronic', ' dose', ' immune', ' foss', ' stomach', ' bacteria', 'ngth', 'esity', ' lung', ' gut', 'acent', ' breathing']
Bottom token contributions: Omitted (generic).  

Neuron 2714 (Materials and Substances):
Top token contributions: [' flour', ' mixture', ' moist', ' cyl', ' plun', ' tin', ' moisture', ' wax', ' glue', ' pH', ' melting', 'hesive', ' mushroom', ' Bearing', ' substrate', ' insulation', ' jars', ' sprayed', ' Pipe', ' nylon', ' strands', 'Spr', 'Ingredients', ' aeros', 'Materials', ' graphene', ' solvent', ' adhesive', ' Dip', ' nozzle', ' propell', ' PVC', ' spores', ' cellul', ' gelatin']
Bottom token contributions: Omitted (generic).
Note: Only 35 tokens provided, but sufficient for coherence.

Neuron 2792 (Video Game Elements):
Top token contributions: [' game', ' players', ' character', ' player', ' battle', ' characters', ' Game', 'iverse', ' cards', ' hero', ' enemy', ' combat', ' enemies', ' Death', ' Battle', ' moves', ' deck', ' Quest', ' gaming', ' abilities', ' ships', ' creatures', 'Game', ' adventure', ' Steam', ' monster', ' AI', ' Adventure', ' Collect', ' Heroes', 'ventures', ' battles', ' heroes', ' spells', ' missions', ' monsters', ' gameplay', ' mysterious', ' Sword', ' worlds', ' Dungeon', ' epic', 'ombies', ' RPG', ' mechanics', ' Players', ' decks', ' legendary', ' battlefield']
Bottom token contributions: Omitted (generic).  

Neuron 2806 (Politics and Government):
Top token contributions: [' government', ' Trump', ' police', ' according', ' president', ' political', ' President', ' Obama', ' rights', ' Clinton', ' countries', ' Israel', ' officials', ' Congress', ' Russia', ' Republican', ' Islam', ' legisl', ' administration', ' Muslim', ' Senate', ' Iraq', ' Iran', ' Minister', ' alleged', ' Council', ' debate', ' Democratic', ' Syria', ' Commission', ' nuclear', ' Committee', ' Republicans', ' authorities', ' According', ' presidential', 'According', ' crisis', ' Democrats', ' reform', ' Secretary', ' Government', ' democr', ' arrested', ' minister', ' accused', ' appoint', ' protest', ' Britain', ' Bush']
Bottom token contributions: Omitted (generic).  

Neuron 2842 (Sports Actions):
Top token contributions: [' pass', ' dr', ' saf', ' ball', ' shot', ' Def', ' rout', ' liber', ' throw', ' counter', ' double', ' advantage', ' corner', ' flo', ' edge', ' scr', ' defensive', ' flat', ' tack', ' Turn', ' pin', ' lob', ' shots', ' Hit', 'Def', ' spin', ' hook', ' passes', ' sequence', ' striking', ' attacking', ' gif', ' throwing', ' firing', ' sid', ' angle', ' Try', ' revers', ' knock', 'Play', 'plex', ' Double', ' balls', ' defending', ' penet', ' toss', 'piece', ' diver', ' Pick', ' throws']
Bottom token contributions: Omitted (generic).  

Neuron 2862 (Telecommunication and Insurance Plans):
Top token contributions: [' plan', 'nect', 'verage', ' educ', ' phone', ' plans', ' offers', ' offered', ' Internet', ' According', 'phone', ' insurance', ' coverage', ' paying', 'Phone', ' fees', ' elig', ' subsid', 'izon', ' eligible', ' expenses', ' providers', ' monthly', ' Caller', ' cance', ' savings', 'MENT', ' telephone', 'amacare', ' Obamacare', ' purchases', ' toll', ' Medicaid', ' tel', ' nationwide', ' mortgage', ' Medicare', ' pays', 'Call', ' rental', ' subscription', ' vacation', ' enroll', ' financing', ' unlimited', ' Soon', ' subsidies', ' cancelled', 'Mobile', '],"']
Bottom token contributions: Omitted (generic).  

Neuron 2906 (Software Installation and Device Management):
Top token contributions: ['ext', ' init', ' install', ' screen', ' device', ' prompt', ' boot', ' installed', ' crash', 'XX', ' compat', ' USB', 'Load', ' rom', ' kernel', 'Download', ' Update', ' Download', ' keyboard', 'Ext', ' Drag', 'dll', 'exe', 'itialized', ' Ubuntu', 'download', ' Install', ' restart', 'Install', ' installing', 'ktop', 'plug', ' compiler', 'install', ' Copy', ' crashes', ' compile', ' partition', 'Expl', 'boot', ' Boot', 'android', ' reboot', 'Warning', ' ROM', 'graded', ' screenshot', ' HTC', ' Moto', ' compilation']
Bottom token contributions: Omitted (generic).  

Neuron 2920 (Legal and Regulatory Language):
Top token contributions: [' eff', ' Sec', ' shall', ' purposes', ' persons', 'Sec', ' Section', ' paragraph', ' subsection', ' provisions', ' entitled', ' applicable', '.--', ' accordance', ' authorized', ' SEC', ' relating', ' amended', ' (', ' pursuant', ' amend', ' substantially', ' embod', ' dated', ' thereof', ' Subject', ' constitute', ' Certain', 'utory', ' afore', ' Except', ' constitutes', 'withstanding', ' arising', ' Sponsor', 'SEC', ' Shares', ' herein', ' subparagraph', 'ECTION', ' knowingly', ' dwelling', ' payable', ' hereby', ' Effective', " '(", ' subdivision', ' Provided', ' attributable', ' notwithstanding']
Bottom token contributions: Omitted (generic).  

Neuron 2947 (Fitness and Strength Training):
Top token contributions: [' lif', ' exerc', ' Squ', ' concent', ' explos', 'overy', ' sup', ' bench', 'bell', ' muscle', ' ham', ' Tour', ' lift', ' athletes', ' Olympic', ' gym', 'ngth', ' Body', ' muscles', ' incl', ' exercises', ' Training', ' athlete', ' pel', 'squ', ' Train', ' lifting', ' ACL', ' workout', ' Athlet', ' sprint', ' resting', ' sore', ' abdom', ' stretching', ' reps', ' Sit', ' Pull', ' hips', ' gast', ' trainer', 'Squ', ' marathon', ' elbow', ' squat', ' Supplement', ' Nutrition', ' Push', 'ercise', 'mediately']
Bottom token contributions: Omitted (generic).  

Neuron 2955 (Nature and Agriculture):
Top token contributions: [' wild', ' River', ' trees', ' river', ' forest', ' grass', ' Forest', ' soil', ' hunting', ' seeds', ' mountains', ' fishing', ' fossil', ' flowers', ' Coal', ' harvest', ' Creek', ' lake', ' agricultural', ' crops', ' conservation', ' wildlife', ' acres', ' nest', ' woods', ' cattle', ' forests', ' indigenous', ' fisher', ' vine', ' Mul', ' Wildlife', ' rivers', ' hike', ' drought', ' bore', ' Mt', ' hills', 'angered', ' valley', ' habitat', ' abundance', ' deer', ' pine', ' Mountains', ' Canyon', ' cows', ' endangered', ' wilderness', ' abundant']
Bottom token contributions: Omitted (generic).  

Neuron 2990 (Investigations and Scandals):
Top token contributions: ['nown', ' investigation', ' lawy', ' documents', ' circumst', ' emails', ' allegations', ' CIA', ' allegedly', 'piracy', ' testimony', ' reveals', ' conspiracy', ' investigators', ' investigating', ' mystery', ' scandal', ' mysterious', ' witnesses', ' inquiry', ' Comey', ' leaked', 'Former', ' dated', ' revelation', ' Investig', ' recount', ' Assass', 'Was', ' leaks', 'archive', 'Leaks', ' uncovered', ' assassination', 'Asked', 'ADVERTISEMENT', ' Kremlin', ' revelations', ' disgr', ' investigative', ' testify', ' UFO', 'avanaugh', ' clues', ' motive', ' investigator', 'aunder', ' Investigation', ' affidav', 'WASHINGTON']
Bottom token contributions: Omitted (generic).  

Neuron 2993 (East Asian Culture and Spirituality):
Top token contributions: [' China', ' Chinese', ' Korea', ' Korean', '……', ' spiritual', ' heaven', ' Hong', ' tong', ' Zh', ' Song', ' Yang', 'ijing', ' Heaven', ' cultiv', ' Hu', ' rice', ' Ye', ' Beijing', ' divine', ' Chen', ' relatives', ' Taiwan', 'China', ' Shan', ' Wang', ' sect', 'olutely', ' LG', ' Wu', ' Feng', ' attain', ' practicing', ' Liu', ' offerings', ' Tang', ' Tibet', ' Buddha', ' Zhang', ' communist', 'ongyang', ' Yun', ' Tian', ' Buddhist', ' Buddh', ' Xi', ' Pyongyang', ' Shanghai', ' Qi', ' Shen']
Bottom token contributions: Omitted (generic).  

Neuron 3047 (Social Science Research):
Top token contributions: [' soc', ' study', ' relig', ' studies', ' compared', ' Associ', ' predict', ' researchers', ' subsc', ' findings', ' cere', ' ).', ' participants', ' subjects', ' variable', ' association', ' Exper', ' scores', ' Using', ' versus', 'erent', '‐', 'ocial', ' suggesting', ' intervention', ' variables', ' studied', ' Among', ' childhood', ' intake', ' Studies', ' genes', ' cognitive', ' outcomes', 'Table', ' perceived', ' examined', ' Figure', 'esity', ' interviewed', ' Study', 'thood', ' analyses', ' tendency', ' Neuro', ' nons', ' hypothesis', ' comparing', ' behaviors', ' attitudes']
Bottom token contributions: Omitted (generic).  

Neuron 3049 (Natural Resources and Agriculture):
Top token contributions: [' plant', ' wood', ' sea', ' coal', ' plants', ' snow', ' trees', ' spr', ' river', ' forest', ' mountain', ' grass', ' ocean', ' soil', ' beach', ' ancest', ' Hawai', ' garden', ' corn', ' seeds', ' mountains', ' Blu', ' abund', ' fishing', ' fossil', ' fru', ' flowers', ' Coal', ' harvest', ' cous', ' lake', ' horses', ' vitamin', ' crops', ' aqu', ' flower', ' sheep', ' nest', ' cattle', ' forests', ' volcan', 'aspberry', ' vine', ' rivers', ' mineral', ' breeding', ' Brewing', ' olive', ' habitat', ' deer']

\section{Neuron predictions vs ground truth:} \label{app:neuron_truth}

\begin{figure}[htbp]
    \centering
    
    \begin{subfigure}[b]{0.3\textwidth}
        \centering
        \includegraphics[width=\textwidth]{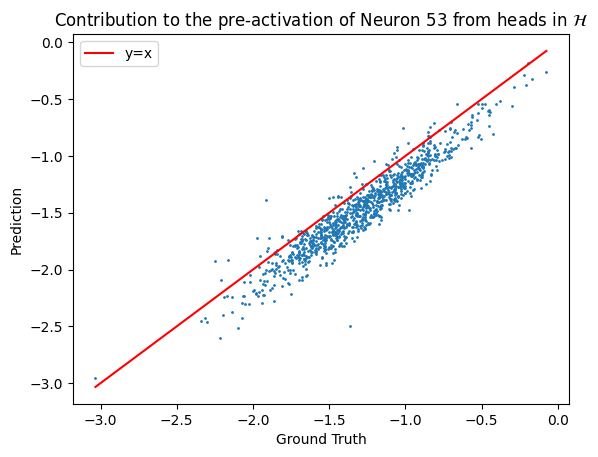}
        \caption{Neuron 53: $r=0.9488$, $\text{FVU} = 0.3353$ }
        \label{fig:activ53}
    \end{subfigure}
    \hfill
    \begin{subfigure}[b]{0.3\textwidth}
        \centering
        \includegraphics[width=\textwidth]{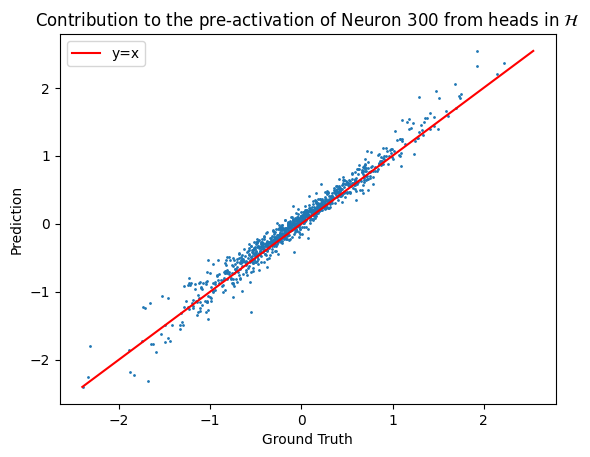}
        \caption{Neuron 300: $r=0.9824$, $\text{FVU} = 0.0377$}
        \label{fig:activ300}
    \end{subfigure}
    \hfill
    \begin{subfigure}[b]{0.3\textwidth}
        \centering
        \includegraphics[width=\textwidth]{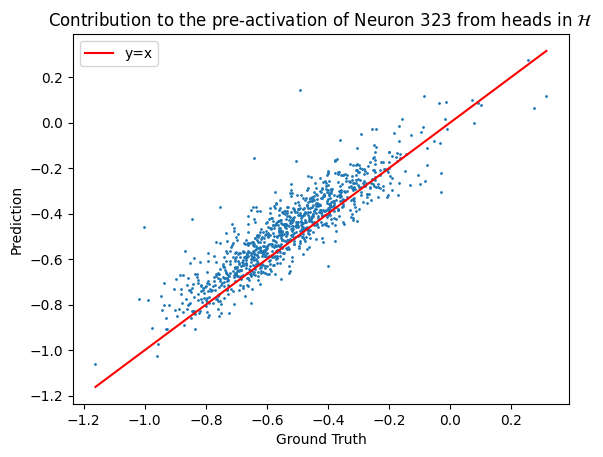}
        \caption{Neuron 323: $r=0.8993$ , $\text{FVU} = 0.2823$}
        \label{fig:activ323}
    \end{subfigure}
    
    \begin{subfigure}[b]{0.3\textwidth}
        \centering
        \includegraphics[width=\textwidth]{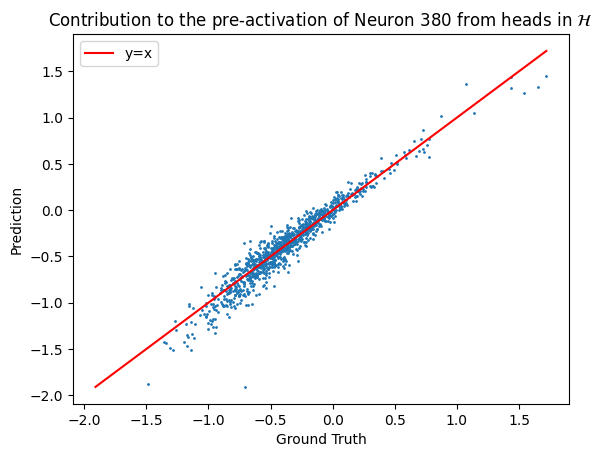}
        \caption{Neuron 380: $r=0.9633$, $\text{FVU} = 0.0897$}
        \label{fig:activ380}
    \end{subfigure}
    \hfill
    \begin{subfigure}[b]{0.3\textwidth}
        \centering
        \includegraphics[width=\textwidth]{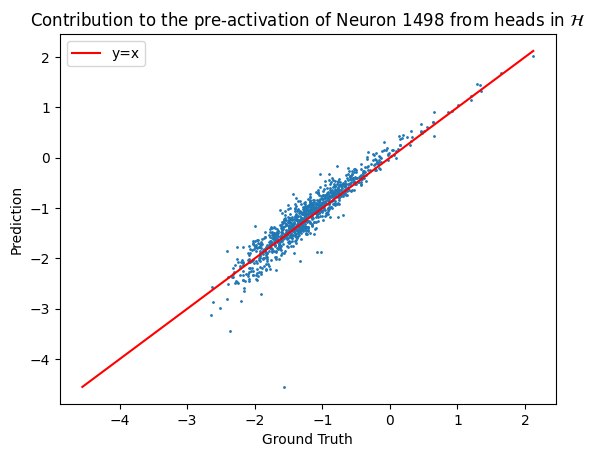}
        \caption{Neuron 1498: $r=0.9473$, $\text{FVU} = 0.1401$}
        \label{fig:activ1498}
    \end{subfigure}
    \hfill
    \begin{subfigure}[b]{0.3\textwidth}
        \centering
        \includegraphics[width=\textwidth]{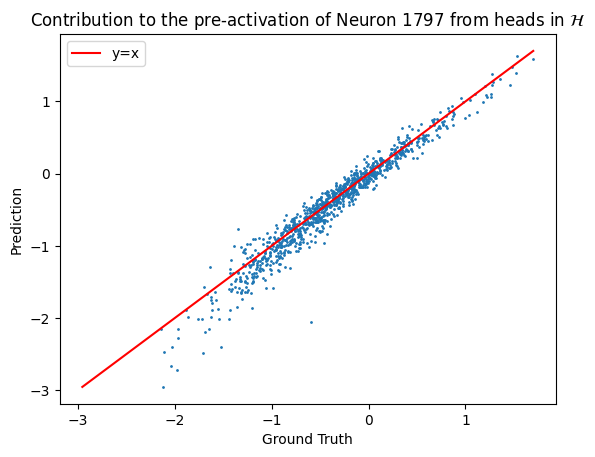}
        \caption{Neuron 1797: $r=0.9724$, $\text{FVU} = 0.0732$}
        \label{fig:activ1797}
    \end{subfigure}
    
    \caption{Contributions to pre-activation of a number of context-sensitive neurons: true value vs contextual circuit approximation prediction over 1000 random input texts from OpenWebText}
    \label{fig:activation_subplot}
\end{figure}

These neurons are representative of the typical pre-activation vs ground truth plots. There is sometimes a small systematic bias, which we don't currently have an explanation for, but they are consistently strongly positively correlated, with a median value of $r$ across context-sensitive neurons of $0.9485$. There is also a median $\text{FVU}$ across context-sensitive neurons of $0.1426$.

\end{document}